\documentclass{article} 
\usepackage{graphicx}
\usepackage[T1,OT1]{fontenc}
\DeclareUnicodeCharacter{201E}{{\fontencoding{T1}\selectfont\quotedblbase}}
\usepackage{iclr2027_conference,times}
\usepackage{bbm}
\usepackage{xspace}
\usepackage{amsfonts,amssymb,amsmath}

\usepackage{amsmath,amsfonts,bm}

\def\eqref#1{equation~\ref{#1}}

\def\1{\bm{1}}

\DeclareMathAlphabet{\mathsfit}{\encodingdefault}{\sfdefault}{m}{sl}
\SetMathAlphabet{\mathsfit}{bold}{\encodingdefault}{\sfdefault}{bx}{n}

\newcommand{\Dict}{\mathcal{D}}
\newcommand{\Wl}{\mathcal{W}}

\usepackage{soul}
\usepackage{xcolor}
\usepackage{CJKutf8}

\newcommand{\arxivglyph}[3]{%
  \leavevmode\pdfliteral direct {/Span << /ActualText <FEFF#1> >> BDC}%
  \raisebox{#2}{#3}%
  \pdfliteral direct {EMC}%
}
\newcommand{\textdevanagari}[1]{\arxivglyph{092A095D}{-.130em}{\includegraphics[height=.751em]{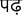}}}
\DeclareUnicodeCharacter{2510}{\arxivglyph{2510}{-.302em}{\includegraphics[height=.628em]{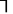}}}
\DeclareUnicodeCharacter{266A}{\arxivglyph{266A}{0pt}{\includegraphics[height=.470em]{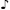}}}
\DeclareUnicodeCharacter{FF1D}{\arxivglyph{FF1D}{0pt}{\includegraphics[height=.503em]{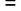}}}
\DeclareUnicodeCharacter{FF5E}{\arxivglyph{FF5E}{0pt}{\includegraphics[height=.432em]{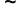}}}

\usepackage{xurl}
\usepackage{hyperref}
\usepackage{fontawesome5}
\definecolor{myblue}{HTML}{2B78D6}
\definecolor{linkblue}{HTML}{1F4E9C}

\hypersetup{
    colorlinks=true,
    citecolor=myblue,
    linkcolor=myblue,
    urlcolor=myblue
}
\usepackage{url}
\usepackage{amsthm}
\newtheorem{proposition}{Proposition}

\usepackage{cleveref}
\usepackage{booktabs}
\usepackage[table]{xcolor}
\usepackage{array}
\usepackage{wrapfig}
\usepackage{subcaption}

\title{Steerable by Design: Conditional Pretraining for Robust Control of Language Models}

\author{Henry Castillo\thanks{Equal contribution.}\\
Cambridge-Boston Alignment Initiative\\
\texttt{henryandrecastillo@gmail.com} \\
\And
Yanchen Liu\footnotemark[1]\\
Massachusetts Institute of Technology \\
\texttt{ychenliu@mit.edu}
\And
Claudio Mayrink Verdun\\
Harvard University \\
Massachusetts Institute of Technology \\
\And
Dylan Hadfield-Menell\\
Massachusetts Institute of Technology \\
}

\colorlet{ourhlcolor}{yellow!45}   

\newcommand{\ourul}[1]{\underline{#1}}

\newcommand{\modelicon}[1]{%
  \raisebox{-0.10em}{\smash{\includegraphics[height=0.78em]{#1}}}%
}
\newcommand{\oursbig}{%
  \modelicon{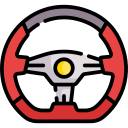}\kern0.12em\textsc{Mancelle-3B}%
}
\newcommand{\ourssmall}{%
  \modelicon{icons/steering1.png}\kern0.12em\textsc{Mancelle-400M}%
}
\newcommand{\pythiabase}{%
  \modelicon{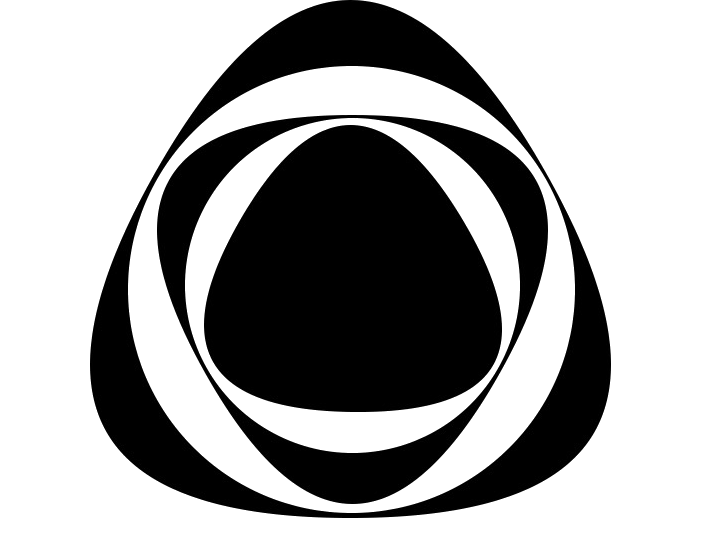}\kern0.12em\textsc{Pythia-2.8B}%
}
\newcommand{\pythia}{%
  \modelicon{icons/eleutherai.png}\kern0.12em\textsc{Pythia-2.8B-SFT}%
}
\newcommand{\qwen}{%
  \modelicon{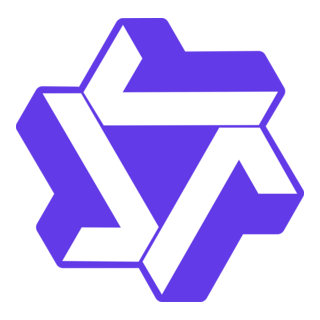}\kern0.12em\textsc{Qwen2.5-3B-Instruct}%
}
\newcommand{\smollm}{%
  \modelicon{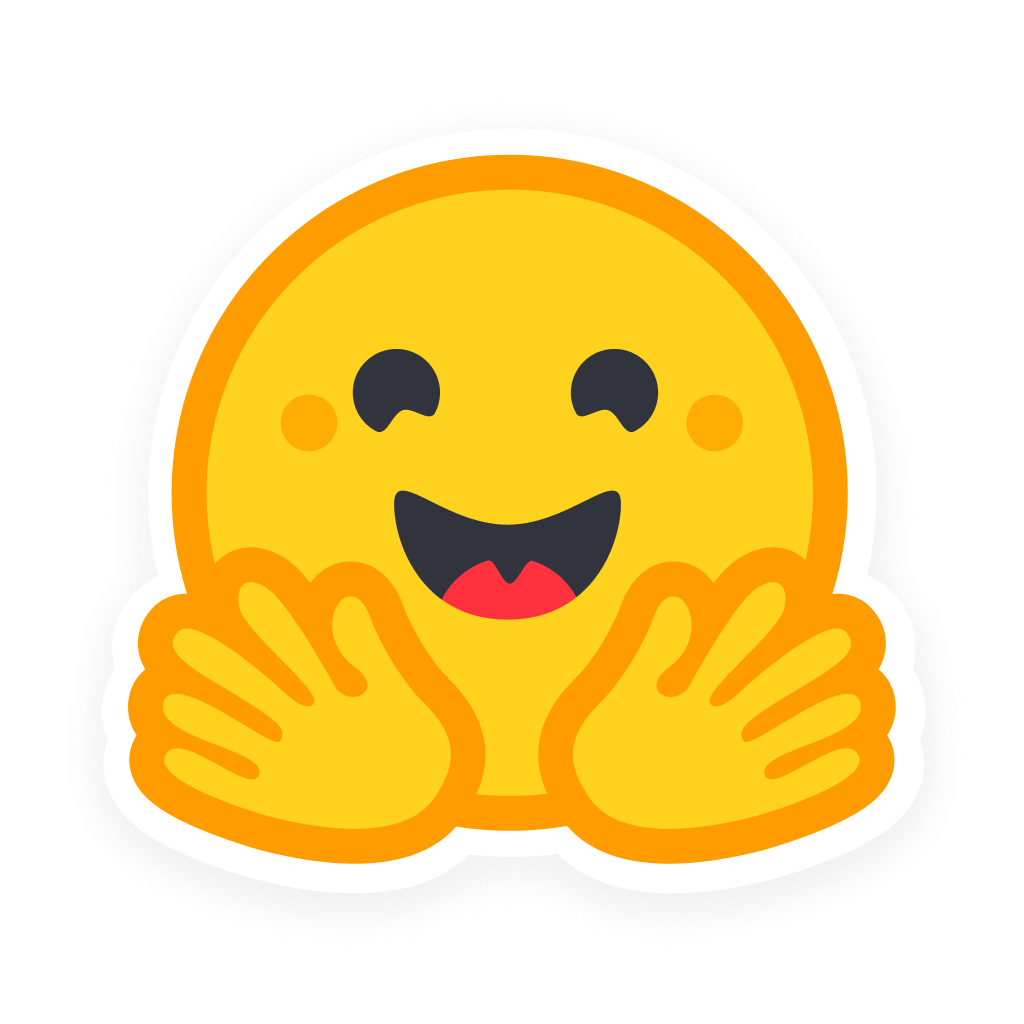}\kern0.12em\textsc{SmolLM3-3B}%
}

\newcommand{\cond}{\emph{Conditioning}} 
\newcommand{\condmask}{\emph{Conditioning (masked)}}
\newcommand{\nlsteer}{\emph{NL Steering}}
\newcommand{\nlencoder}{\emph{NL Steering (encoder)}}
\newcommand{\nlretrieval}{\emph{NL Steering (retrieval)}}
\newcommand{\promptsteer}{\emph{Prompt Steering}}
\newcommand{\lora}{\emph{LoRA}}
\newcommand{\diffmean}{\emph{DiffMean}}
\newcommand{\linearprobe}{\emph{LinearProbe}}

\newcommand{\promptsft}{\emph{Prompt SFT}}
\newcommand{\steerlmsft}{\emph{SteerLM SFT}}

\iclrfinalcopy 
\begin{document}

\newcommand{\yc}[1]{\textcolor{blue}{\textbf{Yanchen:} #1}}
\newcommand{\hc}[1]{\textcolor{red}{\textbf{henry:} #1}}
\newcommand{\cmv}[1]{\textcolor{teal}{\textbf{Claudio:} #1}}
\newcommand{\dhm}[1]{\textcolor{violet}{\textbf{Dylan:} #1}}

\maketitle
\ificlrfinal
\vspace{-1.3em}
{\centering
\href{https://github.com/d0rbu/mancelle}{\textcolor{black}{\faGithub}\kern0.3em\textcolor{myblue}{Code}}\qquad
\href{https://huggingface.co/d0rb/Mancelle-3B}{\raisebox{-0.18em}{\includegraphics[height=1.15em]{icons/huggingface.png}}\kern0.3em\textcolor{myblue}{Models}}\qquad
\href{https://mancelle.com/}{\textcolor{black}{\faGlobe}\kern0.3em\textcolor{myblue}{Website}}\par}
\vspace{1.0em}
\fi

\lhead{}
\renewcommand{\headrulewidth}{0pt}

\begin{abstract}

Our goal is to build language models that are steerable by design. Today's language models are pretrained with next-token prediction, which relegates alignment and control to the final 1\% of training tokens or inference-time interventions. These models repurpose a single input channel for both instructions and data, leading to common failure modes where the model mistakenly treats data as instructions such as intentional jailbreaks and unintentional in-context alignment drift. To correct this, we propose a pretraining method and architecture that separates the two from the very start, producing \oursbig{}. An encoder processes high-level semantic features of upcoming text with cross-attention that prevents the token stream from writing back into this control channel. During pretraining, the model learns to rely on these features to predict the next token, turning the channel into an effective handle for steering the model. \oursbig{} steers more strongly than prompt- and activation-based methods on comparable open models while maintaining fluency, and it withstands adversarial black-box and soft-prompt counter-steering that defeats those baselines. We reproduce this against a more controlled baseline with \ourssmall{}. Our results show that robust control can be built into language models if we do so early and intentionally.

\end{abstract}

\section{Introduction}\label{sec:intro}

We want intelligence whose behavior can be reliably and \textit{robustly} controlled. Despite the efforts and resources that have gone into the modern post-training stack \citep{tie2025survey,lambert2026reinforcement} to build robust control and alignment into models, substantial issues remain. Adversarial prompts have been evolving and keeping up with post-training improvements in a cat-and-mouse game \citep{zou2023universal,hughes2025best,davies2026boundary}; instruction tuning with human demonstrations \citep{ouyang2022training} is complemented by constitutional classifiers \citep{sharma2025constitutionalclassifiersdefendinguniversal} and adversarial training \citep{pmlr-v235-mazeika24a,ICLR2025_1022661f,wang-etal-2026-red}, but adaptive attacks continue to expose vulnerabilities \citep{ICLR2025_63fa7efd,cunningham2026constitutionalclassifiersefficientproductiongrade}. Finetuning can weaken safeguards even when the task is benign \citep{qi2024fine}. Unintentional in-context misalignment is also a concern, especially with recent high-profile incidents \citep{rubygems, collusion_wiki,hf_openai_technical,anthropic2026cyberincidentsalignment}. Alignment drift \citep{lu2026assistantaxissituatingstabilizing} and multi-turn and long-context jailbreaks \citep{russinovich2025greatwritearticlethat,NEURIPS2024_ea456e23,shah2026jailbreakinghaystack} illustrate how strongly context can influence the post-trained persona.

One explanation is that these interventions are post-hoc, guiding a powerful autocompletion engine toward appropriate and correct responses by constructing a chat assistant on top of a base that was optimized for a different objective. High-quality conversation and preference datasets \citep{gao2020pile,penedo2024finewebdatasetsdecantingweb,li2024datacomp} are much smaller than pretraining corpora, leaving models vulnerable to interventions that override their learned behavior. \textit{Context momentum} describes cases where there is conflict between the most likely next token and the most \textit{aligned} next token \citep{anthropic2026cyberincidentsalignment}; \cite{lindsey2025biology} showed this to be a major factor in a successful jailbreak attack on Claude Sonnet 3.5. It is much easier to break down the instruction-data barrier than in other computer systems because it is built on predictive representations, leaving it brittle to methods that can hijack them as illustrated by prefill attacks \citep{ICLR2025_88be0230,struppek2026exposingsystematicvulnerabilityopenweight}.

By virtue of scale and early training exposure, pretraining has outsized effects on the final model's capabilities, behaviors, and generalization \citep{chang2024characterizing,baek2026finetuner,sam2026should,moskvoretskii2026tracing,achille2017critical,feng2026early}. These findings raise the question of whether behavioral control can be made more robust and persistent with earlier, more generalizable interventions. Approaches include model spec midtraining ~\citep{li2026modelspecmidtrainingimproving}, synthetic persona pretraining ~\citep{minder2026syntheticpersonapretrainingalignment}, and synthetic document finetuning ~\citep{wang2025modifying} to shape model priors in beneficial ways. \cite{o2026deep} filter harmful knowledge while \cite{maini2026safety} also rephrase and annotate harmful data. Building on conditional pretraining with binary preference labels \citep{korbak2023pretraininglanguagemodelshuman}, we ask how to extend to a broad vocabulary of semantic controls while maintaining their influence when the text context conflicts with the intended behavior. As opposed to the other described interventions, our aim is twofold: flexible, alignable \textit{structure} that does not require pre-committing to a particular set of values and a rigorous investigation into the efficacy of pretraining interventions for effective and robust control of language generation.

\begin{figure}[!t]
    \centering    \includegraphics[width=1\textwidth]{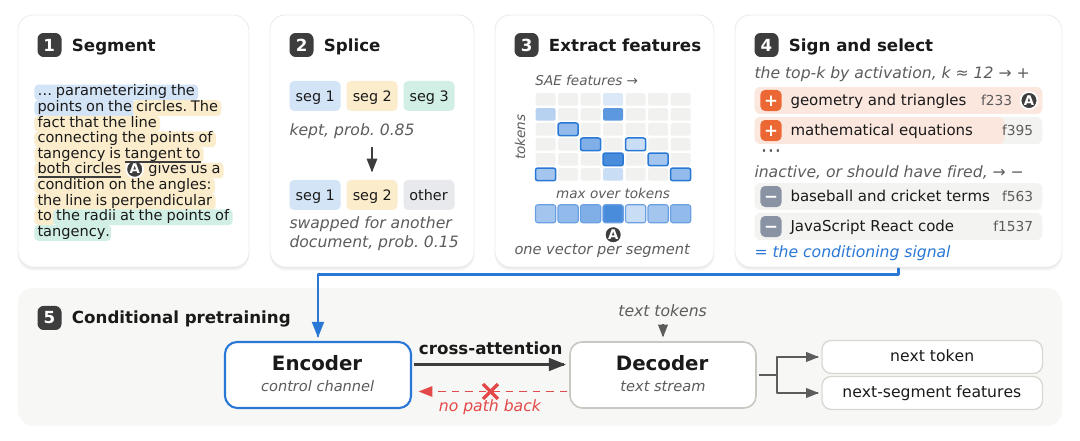}
    \caption{We introduce conditional pretraining with a separate semantic control channel to build robustly controllable models. (1) Documents are tokenized and divided into segments. (2) Segments are spliced to encourage reliance on conditioning signals over previous token context. (3) SAE activations are extracted independently for each segment and max-pooled over token positions; filtering, clustering and selection produce a curated vocabulary of conditioning features. (4) Selected active and inactive conditioning features are encoded as positive and negative tokens, respectively; unselected features are omitted. (5) A dedicated encoder processes these feature tokens and sends them to a decoder through cross-attention. Training combines next-token prediction with an auxiliary objective for predicting the next segment’s features.}  \label{fig:sae_conditional_pretraining}
\end{figure}

We propose a conditional pretraining method that separates semantic control signals from the text being modelled (Figure~\ref{fig:sae_conditional_pretraining}). A feature encoder processes the signals and an autoregressive decoder attends to its representations through one-way cross-attention: the decoder can use the control channel to predict text but the text stream cannot update the encoder's representations, separating instructions from data. The encoder vocabulary is a set of sparse autoencoder (SAE) features we filter and consolidate into semantic conditioning units to improve quality and minimize redundancy. We provide a sparse conditioning signal as positive and negative feature tokens to the encoder for conditional next-token prediction. We add an auxiliary objective for predicting the next segment's features, enabling self-conditioning for fully autoregressive generation with a built-in control channel. We also splice pretraining data to reduce text predictability and encourage reliance on these signals.

We introduce \oursbig{} and \ourssmall{}\footnote{Amédée Bollée's La Mancelle was one of the earliest automobiles with a steering wheel, the first with a bell-crank steering linkage, and the first to be manufactured in series, with fifty built \citep{simionescu2022steering}.}. We evaluate robustness against an adversary who controls model inputs through optimized black-box prompts and white-box soft prompts. Against \pythia{}, \oursbig{} excels in robustness and steering strength without losing fluency or instruction-following ability. Against \qwen{} and \smollm{}, \oursbig{} achieves competitive steering and much higher robustness with less than 5\% of training tokens (\Cref{sec:results_3B}). Against a recipe- and parameter-matched decoder-only model, \ourssmall{} also dominates in steerability and robustness (\Cref{sec:results_400M}).

Thus, we provide three primary contributions: our \textbf{training recipe} including the conditioning signals, dataset splicing, architecture, and post-training pipeline; our \textbf{models and datasets} with feature annotations; and our \textbf{analysis} against similarly-sized open models and a controlled 400M baseline.

\paragraph{Code, data, and model availability.}
We release a \href{https://github.com/d0rbu/mancelle/tree/0a80d5a8fdb14aeef514874fa089fe740aa04a7f}{snapshot of the research code} including our robustness and steering evaluations alongside the \href{https://huggingface.co/d0rb/Mancelle-3B/tree/307c97be4900f9eaf2f9375c7918346c94b07295}{final SFT Mancelle-3B checkpoint}. The \href{https://huggingface.co/buckets/henryac/nano-gpt-pretrain-steer}{pretraining feature data} is available on Hugging Face. The \href{https://github.com/d0rbu/mancelle/releases/tag/paper-data-v1}{supervised fine-tuning dataset} and \href{https://github.com/d0rbu/mancelle/releases/tag/paper-evaluations-v1}{evaluation results and figure reproduction files} are available through GitHub Releases.

\section{Related Work}

\paragraph{Post-training alignment and steering.}  Language models are typically aligned after pretraining through supervised instruction tuning and reinforcement learning from human feedback \citep{ouyang2022training}, AI feedback \citep{bai2022constitutional}, or direct preference optimization \citep{rafailov2023direct}. Complementary methods shape model behavior through prompting, parameter-efficient adaptation such as LoRA \citep{hu2022lora}, or inference-time latent activation interventions including contrastive activation addition \citep{turner2023steering,marks2023geometry,rimsky2024steering}. AxBench compares these approaches and finds steering vectors perform poorly compared to simple prompting and finetuning \citep{wu2025axbenchsteeringllmssimple}. However, finetuning can compromise safety alignment \citep{qi2024fine}. Tamper-resistant training directly addresses attacks that modify model weights \citep{tamirisa2025tamper,o2026deep}. We instead focus on conditional pretraining and evaluate our models' persistence under adversarial prompt and steering vector interventions.

\paragraph{Alignment during pretraining.} \cite{korbak2023pretraininglanguagemodelshuman} compare preference-aware objectives and show that segment-level conditioning on preference labels can reduce undesirable generation while retaining downstream performance, including under adversarial prompting. \cite{maini2026safety} combine filtering, rephrasing, refusal-oriented data, and harmfulness tags while \cite{o2026deep} instead filter dual-use knowledge from pretraining data to improve adversarial finetuning resistance. \cite{minder2026syntheticpersonapretrainingalignment} augment pretraining data with first-person reflections to install a particular persona which then gets bound to the assistant identity. Our work instead focuses on learning a flexible semantic control interface during pretraining with the goal of instruction and data separation, and we evaluate whether this interface remains effective under adversarial prompts and steering vectors.

\paragraph{Conditional language modelling.} Conditioning generation on explicit attributes predates modern LLMs \citep{ficler2017controlling}. CTRL scales this approach using control codes derived from naturally occurring text structure \citep{keskar2019ctrl} such as subreddit source and URL metadata. MeCo uses metadata in pretraining to improve efficiency and enable inference-time steering, including reductions in harmful generations \citep{gao2025metadata}. \cite{fanurls,fan2026beyond} examine how metadata type, granularity, and position affect learning, demonstrating controllable generation and auxiliary metadata prediction. Building on this lineage, we draw on sparse feature discovery to derive conditioning signals from a curated dictionary \citep{cunningham2023sparseautoencodershighlyinterpretable}. Critically, our conditioning is segment-level, providing examples of conditioning signals evolving within a single document and promoting in-context steering rather than whole-document conditional generation. Furthermore, existing techniques only steer within restricted domains due to limited conditioning signal coverage.

\paragraph{Separating control from context.} \cite{wallace2024instruction} train models to respect instruction priority, while StruQ combines structured queries with instruction tuning \citep{chen2025struq}. ASIDE separates instruction and data embeddings through an orthogonal transformation \citep{zverev2026aside}. System-Prompt Anchoring with Cross-Attention Layers adds a dedicated route to system-prompt representations while freezing the pretrained backbone \citep{lixing2026calyrex}. Our model instead learns a feature-based control interface from scratch as the instruction channel, separate from text tokens.

\section{Conditional Pretraining with SAE Features}
\label{sec:method}

We train a language model to predict text conditioned on preceding tokens and a separate semantic signal. Building on conditional pretraining with binary preference labels (\texttt{<good>} and \texttt{<bad>} tokens) \citep{korbak2023pretraininglanguagemodelshuman}, we use a much larger vocabulary of SAE-derived conditioning features and a dedicated encoder to process these signals for prediction. To illustrate the effect of conditioning, consider a simplified model of metadata conditioning \citep{gao2025metadata}. Suppose the training corpus mixes two sources with distinguishable distributions: $f \in \{\texttt{Wikipedia}, \texttt{Reddit}\}$. A model trained on next-token prediction must infer it from the context $x_{:t} = (x_1, \dots, x_{t-1})$:
\begin{equation}
p(x_t \mid x_{:t}) = \sum_{f} p(x_t \mid x_{:t}, f)\, p(f \mid x_{:t}),
\label{eq:marginalize}
\end{equation}
so the belief $p(f \mid x_{:t})$ belongs to whoever writes the context. A few Reddit-like sentences could shift the data source belief and cause the model to fall into a "Reddit mode". A model conditioned with $f$ as an input, on the other hand, predicts $p(x_t \mid x_{:t}, f)$, dropping $p(f \mid x_{:t})$ from the prediction.

We extend from a single binary to a sparse set of positive and negative features about the upcoming text based on sparse autoencoder (SAE) features. A separate feature encoder serves as a channel the token stream can read but not write. At inference, we have the model self-condition (predict $p(f\mid x_{:t})$) so that it can autoregressively generate while still enabling this steering interface.

\subsection{Conditioning Vocabulary and Signal Construction}\label{subsec:conditioning}

A major issue with conditional pretraining is pre-committing. If we hand-pick a small set of signals we deem important and later realize we need to steer along other axes, we need to restart the whole training pipeline from scratch. To address this, we utilize a high-coverage, expressive set of features based on a 65k-latent SAE trained on layer 22 of \texttt{gemma-3-4b-pt} \citep{gemmateam2025gemma3}. We start by filtering features for quality followed by two-stage clustering: we consolidate near-synonyms before grouping related features. We then select the clusters that best cover the corpus. Our aim is expressivity such that any target can be described with these features and steered towards or against.

\paragraph{Curation.}
The quality of SAE features is uneven: some are polysemantic, some activate inconsistently, and many are simply uninterpretable. We retain $16{,}261$ of the $65{,}536$ features by frequency (activates on at least 0.001\% but no more than half of segments) and by quality, judged by GPT-5.6 Luna from Neuronpedia~\citep{neuronpedia} descriptions and top-activating examples.

\paragraph{Consolidation and selection.}
The curated features remain redundant due to feature splitting and absorption. We therefore cluster their description embeddings in two stages, first merging near-synonyms and then grouping related clusters. Next, we select the subset of clusters that best covers the corpus under a budget of at most $100$ active units per segment, which removes the generic, high-frequency clusters that would otherwise appear in every signal. For \oursbig{}, this yields a vocabulary of $844$ units covering $7{,}208$ SAE features (median $5$ members), each with a natural-language description (\Cref{subsec:nl}). \ourssmall{} was trained before the selection step and uses the $2{,}074$-unit vocabulary that the two clustering stages produce; more details in \Cref{app:clustering}.

Let $\mathcal{V}$ be the token vocabulary and $p$ the distribution over documents. For a segment length $L$ and a position $t$, the context and the continuation are
\begin{equation}
  c = x_{:t} = (x_1,\dots,x_{t-1}), \qquad
  y = x_{t:t+L} = (x_t,\dots,x_{t+L-1}).
\end{equation}
The conditioning vocabulary is a finite set $\mathcal{D}$ of $M = |\mathcal{D}|$ units \mbox{($M = 844$ for \oursbig{})}. A \emph{signal} is a map $f:\mathcal{D}\to\{-1,0,+1\}$; we write $f^{+} = \{i : f(i) = +1\}$ for active features and $f^{-} = \{i : f(i) = -1\}$ for inactive ones. We denote by $C$ the context and by $Y$ the continuation drawn from the training distribution.

We write $h_{n,\ell}(y)$ for the activation of feature $n$ at position $\ell$ in segment $y$. Segment-level activity is
\begin{equation}
  u_n(y) = \max_{t \le \ell \le t+L} h_{n,\ell}(y), \qquad
  U_n(y) = \mathbf{1}\{u_n(y) > 0\}.
\end{equation}
Sparsity at a single token does not imply sparsity of $U_n$ over a $128$-token segment. Curation keeps a whitelist $\Wl\subseteq\{1,\dots,N\}$ of latents, and two clustering stages partition $\Wl$ into groups. Each unit $i$ represents a set $G_i$ of retained SAE latents. A unit's activation on a segment is that of its most active member, and the unit is active whenever any member is,
\begin{equation}
z_i(y)=\max_{n\in G_i} u_n(y), \qquad
B_i(y)=\mathbf{1}\{z_i(y)>0\}=\max_{n\in G_i} U_n(y).
\label{eq:unit-activity}
\end{equation}
We write $B(y)=(B_i(y))_{i\in\Dict}\in\{0,1\}^M$ for the activity vector, $A(y)=\{i: B_i(y)=1\}$ for the active set and $I(y)=\Dict\setminus A(y)$ for the inactive set. The firing rate of a unit is the fraction of segments in which it is active, marginally and given a context,
\begin{equation}
s_i=\Pr\big(B_i(Y)=1\big), \qquad
s_i(c)=\Pr\big(B_i(Y)=1\mid C=c\big),
\label{eq:rates}
\end{equation}
where the probability is over a segment $(C,Y)$ drawn from the corpus $p$. Because $B_i(y)$ is a maximum over the $L$ positions of the segment and over the members of the unit, $s_i$ is at least the segment-level rate of each member and far larger than the per-token sparsity of any single latent.

\subsection{Architecture and Training Objectives}\label{subsec:architecture}
A document $x$ is partitioned into $J$ consecutive segments $y_1, \dots, y_J$. Segment $j$ has length $L_j$ under the decoder tokenizer (segments are fixed-length in the SAE tokenizer so $L_j$ varies slightly), context $c_j = (y_1, \dots, y_{j-1})$, and signal $f_j = f(y_j)$
constructed as above. We write $N_{\mathrm{tok}} = \sum_{j=1}^{J} L_j$, and all losses are averaged over documents in a batch. The conditional next-token loss is
$$
\mathcal{L}_{\mathrm{LM}}
=-\frac{1}{N_{\mathrm{tok}}}
\sum_{j=1}^{J}\sum_{\ell=1}^{L_j}
\log p_\theta\!\left(
y_{j,\ell}\mid c_j,y_{j,<\ell},f_{\leq j}
\right).
$$
A feature prediction head $W_{\mathrm{feat}} \in \mathbb{R}^{M \times d}$, where $d$ is the hidden width, sits on the final decoder residual and produces logits $o_{j,\ell} \in \mathbb{R}^M$ at every position. The prediction for segment $j$ is max-pooled over its positions $\hat o_j = \max_{\ell \le L_j} o_{j,\ell}$ elementwise, and is trained to predict the feature activation vector of the next segment, $b_{j+1} = B(y_{j+1})$ with multi-label cross-entropy:
\begin{equation}
\mathcal{L}_{\mathrm{feat}} = -\frac{1}{(J-1)\, M} \sum_{j=1}^{J-1} \sum_{i=1}^{M} \Big[ b_{j+1,i} \log \sigma(\hat o_{j,i}) + (1 - b_{j+1,i}) \log\big(1 - \sigma(\hat o_{j,i})\big) \Big],
\label{eq:feat-loss}
\end{equation}
where $\sigma$ is the logistic function. The total loss is $\mathcal{L}=\mathcal{L}_{\mathrm{LM}}+\lambda\mathcal{L}_{\mathrm{feat}}$ with $\lambda=0.1$ (Table~\ref{tab:training-hyperparameters}). Because $f_j$ is computed from $y_j$, the decoder states of segment $j$ see the same text as $c_{j+1}$ (the counterfactual negatives add information only about a segment outside the document), so at the optimum of $\mathcal{L}_{\mathrm{feat}}$ the head recovers the conditional firing rates of Equation \ref{eq:rates}, $\sigma(\hat o_{j,i})=s_i(c_{j+1})$.

The model is an encoder-decoder transformer with an encoder of the same depth and width as the decoder. The decoder's vocabulary is based on our tokenizer and the encoder's is $\mathcal{D}$, with a special negative-token embedding that is injected additively to token embeddings similar to positional embeddings in \cite{vaswani2017attention}. Signals are therefore ternary, but with $M$ units it is far more expressive than \cite{korbak2023pretraininglanguagemodelshuman}. The encoder processes the feature tokens of a document under a block-diagonal mask according to segment boundaries. This mask is also applied to cross-attention so that decoder positions in $y_j$ attend to $f_j$ and nothing flows from the decoder back into the encoder.

During training, $f_j$ is computed from its corresponding segment $y_j$. In \Cref{sec:theory} we show that the loss floor of a regular token-space decoder is $H(Y\mid C)$ whereas our conditional model's is $H(Y\mid C,F)$ per segment. The gap $I(Y;F\mid C)$ is information not contained in the token context. Therefore, $\mathcal{L}_{\mathrm{LM}}$ teaches the decoder to read the channel, giving us a steering handle in the form of this learned dependency. Signal construction and splicing are chosen to maximize the conditional mutual information, incentivizing the model to incorporate the signals into its predictions more strongly. At inference the model conditions on its own next-segment feature predictions. After segment $j$ is generated, $k^{\pm}$ are sampled from geometric distributions with mean 12, wrapped by a modulo operation into $[1, 48]$ and $[0, 48]$, respectively. $\hat f_{j+1}$ takes the $k^{+}$ largest and $k^{-}$ smallest entries of $\hat o_j$ as positive and negative units.

\subsection{Post-Training}
\label{subsec:posttraining}

Post-training for our models is supervised finetuning (SFT). We start with a 396M-token mixture of SmolTalk \citep{allal2025smollm2smolgoesbig}, the MMLU training set \citep{hendrycks2021measuringmassivemultitasklanguage}, the GSM8k training set \citep{cobbe2021trainingverifierssolvemath}, and bespoke Nanochat-specific datasets for spelling and model identity \citep{nanochat}. Then, we generate 140M tokens of steered synthetic data based on SmolTalk to provide more diversity and examples of steering in a more natural, conversational context. For the synthetic data, we prefer GPT-4.1-mini to small GPT-5 variants. We also add synthetic samples for eliciting multiple-choice answers from internal model knowledge. More details in \Cref{app:training}.

\subsection{Natural Language Steering}
\label{subsec:nl}
Steering through the encoder channel requires a conditioning vector over the encoder units, and fitting one from labelled text for every new concept is a nontrivial cost. We therefore propose and implement a method that maps a one-sentence description of the target behavior to the unit vocabulary. We first embed the sentence and every unit's description with the same sentence-embedding model and score units by cosine similarity. However, descriptions are an imperfect proxy for units' behavioral effects---retrieval alone reaches a similarity of only $0.03$ to the fitted targets---so \nlsteer{} adds a learned correction on top of the retrieval vector: a rank-$32$ residual, initialized to zero and fit once against unit-space difference-of-means targets on concepts disjoint from those we evaluate on. This only requires a sentence description of a concept, and on held-out concepts the encoder raises the cosine similarity to the fitted target from $0.03$ to $0.41$. More details in \Cref{app:nl}.

\section{Experimental Setup}
\label{sec:experimental_setup}

We train encoder-decoder models \oursbig{} and \ourssmall{}. Each model's pretraining dataset requires heavy post-processing to implement our recipe, visualized in Figure~\ref{fig:sae_conditional_pretraining}. \Cref{tab:hparams} reports all relevant hyperparameters, with more details available in \Cref{app:training}. With \texttt{Gemma-3-4b-pt} as our trained base SAE model, we split up pretraining documents from FineWeb-Edu \citep{penedo2024finewebdatasetsdecantingweb} into segments of length 128, which roughly translates to 256-token segments on our tokenizer. For conversation-format SFT data, we split each conversation turn into its own segment. Each turn is further segmented into fixed-length 32-token sequences ($\approx$64 tokens on our tokenizer). The numbers were chosen due to computational trade-offs, i.e. having smaller, more frequent segments increases the amount of required memory and time for pretraining. As shown in \Cref{fig:sae_conditional_pretraining}, we also splice the actual token sequences in the dataset to make the documents less predictable from context, further encouraging the model to rely on conditioning signals.

\paragraph{SAE activation extraction.} We run each segment as its \textit{own independent sequence} through the SAE extraction LLM and SAE encoder, giving us sparse activations in SAE feature space for each token before max-pooling over each segment. This is done to reduce compute requirements and ensure conditioning signals can contain only information about the relevant segment rather than resurfacing signals that are redundant with the previous context.

\paragraph{Steering methods.}
For each concept in AxBench, the Concept500 dataset \citep{wu2025axbenchsteeringllmssimple} provides a positive set of concept-relevant completions and a negative set. These are used as contrastive sets for the latent-space baselines \diffmean{} and \linearprobe{} and as datasets for concept-specific finetuning with \lora{}. The final baseline method is simple \promptsteer{}. We also post-train \pythia{} on \oursbig{}'s steered conversations to follow a steering signal in its prompt (\Cref{app:baselines}): an instruction as in \promptsteer{} (\promptsft{}), or a SteerLM-style \citep{dong2023steerlmattributeconditionedsft} list of our conditioning units with levels (\steerlmsft{}).

We also use these datasets to construct vectors to steer through the control channel of \oursbig{}. These are computed as a difference-of-means, where we compute max-pooled SAE percentile activations for the positive samples and subtract the equivalent activations from the negative samples. We then steer via interpolation (\cond{}) between the model's generated self-conditioning signals and the computed unit-space vector at every segment. We also map a natural-language description of the concept to a conditioning vector with a trained natural-language encoder, which we then use for interpolation steering (\nlsteer{}). We define a strength parameter which controls the steering vector norm for the latent methods and the interpolation parameter for \cond{} and \nlsteer{}. We sweep over 12 steering strength values in the evaluation.

\paragraph{Concepts, prompts and judge.} We use AxBench to evaluate steering: an LLM judge scores each response from 0 to 2 along the axes of concept relevance, instruction-following, and fluency. The clean evaluation uses 100 concepts while the attacks use the first 50, with 10 instructions each.

\paragraph{Attacks.} We consider black-box and white-box attacks on the model's input channel, both of which append a suffix to the instruction to remove the steered concept from the reply. A black-box attacker sees only the steered model's replies while a white-box attacker can take gradients over soft tokens. We attack each method at its best steering strength, the one with the highest harmonic mean of concept and fluency scores. The desired behavior is that the concept survives, or that removing it also breaks the reply; the failure case is an attack that undoes the steering and leaves a usable model.

\underline{GEPA.} GEPA~\citep{agrawal2026gepa} uses a reflective LLM to propose a suffix appended to every instruction, keeping a proposal only when it lowers the concept score on a minibatch. The attack budget $k \in \{0,1,2,4,8\}$ is the number of proposals per concept. 

\underline{RL red-teaming.} Following \citet{perez2022red}, we train an attacker model to write the suffix: \textsc{Qwen2.5-3B-Instruct} \citep{qwen2025qwen25technicalreport} with a LoRA adapter \citep{hu2022lora} reads the concept description, writes a $64$-token suffix, and is trained with GRPO \citep{shao2024deepseekmath} for $40$ steps per model, method, and concept. The reward is the drop in concept score from a local judge we finetuned for this purpose, and it is zero when fluency drops below a floor (\Cref{app:rl-redteam}).

\underline{Soft prompts.} We append a soft prompt composed of 4 continuous tokens \citep{lester2021power,schwinn2024soft} to every instruction and optimize them by gradient descent to prefer reference responses that do not express the concept of interest. We consider two variants (\Cref{app:softprompt}): a free soft prompt, which optimizes the tokens' input embeddings directly, and a convex-hull soft prompt, whose tokens are softmax-weighted mixtures of real token embeddings \citep{guo2021gradient}. On \oursbig{}, we use kernel estimation to map free input vectors to value embeddings to make the attack comparable to open models that do not make use of value embeddings (\Cref{app:softprompt-attack}).

\section{Results}
\label{sec:results_3B}

\paragraph{Baselines at 3B.}
The closest baseline is \pythiabase{}, as it is similar in scale and has been pretrained for a similar amount of tokens at 300B vs our model's 500B (54B over multiple epochs). We finetune it on \oursbig{}'s SFT data without the steered conversations, yielding \pythia{}, which we steer with the four post-hoc methods. Finetuning \pythia{} further on the steered conversations, with the concept in the prompt, gives the two post-trained baselines \promptsft{} and \steerlmsft{} (\Cref{app:baselines}). \qwen{} and \smollm{} are much stronger, having been pretrained on over 10T tokens and post-trained far more extensively.

\subsection{Clean Steering}
\label{sec:clean-3b}

\begin{figure}[!ht]
  \centering
\includegraphics[width=1\textwidth]{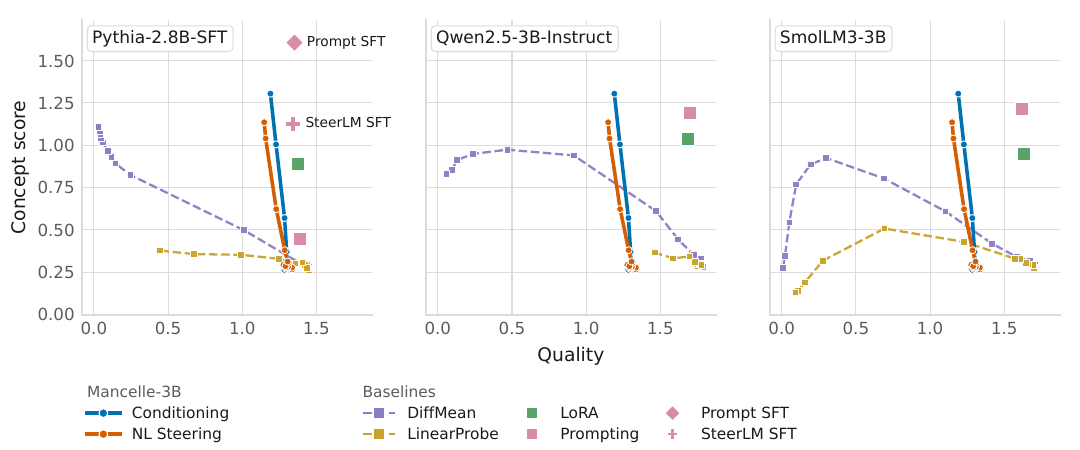}
  \caption{%
    \textbf{Steering comparison: \oursbig{} vs baselines.}
    Against \pythia{} (\underline{left}), \oursbig{} steers more strongly than the four post-hoc methods; \diffmean{} comes close only at strengths where its quality has collapsed. Of the two baselines finetuned on our steered data, \steerlmsft{} steers less strongly and \promptsft{}, which the comparison favors (\Cref{app:baselines}), more strongly, but both are less robust to adversarial prompts (\Cref{sec:blackbox,sec:whitebox}). Against \qwen{} and \smollm{} (\underline{middle}, \underline{right}), we reach comparable steering to the strong \promptsteer{} baseline \citep{wu2025axbenchsteeringllmssimple} with less than 5\% the amount of training tokens, though we lag in overall quality. Concept score measures the LLM-judged presence of a concept in a generated response and quality is measured as the harmonic mean of LLM-judged instruction-following and fluency. More details available in \Cref{app:clean3b}.
  }
  \label{fig:steering_main}
\end{figure}

We ask how much steering each method buys and what it costs in terms of response quality. \Cref{fig:steering_main} plots, for every method and steering strength on the baseline models, the concept score against the harmonic mean of fluency and instruction-following; up and to the right is better.

\paragraph{Inference-time steering cost is not homogeneous.} \lora{} requires a finetune for each concept while \diffmean{} and \linearprobe{} require $144$ examples generated by a language model for each concept before fitting a steering vector~\citep{wu2025axbenchsteeringllmssimple}. \cond{} derives its feature vectors from the same examples (without gradient steps) while \nlsteer{} is as cheap as \promptsteer{}, which likewise asks for nothing more than a sentence.

\subsection{Robustness to Black-Box Adversarial Prompts}
\label{sec:blackbox}

\begin{figure}[!ht]
  \centering
  \includegraphics[width=1\linewidth]{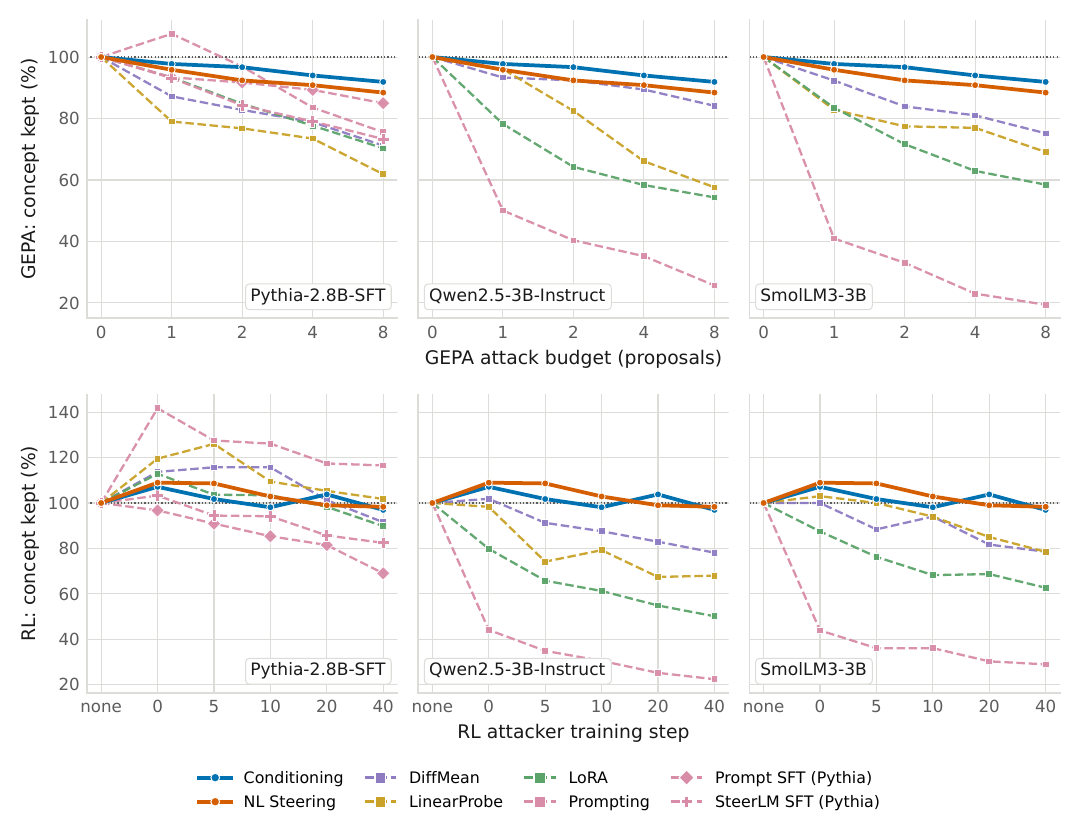}
  \caption{%
 \textbf{Robustness to black-box adversarial prompts.} Concept score kept after the attack, as a proportion of each method's unattacked score, averaged over $50$ concepts. \underline{Top}: GEPA, against the number of proposals. \underline{Bottom}: the RL attacker, against its training step. In both rows the most effective post-hoc baseline methods, \promptsteer{} and \lora{}, are also the most brittle, while \oursbig{}'s steering holds. On \pythia{}, RL curves above $100\%$ come from concepts that were barely steered to begin with. Per-method numbers are in \Cref{app:gepa-table,app:rl-redteam}.
}
  \label{fig:blackbox}
\end{figure}

We attack the steered models with adversarial suffixes written by GEPA~\citep{agrawal2026gepa}, which searches over suffixes proposed by a reflective LLM, and by an attacker model trained with reinforcement learning (\Cref{sec:experimental_setup}). Both see only the steered model's replies, minimize the concept score, and keep the reply fluent. \Cref{fig:blackbox} shows the proportion of each method's concept score that survives, and \Cref{app:gepa,app:rl-redteam} detail the two attacks.

\paragraph{GEPA.} Under counter-steering pressure, our methods keep their steering power robustly whereas the most fluent and effective baselines are much more brittle. \promptsteer{} on \qwen{} and \smollm{} keep $26\%$ and $19\%$ of their original concept scores, respectively, while \cond{} and \nlencoder{} on \oursbig{} keep $92\%$ and $88\%$. The two post-trained baselines on \pythia{} fall in between, keeping $85\%$ (\promptsft{}) and $73\%$ (\steerlmsft{}).

\paragraph{RL red-teaming.} After 40 training steps, the attacker breaks up to 95\% of steered concepts against the baselines. Against \oursbig{}, it only breaks 2 of 37 steered concepts under \cond{} and 1 of 33 under \nlencoder{}; \Cref{app:rl-redteam} defines how we count steered and broken concepts. \pythia{}'s post-hoc methods notably \textit{appear} more robust here, though mostly because they barely steer it to begin with (\Cref{app:rl-counting}). The two post-trained baselines, which do steer it, lose far more than \oursbig{}: the attacker breaks 21 of 47 steered concepts under \promptsft{} and 11 of 33 under \steerlmsft{}, which keep $69\%$ and $82\%$ of their concept score.

\subsection{Robustness to White-Box Adversarial Prompts}
\label{sec:whitebox}

\begin{figure}[!ht]
  \centering
  \includegraphics[width=\linewidth]{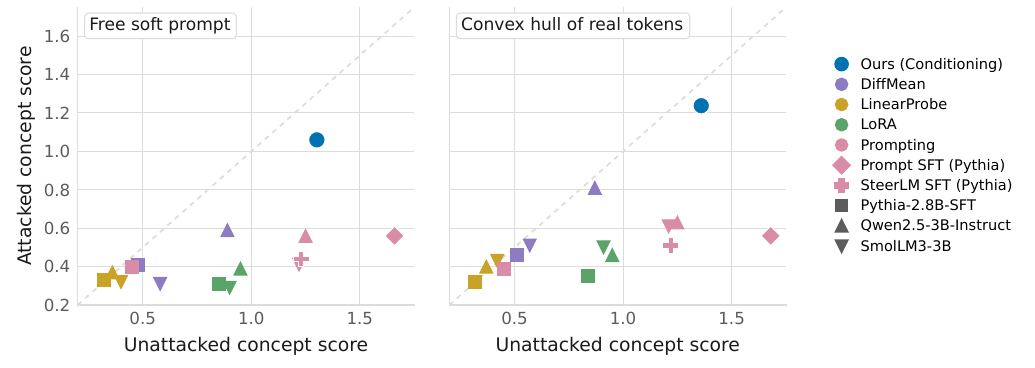}
  \caption{\textbf{Robustness to white-box adversarial prompts.} Attacked concept score against the unattacked score. \underline{Left}: free soft tokens; \oursbig{}'s value embeddings are derived from them. \underline{Right}: soft tokens restricted to mixtures of real tokens. Per-method numbers are in \Cref{app:softprompt}.}
  \label{fig:whitebox}
\end{figure}

We also model an attacker that can optimize a soft prompt against the model, as an intentionally overly adversarial setting that sets an upper bound for token-space attacks (\Cref{sec:experimental_setup}). \Cref{fig:whitebox} shows \oursbig{} still keeps the highest concept score under attack. With free soft tokens, it keeps $81\%$ of its concept score under both \cond{} and \nlencoder{}, with the absolute scores of $1.06$ and $0.97$ staying above that of every baseline method (at most $0.60$). Restricted to mixtures of real tokens \oursbig{} keeps $91\%$ under \cond{} and $92\%$ under \nlencoder{} despite four times as many optimization steps. Its attacked score exceeds every baseline method's by at least $0.42$ and $0.25$, respectively. \promptsteer{} and \lora{} keep about half of their concept score or less while \diffmean{} better retains its already-low starting point. The soft prompts remove far more from the two post-trained baselines on \pythia{} than the black-box attacks did (\Cref{sec:blackbox}): they keep $33$--$42\%$ of their concept score, against $73$--$85\%$ under GEPA and $69$--$82\%$ under RL. \promptsft{}, which starts higher than any other method, ends at $0.56$ under either attack, below both of \oursbig{}'s attacked scores.

When the attack also optimizes \oursbig{}'s value embeddings directly, it removes most of the steering (\Cref{app:softprompt-free}) because it can directly write vectors into $14$ layers unlike the baselines. More generally, attacks that write into the model's internals directly, such as additive steering in the residual stream, weaken our steering about as much as post-hoc steering (\Cref{app:counter-3b}).

\subsection{A Controlled Study at 400M}\label{sec:results_400M}

\begin{figure}[!htbp]
  \centering
  \includegraphics[width=1\linewidth]{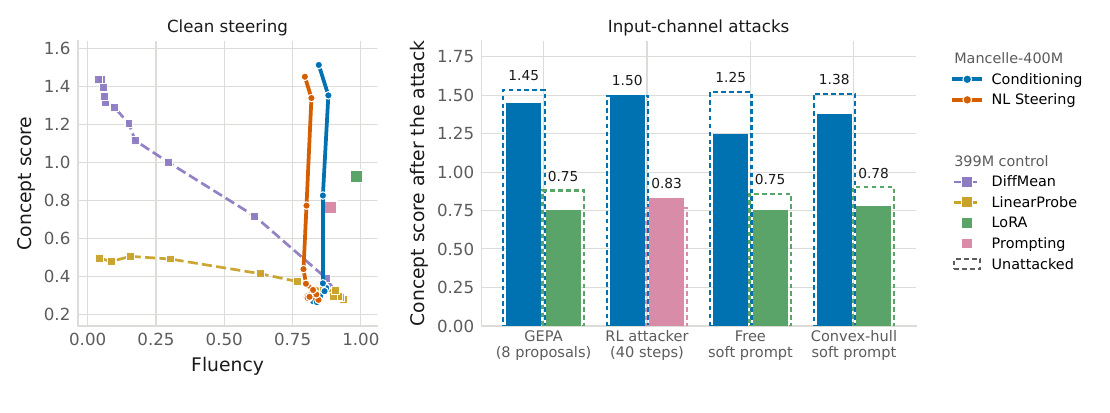}
  \caption{\textbf{Conditioning vs a matched 400M control.} \underline{Left}: similar to \Cref{fig:steering_main}, \ourssmall{} steers far more strongly or fluently than the baselines. \underline{Right}: concept score after each input-channel attack of \Cref{sec:blackbox,sec:whitebox} (bars) and before it (dashed outlines), $50$ concepts, for \cond{} and for the control's method with the highest attacked score; \ourssmall{} ends at $1.25$--$1.50$, against at most $0.83$. All curves and latent counter-steering are in \Cref{app:m400}.}
  \label{fig:m400}
\end{figure}

As a strict control, our \ourssmall{} is compared against decoder-only models trained with the same data and recipe. In \Cref{fig:m400} we present results with a 399M decoder-only model matching our model's 412M parameters, and in \Cref{app:m400} we compare against a 286M model equivalent to ours but without an encoder or cross-attention; the two sets of results agree. All analyses here use the same AxBench-based \citep{wu2025axbenchsteeringllmssimple} setup as in \Cref{sec:results_3B}.

\subsection{Limitations}

\paragraph{Benchmark performance.} \Cref{tab:chatcore-letter-answer-text} shows performance on reasoning, knowledge, and coding benchmarks. Although our model sees a minor performance drop against \pythia{}, we attribute this to nearly 10 epochs' worth of data repetition. More details available in \Cref{app:capability}.

\begin{table*}[t]
\centering
\small
\setlength{\tabcolsep}{2pt}
\begin{tabular}{lrrrrrr}
\toprule
Model & ChatCORE & ARC-E & ARC-C & MMLU & HumanEval \\
\midrule
\oursbig{} & 0.3170 & 0.5500 & 0.4240 & 0.3500 & 0.1402 \\
\addlinespace
\pythia{} & 0.3383 & 0.5960 & 0.4140 & 0.3860 & 0.1524 \\
\addlinespace
\qwen{} & 0.4313 & 0.8520 & 0.7520 & 0.5700 & 0.6890 \\
\addlinespace
\smollm{} & 0.4377 & 0.8860 & 0.7500 & 0.5760 & 0.6768 \\
\bottomrule
\end{tabular}
\caption{ChatCORE \citep{nanochat} scores and accuracies on a subset of tasks. We omit GSM8K and SpellingBee for brevity; \Cref{tab:chatcore-comparison} is available for completeness.}
\label{tab:chatcore-letter-answer-text}
\end{table*}

\paragraph{Dataset processing.} Our datasets are small and repeated for multiple epochs due to compute and storage costs. The compressed dataset for \oursbig{} takes up 2.7TB and requires 5-10\% of the compute budget despite the annotation pipeline running at $\approx$50\% MFU. Labelling an equivalent single-epoch dataset would take around 50\% of the compute needed for the actual pretraining run.

\paragraph{Control channel vulnerability.} An adversary could potentially manipulate token inputs to modify the model's self-predicted features adversarially, though we expect this to require white-box access.

\section{Conclusion}

Current steering methods suffer from brittleness, high test-time compute, or severe model degradation. We are the first to leverage curated, controlled information leakage with conditional pretraining to build a steerable autocompletion engine without these drawbacks. Against strong prompt steering and concept-specific LoRA baselines on more capable models, \oursbig{} steers at comparable strength while holding up more robustly against counter-steering pressure. Against steering vectors, \oursbig{} achieves similar steering with far greater response quality.

Our results show that if we want to achieve behavior that is much more robustly predictable and controllable than what we have today, the promising and scalable approach is mixing alignment into the batter prior to putting the model in the oven rather than only as the icing on top.

\subsection*{AI use statement}

In this work, we used generative AI tools for implementation of experiments, selection of high-quality SAE features for our method (clean and reformat dataset), generating a synthetic post-training dataset, assist in the writing of proofs through typesetting support, and support of qualitative and thematic data analysis through visualization.
We have not used generative AI tools to discover research topics or identify gaps, propose a title or keywords for the research paper, suggest a structure for the research paper, brainstorm, summarize or analyze existing literature, draft parts of the research paper, suggest experimental parameters, interpret results, design or provide feedback on research methodology or experiments, propose or refine hypotheses, help develop theoretical models or conceptual frameworks, formulate mathematical claims, or provide critical ingredients for proving mathematical claims,
and assisting with translation, formulating questions for surveys or interviews, and transcribing recordings of research material are not applicable to this work.
Additionally, we used generative AI tools for running of experiments, launch and management of dataset processing and training runs (creation of artifacts), creation of figures, writing of the majority of our codebase, finding relevant literature, finding information about open-weight models to compare against (sourcing/searching for information), finding BibTeX-formatted citations for papers (format references), and editing the research paper to improve readability. We have reviewed all AI-assisted work. Typeset equations were checked for notation and that they matched the mathematical statements we were writing. AI-written code was reviewed by at least one author. Produced datasets and artifacts were spot-checked by at least one author manually looking at data and model outputs. We take responsibility for the final content of this work, including text, claims or artifacts produced with the aid of generative AI.








\bibliography{iclr2027_conference}

\begin{thebibliography}{86}
\providecommand{\natexlab}[1]{#1}
\providecommand{\url}[1]{\texttt{#1}}
\expandafter\ifx\csname urlstyle\endcsname\relax
  \providecommand{\doi}[1]{doi: #1}\else
  \providecommand{\doi}{doi: \begingroup \urlstyle{rm}\Url}\fi

\bibitem[Achille et~al.(2019)Achille, Rovere, and Soatto]{achille2017critical}
Alessandro Achille, Matteo Rovere, and Stefano Soatto.
\newblock Critical learning periods in deep networks.
\newblock In \emph{International Conference on Learning Representations}, 2019.
\newblock URL \url{https://openreview.net/forum?id=BkeStsCcKQ}.

\bibitem[Agrawal et~al.(2026)Agrawal, Tan, Soylu, Ziems, Khare, Opsahl-Ong,
  Singhvi, Shandilya, Ryan, Jiang, et~al.]{agrawal2026gepa}
Lakshya~A Agrawal, Shangyin Tan, Dilara Soylu, Noah Ziems, Rishi Khare, Krista
  Opsahl-Ong, Arnav Singhvi, Herumb Shandilya, Michael~J Ryan, Meng Jiang,
  et~al.
\newblock Gepa: Reflective prompt evolution can outperform reinforcement
  learning.
\newblock In \emph{International Conference on Learning Representations},
  volume 2026, pp.\  8479--8565, 2026.

\bibitem[Allal et~al.(2025)Allal, Lozhkov, Bakouch, Blázquez, Penedo,
  Tunstall, Marafioti, Lajarín, Kydlíček, Srivastav, Lochner, Fahlgren,
  Nguyen, Burtenshaw, Fourrier, Zhao, Larcher, Morlon, Zakka, Raffel, von
  Werra, and Wolf]{allal2025smollm2smolgoesbig}
Loubna~Ben Allal, Anton Lozhkov, Elie Bakouch, Gabriel~Martín Blázquez,
  Guilherme Penedo, Lewis Tunstall, Andrés Marafioti, Agustín~Piqueres
  Lajarín, Hynek Kydlíček, Vaibhav Srivastav, Joshua Lochner, Caleb
  Fahlgren, Xuan-Son Nguyen, Ben Burtenshaw, Clémentine Fourrier, Haojun Zhao,
  Hugo Larcher, Mathieu Morlon, Cyril Zakka, Colin Raffel, Leandro von Werra,
  and Thomas Wolf.
\newblock {SmolLM2}: When smol goes big --- data-centric training of a fully
  open small language model.
\newblock In \emph{Conference on Language Modeling}, 2025.
\newblock URL \url{https://openreview.net/forum?id=3JiCl2A14H}.

\bibitem[Andriushchenko et~al.(2025)Andriushchenko, croce, and
  Flammarion]{ICLR2025_63fa7efd}
Maksym Andriushchenko, francesco croce, and Nicolas Flammarion.
\newblock Jailbreaking leading safety-aligned llms with simple adaptive
  attacks.
\newblock In Y.~Yue, A.~Garg, N.~Peng, F.~Sha, and R.~Yu (eds.),
  \emph{International Conference on Learning Representations}, volume 2025,
  pp.\  40116--40143, 2025.
\newblock URL
  \url{https://proceedings.iclr.cc/paper_files/paper/2025/file/63fa7efdd3bcf944a4bd6e0ff6a50041-Paper-Conference.pdf}.

\bibitem[Anil et~al.(2024)Anil, Durmus, Panickssery, Sharma, Benton, Kundu,
  Batson, Tong, Mu, Ford, Mosconi, Agrawal, Schaeffer, Bashkansky, Svenningsen,
  Lambert, Radhakrishnan, Denison, Hubinger, Bai, Bricken, Maxwell, Schiefer,
  Sully, Tamkin, Lanhan, Nguyen, Korbak, Kaplan, Ganguli, Bowman, Perez,
  Grosse, and Duvenaud]{NEURIPS2024_ea456e23}
Cem Anil, Esin Durmus, Nina Panickssery, Mrinank Sharma, Joe Benton, Sandipan
  Kundu, Joshua Batson, Meg Tong, Jesse Mu, Daniel Ford, Fracesco Mosconi,
  Rajashree Agrawal, Rylan Schaeffer, Naomi Bashkansky, Samuel Svenningsen,
  Mike Lambert, Ansh Radhakrishnan, Carson Denison, Evan~J Hubinger, Yuntao
  Bai, Trenton Bricken, Timothy Maxwell, Nicholas Schiefer, James Sully, Alex
  Tamkin, Tamera Lanhan, Karina Nguyen, Tomasz Korbak, Jared Kaplan, Deep
  Ganguli, Samuel~R. Bowman, Ethan Perez, Roger Grosse, and David Duvenaud.
\newblock Many-shot jailbreaking.
\newblock In A.~Globerson, L.~Mackey, D.~Belgrave, A.~Fan, U.~Paquet,
  J.~Tomczak, and C.~Zhang (eds.), \emph{Advances in Neural Information
  Processing Systems}, volume~37, pp.\  129696--129742. Curran Associates,
  Inc., 2024.
\newblock \doi{10.52202/079017-4121}.
\newblock URL
  \url{https://proceedings.neurips.cc/paper_files/paper/2024/file/ea456e232efb72d261715e33ce25f208-Paper-Conference.pdf}.

\bibitem[Baek et~al.(2026)Baek, Monti, Schwab, Abbas, Adiga, Blakeney,
  B{\"o}ther, Burstein, Carranza, Deng, et~al.]{baek2026finetuner}
Christina Baek, Ricardo~Pio Monti, David Schwab, Amro Abbas, Rishabh Adiga,
  Cody Blakeney, Maximilian B{\"o}ther, Paul Burstein, Aldo~Gael Carranza,
  Alvin Deng, et~al.
\newblock The finetuner's fallacy: When to pretrain with your finetuning data.
\newblock In \emph{Conference on Language Modeling}, 2026.
\newblock URL \url{https://arxiv.org/abs/2603.16177}.

\bibitem[Bai et~al.(2022)Bai, Kadavath, Kundu, Askell, Kernion, Jones, Chen,
  Goldie, Mirhoseini, McKinnon, et~al.]{bai2022constitutional}
Yuntao Bai, Saurav Kadavath, Sandipan Kundu, Amanda Askell, Jackson Kernion,
  Andy Jones, Anna Chen, Anna Goldie, Azalia Mirhoseini, Cameron McKinnon,
  et~al.
\newblock Constitutional ai: Harmlessness from ai feedback.
\newblock \emph{arXiv preprint arXiv:2212.08073}, 2022.

\bibitem[Bakouch et~al.(2025)Bakouch, Ben~Allal, Lozhkov, Tazi, Tunstall,
  Patiño, Beeching, Roucher, Reedi, Gallouédec, Rasul, Habib, Fourrier,
  Kydlicek, Penedo, Larcher, Morlon, Srivastav, Lochner, Nguyen, Raffel, von
  Werra, and Wolf]{bakouch2025smollm3}
Elie Bakouch, Loubna Ben~Allal, Anton Lozhkov, Nouamane Tazi, Lewis Tunstall,
  Carlos~Miguel Patiño, Edward Beeching, Aymeric Roucher, Aksel~Joonas Reedi,
  Quentin Gallouédec, Kashif Rasul, Nathan Habib, Clémentine Fourrier, Hynek
  Kydlicek, Guilherme Penedo, Hugo Larcher, Mathieu Morlon, Vaibhav Srivastav,
  Joshua Lochner, Xuan-Son Nguyen, Colin Raffel, Leandro von Werra, and Thomas
  Wolf.
\newblock {SmolLM3: smol, multilingual, long-context reasoner}.
\newblock \url{https://huggingface.co/blog/smollm3}, 2025.

\bibitem[Biderman et~al.(2023)Biderman, Schoelkopf, Anthony, Bradley, O'Brien,
  Hallahan, Khan, Purohit, Prashanth, Raff, Skowron, Sutawika, and Van
  Der~Wal]{pmlr-v202-biderman23a}
Stella Biderman, Hailey Schoelkopf, Quentin~Gregory Anthony, Herbie Bradley,
  Kyle O'Brien, Eric Hallahan, Mohammad~Aflah Khan, Shivanshu Purohit,
  Usvsn~Sai Prashanth, Edward Raff, Aviya Skowron, Lintang Sutawika, and Oskar
  Van Der~Wal.
\newblock Pythia: A suite for analyzing large language models across training
  and scaling.
\newblock In Andreas Krause, Emma Brunskill, Kyunghyun Cho, Barbara Engelhardt,
  Sivan Sabato, and Jonathan Scarlett (eds.), \emph{Proceedings of the 40th
  International Conference on Machine Learning}, volume 202 of
  \emph{Proceedings of Machine Learning Research}, pp.\  2397--2430. PMLR,
  23--29 Jul 2023.
\newblock URL \url{https://proceedings.mlr.press/v202/biderman23a.html}.

\bibitem[Bogdan et~al.(2026)Bogdan, Qi, Eaton, Kennedy, Roger, Glynn, Chen,
  Wright, Stegmaier, Kutasov, Foreman-Mackey, Bricken, Carr, Carter,
  MacDiarmid, Marks, Pearce, Simon, Carlini, Burns, Lindsey, Price, and
  Kantamneni]{anthropic2026cyberincidentsalignment}
Paul~C. Bogdan, Richard Qi, Jake Eaton, Sam Kennedy, Fabien Roger, Alex Glynn,
  Runjin Chen, Ben Wright, Otto Stegmaier, Jon Kutasov, Dan Foreman-Mackey,
  Trenton Bricken, Sylvie Carr, Shan Carter, Monte MacDiarmid, Samuel Marks,
  Adam Pearce, Elana Simon, Nicholas Carlini, Collin Burns, Jack Lindsey, Sara
  Price, and Subhash Kantamneni.
\newblock An alignment assessment of recent cybersecurity incidents.
\newblock Anthropic, September 2026.
\newblock URL
  \url{https://www.anthropic.com/research/alignment-assessment-cybersecurity-incidents}.
\newblock Sara Price and Subhash Kantamneni share senior authorship.
  Correspondence to subhash@anthropic.com.

\bibitem[Brown et~al.(2020)Brown, Mann, Ryder, Subbiah, Kaplan, Dhariwal,
  Neelakantan, Shyam, Sastry, Askell, Agarwal, Herbert-Voss, Krueger, Henighan,
  Child, Ramesh, Ziegler, Wu, Winter, Hesse, Chen, Sigler, Litwin, Gray, Chess,
  Clark, Berner, McCandlish, Radford, Sutskever, and
  Amodei]{brown2020languagemodelsfewshotlearners}
Tom Brown, Benjamin Mann, Nick Ryder, Melanie Subbiah, Jared~D Kaplan, Prafulla
  Dhariwal, Arvind Neelakantan, Pranav Shyam, Girish Sastry, Amanda Askell,
  Sandhini Agarwal, Ariel Herbert-Voss, Gretchen Krueger, Tom Henighan, Rewon
  Child, Aditya Ramesh, Daniel Ziegler, Jeffrey Wu, Clemens Winter, Chris
  Hesse, Mark Chen, Eric Sigler, Mateusz Litwin, Scott Gray, Benjamin Chess,
  Jack Clark, Christopher Berner, Sam McCandlish, Alec Radford, Ilya Sutskever,
  and Dario Amodei.
\newblock Language models are few-shot learners.
\newblock In \emph{Advances in Neural Information Processing Systems},
  volume~33, pp.\  1877--1901. Curran Associates, Inc., 2020.
\newblock URL
  \url{https://proceedings.neurips.cc/paper/2020/hash/1457c0d6bfcb4967418bfb8ac142f64a-Abstract.html}.

\bibitem[Chang et~al.(2024)Chang, Tu, and Bergen]{chang2024characterizing}
Tyler~A Chang, Zhuowen Tu, and Benjamin~K Bergen.
\newblock Characterizing learning curves during language model pre-training:
  Learning, forgetting, and stability.
\newblock \emph{Transactions of the Association for Computational Linguistics},
  12:\penalty0 1346--1362, 2024.

\bibitem[Chen et~al.(2025)Chen, Piet, Sitawarin, and Wagner]{chen2025struq}
Sizhe Chen, Julien Piet, Chawin Sitawarin, and David Wagner.
\newblock $\{$StruQ$\}$: Defending against prompt injection with structured
  queries.
\newblock In \emph{34th USENIX Security Symposium (USENIX Security 25)}, pp.\
  2383--2400, 2025.

\bibitem[Child et~al.(2019)Child, Gray, Radford, and
  Sutskever]{child2019generatinglongsequencessparse}
Rewon Child, Scott Gray, Alec Radford, and Ilya Sutskever.
\newblock Generating long sequences with sparse transformers, 2019.
\newblock URL \url{https://arxiv.org/abs/1904.10509}.

\bibitem[Cobbe et~al.(2021)Cobbe, Kosaraju, Bavarian, Chen, Jun, Kaiser,
  Plappert, Tworek, Hilton, Nakano, Hesse, and
  Schulman]{cobbe2021trainingverifierssolvemath}
Karl Cobbe, Vineet Kosaraju, Mohammad Bavarian, Mark Chen, Heewoo Jun, Lukasz
  Kaiser, Matthias Plappert, Jerry Tworek, Jacob Hilton, Reiichiro Nakano,
  Christopher Hesse, and John Schulman.
\newblock Training verifiers to solve math word problems, 2021.
\newblock URL \url{https://arxiv.org/abs/2110.14168}.

\bibitem[Cunningham et~al.(2026)Cunningham, Wei, Wang, Persic, Peng,
  Abderrachid, Agarwal, Chen, Dau, Dimitriev, Howard, Hua, Gilson, Lin, Liu,
  Mikulik, Mittapalli, O'Hara, Pan, Saxena, Silverstein, Song, Zhou, Leike,
  Kaplan, Perez, and
  Sharma]{cunningham2026constitutionalclassifiersefficientproductiongrade}
Hoagy Cunningham, Jerry Wei, Zihan Wang, Andrew Persic, Alwin Peng, Jordan
  Abderrachid, Raj Agarwal, Bobby Chen, Andy Dau, Alek Dimitriev, Logan Howard,
  Yijin Hua, Rob Gilson, Mu~Lin, Christopher Liu, Vladimir Mikulik, Rohit
  Mittapalli, Clare O'Hara, Jin Pan, Nikhil Saxena, Alex Silverstein, Yue Song,
  Giulio Zhou, Jan Leike, Jared Kaplan, Ethan Perez, and Mrinank Sharma.
\newblock Constitutional classifiers++: Efficient production-grade defenses
  against universal jailbreaks.
\newblock In \emph{International Conference on Learning Representations},
  volume 2026, pp.\  48038--48054, 2026.
\newblock URL
  \url{https://proceedings.iclr.cc/paper_files/paper/2026/hash/4f697305ef1f868ad77c3c0027989a6f-Abstract-Conference.html}.

\bibitem[Davies et~al.(2026)Davies, Giglemiani, Lau, Winsor, Irving, and
  Gal]{davies2026boundary}
Xander Davies, Giorgi Giglemiani, Edmund Lau, Eric Winsor, Geoffrey Irving, and
  Yarin Gal.
\newblock Boundary point jailbreaking of black-box {LLMs}, 2026.
\newblock URL \url{https://arxiv.org/abs/2602.15001}.

\bibitem[Dehghani et~al.(2023)Dehghani, Djolonga, Mustafa, Padlewski, Heek,
  Gilmer, Steiner, Caron, Geirhos, Alabdulmohsin, et~al.]{dehghani2023vit22b}
Mostafa Dehghani, Josip Djolonga, Basil Mustafa, Piotr Padlewski, Jonathan
  Heek, Justin Gilmer, Andreas Steiner, Mathilde Caron, Robert Geirhos, Ibrahim
  Alabdulmohsin, et~al.
\newblock Scaling vision transformers to 22 billion parameters.
\newblock In \emph{Proceedings of the 40th International Conference on Machine
  Learning}, volume 202 of \emph{Proceedings of Machine Learning Research},
  pp.\  7480--7512. PMLR, 2023.
\newblock URL \url{https://arxiv.org/abs/2302.05442}.

\bibitem[Dong et~al.(2023)Dong, Wang, Sreedhar, Wu, and
  Kuchaiev]{dong2023steerlmattributeconditionedsft}
Yi~Dong, Zhilin Wang, Makesh Sreedhar, Xianchao Wu, and Oleksii Kuchaiev.
\newblock Steerlm: Attribute conditioned sft as an (user-steerable) alternative
  to rlhf.
\newblock In \emph{Findings of the Association for Computational Linguistics:
  EMNLP 2023}, pp.\  11275--11288, 2023.

\bibitem[Fan et~al.(2025)Fan, Sabol{\v{c}}ec, and Jaggi]{fanurls}
Dongyang Fan, Vinko Sabol{\v{c}}ec, and Martin Jaggi.
\newblock Urls help, topics guide: Understanding metadata utility in llm
  training.
\newblock In \emph{The Thirty-ninth Annual Conference on Neural Information
  Processing Systems}, 2025.

\bibitem[Fan et~al.(2026)Fan, Hashemi, Karimireddy, and Jaggi]{fan2026beyond}
Dongyang Fan, Diba Hashemi, Sai Karimireddy, and Martin Jaggi.
\newblock Beyond urls: Metadata diversity and position for efficient llm
  pretraining.
\newblock In \emph{International Conference on Learning Representations},
  volume 2026, pp.\  44247--44264, 2026.

\bibitem[Feng et~al.(2026)Feng, Ghosal, Springer, Zhong, and
  Raghunathan]{feng2026early}
Lawrence Feng, Gaurav~R Ghosal, Jacob~Mitchell Springer, Ziqian Zhong, and
  Aditi Raghunathan.
\newblock Early data exposure improves robustness to subsequent fine-tuning.
\newblock In \emph{Conference on Language Modeling}, 2026.
\newblock URL \url{https://arxiv.org/abs/2605.12705}.

\bibitem[Ficler \& Goldberg(2017)Ficler and Goldberg]{ficler2017controlling}
Jessica Ficler and Yoav Goldberg.
\newblock Controlling linguistic style aspects in neural language generation.
\newblock In \emph{Proceedings of the Workshop on Stylistic Variation}, pp.\
  94--104, 2017.

\bibitem[Gao et~al.(2020)Gao, Biderman, Black, Golding, Hoppe, Foster, Phang,
  He, Thite, Nabeshima, et~al.]{gao2020pile}
Leo Gao, Stella Biderman, Sid Black, Laurence Golding, Travis Hoppe, Charles
  Foster, Jason Phang, Horace He, Anish Thite, Noa Nabeshima, et~al.
\newblock The pile: An 800gb dataset of diverse text for language modeling.
\newblock \emph{arXiv preprint arXiv:2101.00027}, 2020.

\bibitem[Gao et~al.(2025)Gao, Wettig, He, Dong, Malladi, and
  Chen]{gao2025metadata}
Tianyu Gao, Alexander Wettig, Luxi He, Yihe Dong, Sadhika Malladi, and Danqi
  Chen.
\newblock Metadata conditioning accelerates language model pre-training.
\newblock In \emph{International Conference on Machine Learning}, pp.\
  18612--18629. PMLR, 2025.

\bibitem[{Gemma Team}(2025)]{gemmateam2025gemma3}
{Gemma Team}.
\newblock Gemma 3 technical report, 2025.
\newblock URL \url{https://arxiv.org/abs/2503.19786}.

\bibitem[Guo et~al.(2021)Guo, Sablayrolles, J{\'e}gou, and
  Kiela]{guo2021gradient}
Chuan Guo, Alexandre Sablayrolles, Herv{\'e} J{\'e}gou, and Douwe Kiela.
\newblock Gradient-based adversarial attacks against text transformers.
\newblock In \emph{Proceedings of the 2021 Conference on Empirical Methods in
  Natural Language Processing}, pp.\  5747--5757, 2021.

\bibitem[Hendrycks et~al.(2021)Hendrycks, Burns, Basart, Zou, Mazeika, Song,
  and Steinhardt]{hendrycks2021measuringmassivemultitasklanguage}
Dan Hendrycks, Collin Burns, Steven Basart, Andy Zou, Mantas Mazeika, Dawn
  Song, and Jacob Steinhardt.
\newblock Measuring massive multitask language understanding.
\newblock In \emph{International Conference on Learning Representations}, 2021.
\newblock URL \url{https://openreview.net/forum?id=d7KBjmJ3GmQ}.

\bibitem[Henry et~al.(2020)Henry, Dachapally, Pawar, and Chen]{henry2020qknorm}
Alex Henry, Prudhvi~Raj Dachapally, Shubham~Shantaram Pawar, and Yuxuan Chen.
\newblock Query-key normalization for transformers.
\newblock In \emph{Findings of the Association for Computational Linguistics:
  EMNLP 2020}, pp.\  4246--4253. Association for Computational Linguistics,
  2020.
\newblock \doi{10.18653/v1/2020.findings-emnlp.379}.

\bibitem[Hu et~al.(2022)Hu, Shen, Wallis, Allen-Zhu, Li, Wang, Wang, and
  Chen]{hu2022lora}
Edward~J Hu, Yelong Shen, Phillip Wallis, Zeyuan Allen-Zhu, Yuanzhi Li, Shean
  Wang, Lu~Wang, and Weizhu Chen.
\newblock {LoRA}: Low-rank adaptation of large language models.
\newblock In \emph{International Conference on Learning Representations}, 2022.
\newblock URL \url{https://openreview.net/forum?id=nZeVKeeFYf9}.

\bibitem[Huben et~al.(2024)Huben, Cunningham, Smith, Ewart, and
  Sharkey]{cunningham2023sparseautoencodershighlyinterpretable}
Robert Huben, Hoagy Cunningham, Logan Smith, Aidan Ewart, and Lee Sharkey.
\newblock Sparse autoencoders find highly interpretable features in language
  models.
\newblock In \emph{International Conference on Learning Representations},
  volume 2024, pp.\  7827--7845, 2024.
\newblock URL
  \url{https://proceedings.iclr.cc/paper_files/paper/2024/hash/1fa1ab11f4bd5f94b2ec20e794dbfa3b-Abstract-Conference.html}.

\bibitem[Hughes et~al.(2025)Hughes, Price, Lynch, Schaeffer, Barez, Somani,
  Koyejo, Sleight, Jones, Perez, and Sharma]{hughes2025best}
John Hughes, Sara Price, Aengus Lynch, Rylan Schaeffer, Fazl Barez, Arushi
  Somani, Sanmi Koyejo, Henry Sleight, Erik Jones, Ethan Perez, and Mrinank
  Sharma.
\newblock Best-of-n jailbreaking.
\newblock In \emph{Advances in Neural Information Processing Systems},
  volume~38, 2025.
\newblock URL
  \url{https://proceedings.neurips.cc/paper_files/paper/2025/hash/69f3eb242c7c9df9ea2f2b66ea8b3c0f-Abstract-Conference.html}.

\bibitem[Jordan et~al.(2024)Jordan, Bernstein, Rappazzo, @fernbear.bsky.social,
  Vlado, Jiacheng, Cesista, Koszarsky, and @Grad62304977]{modded_nanogpt_2024}
Keller Jordan, Jeremy Bernstein, Brendan Rappazzo, @fernbear.bsky.social, Boza
  Vlado, You Jiacheng, Franz Cesista, Braden Koszarsky, and @Grad62304977.
\newblock modded-nanogpt: Speedrunning the {NanoGPT} baseline, 2024.
\newblock URL \url{https://github.com/KellerJordan/modded-nanogpt}.

\bibitem[Karpathy(2025)]{nanochat}
Andrej Karpathy.
\newblock nanochat: The best chatgpt that \$100 can buy, 2025.
\newblock URL \url{https://github.com/karpathy/nanochat}.

\bibitem[Karpathy(2026)]{karpathy2026rustbpe}
Andrej Karpathy.
\newblock rustbpe: The missing tiktoken training code, 2026.
\newblock URL \url{https://github.com/karpathy/rustbpe}.

\bibitem[Keskar et~al.(2019)Keskar, McCann, Varshney, Xiong, and
  Socher]{keskar2019ctrl}
Nitish~Shirish Keskar, Bryan McCann, Lav~R Varshney, Caiming Xiong, and Richard
  Socher.
\newblock Ctrl: A conditional transformer language model for controllable
  generation.
\newblock \emph{arXiv preprint arXiv:1909.05858}, 2019.

\bibitem[Kitts et~al.(2026)Kitts, Larsen, and Von~Arx]{rubygems}
Spencer Kitts, Thomas Larsen, and Sydney Von~Arx.
\newblock {OpenAI} agents carried out an undisclosed cyber-attack on
  {RubyGems}.
\newblock rubyhack.ai, September 2026.
\newblock URL \url{https://www.rubyhack.ai/}.
\newblock Accessed 2026-09-16.

\bibitem[Korbak et~al.(2023)Korbak, Shi, Chen, Bhalerao, Buckley, Phang,
  Bowman, and Perez]{korbak2023pretraininglanguagemodelshuman}
Tomasz Korbak, Kejian Shi, Angelica Chen, Rasika Bhalerao, Christopher~L.
  Buckley, Jason Phang, Samuel~R. Bowman, and Ethan Perez.
\newblock Pretraining language models with human preferences.
\newblock In \emph{Proceedings of the 40th International Conference on Machine
  Learning}, ICML'23. JMLR.org, 2023.

\bibitem[Lambert(2026)]{lambert2026reinforcement}
Nathan Lambert.
\newblock \emph{Reinforcement Learning from Human Feedback: Alignment and
  post-training of {LLMs}}.
\newblock Manning Publications, 2026.
\newblock ISBN 9781633434301.
\newblock URL
  \url{https://www.manning.com/books/reinforcement-learning-from-human-feedback}.

\bibitem[Lester et~al.(2021)Lester, Al-Rfou, and Constant]{lester2021power}
Brian Lester, Rami Al-Rfou, and Noah Constant.
\newblock The power of scale for parameter-efficient prompt tuning.
\newblock In \emph{Proceedings of the 2021 conference on empirical methods in
  natural language processing}, pp.\  3045--3059, 2021.

\bibitem[Li et~al.(2026)Li, Wichers, Price, Marks, and
  Kutasov]{li2026modelspecmidtrainingimproving}
Chloe Li, Nevan Wichers, Sara Price, Samuel Marks, and Jon Kutasov.
\newblock Model spec midtraining: Improving how alignment training generalizes,
  2026.
\newblock URL \url{https://arxiv.org/abs/2605.02087}.

\bibitem[Li et~al.(2024)Li, Fang, Smyrnis, Ivgi, Jordan, Gadre, Bansal, Guha,
  Keh, Arora, Garg, Xin, Muennighoff, Heckel, Mercat, Chen, Gururangan,
  Wortsman, Albalak, Bitton, Nezhurina, Abbas, Hsieh, Ghosh, Gardner, Kilian,
  Zhang, Shao, Pratt, Sanyal, Ilharco, Daras, Marathe, Gokaslan, Zhang, Chandu,
  Nguyen, Vasiljevic, Kakade, Song, Sanghavi, Faghri, Oh, Zettlemoyer, Lo,
  El-Nouby, Pouransari, Toshev, Wang, Groeneveld, Soldaini, Koh, Jitsev,
  Kollar, Dimakis, Carmon, Dave, Schmidt, and Shankar]{li2024datacomp}
Jeffrey Li, Alex Fang, Georgios Smyrnis, Maor Ivgi, Matt Jordan, Samir Gadre,
  Hritik Bansal, Etash Guha, Sedrick Keh, Kushal Arora, Saurabh Garg, Rui Xin,
  Niklas Muennighoff, Reinhard Heckel, Jean Mercat, Mayee Chen, Suchin
  Gururangan, Mitchell Wortsman, Alon Albalak, Yonatan Bitton, Marianna
  Nezhurina, Amro Abbas, Cheng-Yu Hsieh, Dhruba Ghosh, Josh Gardner, Maciej
  Kilian, Hanlin Zhang, Rulin Shao, Sarah Pratt, Sunny Sanyal, Gabriel Ilharco,
  Giannis Daras, Kalyani Marathe, Aaron Gokaslan, Jieyu Zhang, Khyathi Chandu,
  Thao Nguyen, Igor Vasiljevic, Sham Kakade, Shuran Song, Sujay Sanghavi,
  Fartash Faghri, Sewoong Oh, Luke Zettlemoyer, Kyle Lo, Alaaeldin El-Nouby,
  Hadi Pouransari, Alexander Toshev, Stephanie Wang, Dirk Groeneveld, Luca
  Soldaini, Pang~Wei Koh, Jenia Jitsev, Thomas Kollar, Alexandros~G. Dimakis,
  Yair Carmon, Achal Dave, Ludwig Schmidt, and Vaishaal Shankar.
\newblock {DataComp-LM}: In search of the next generation of training sets for
  language models.
\newblock In \emph{Advances in Neural Information Processing Systems},
  volume~37, pp.\  14200--14282, 2024.

\bibitem[Lin(2023)]{neuronpedia}
Johnny Lin.
\newblock Neuronpedia: Interactive reference and tooling for analyzing neural
  networks, 2023.
\newblock URL \url{https://www.neuronpedia.org}.
\newblock Software available from neuronpedia.org.

\bibitem[Lindsey et~al.(2025)Lindsey, Gurnee, Ameisen, Chen, Pearce, Turner,
  Citro, Abrahams, Carter, Hosmer, Marcus, Sklar, Templeton, Bricken,
  McDougall, Cunningham, Henighan, Jermyn, Jones, Persic, Qi, Thompson,
  Zimmerman, Rivoire, Conerly, Olah, and Batson]{lindsey2025biology}
Jack Lindsey, Wes Gurnee, Emmanuel Ameisen, Brian Chen, Adam Pearce,
  Nicholas~L. Turner, Craig Citro, David Abrahams, Shan Carter, Basil Hosmer,
  Jonathan Marcus, Michael Sklar, Adly Templeton, Trenton Bricken, Callum
  McDougall, Hoagy Cunningham, Thomas Henighan, Adam Jermyn, Andy Jones, Andrew
  Persic, Zhenyi Qi, T.~Ben Thompson, Sam Zimmerman, Kelley Rivoire, Thomas
  Conerly, Chris Olah, and Joshua Batson.
\newblock On the biology of a large language model.
\newblock \emph{Transformer Circuits Thread}, 2025.
\newblock URL
  \url{https://transformer-circuits.pub/2025/attribution-graphs/biology.html}.

\bibitem[Lixing(2026)]{lixing2026calyrex}
Li~Lixing.
\newblock System-prompt anchoring with cross-attention layers.
\newblock \emph{arXiv preprint arXiv:2605.09737}, 2026.
\newblock \doi{10.48550/arXiv.2605.09737}.
\newblock URL \url{https://arxiv.org/abs/2605.09737v2}.

\bibitem[Lu et~al.(2026)Lu, Gallagher, Michala, Fish, and
  Lindsey]{lu2026assistantaxissituatingstabilizing}
Christina Lu, Jack Gallagher, Jonathan Michala, Kyle Fish, and Jack Lindsey.
\newblock The assistant axis: Situating and stabilizing the default persona of
  language models.
\newblock In \emph{Proceedings of the 43rd International Conference on Machine
  Learning}, 2026.
\newblock URL \url{https://openreview.net/forum?id=q17jVxMtwS}.

\bibitem[Maini et~al.(2025)Maini, Goyal, Sam, Robey, Savani, Jiang, Zou,
  Fredrikson, Lipton, and Kolter]{maini2026safety}
Pratyush Maini, Sachin Goyal, Dylan Sam, Alexander Robey, Yash Savani, Yiding
  Jiang, Andy Zou, Matt Fredrikson, Zachary Lipton, and Zico Kolter.
\newblock Safety pretraining: Toward the next generation of safe ai.
\newblock In \emph{Advances in Neural Information Processing Systems},
  volume~38, pp.\  43971--44008, 2025.

\bibitem[Marks \& Tegmark(2024)Marks and Tegmark]{marks2023geometry}
Samuel Marks and Max Tegmark.
\newblock The geometry of truth: Emergent linear structure in large language
  model representations of true/false datasets.
\newblock In \emph{Conference on Language Modeling}, 2024.
\newblock URL \url{https://openreview.net/forum?id=aajyHYjjsk}.

\bibitem[Mazeika et~al.(2024)Mazeika, Phan, Yin, Zou, Wang, Mu, Sakhaee, Li,
  Basart, Li, Forsyth, and Hendrycks]{pmlr-v235-mazeika24a}
Mantas Mazeika, Long Phan, Xuwang Yin, Andy Zou, Zifan Wang, Norman Mu, Elham
  Sakhaee, Nathaniel Li, Steven Basart, Bo~Li, David Forsyth, and Dan
  Hendrycks.
\newblock {H}arm{B}ench: A standardized evaluation framework for automated red
  teaming and robust refusal.
\newblock In Ruslan Salakhutdinov, Zico Kolter, Katherine Heller, Adrian
  Weller, Nuria Oliver, Jonathan Scarlett, and Felix Berkenkamp (eds.),
  \emph{Proceedings of the 41st International Conference on Machine Learning},
  volume 235 of \emph{Proceedings of Machine Learning Research}, pp.\
  35181--35224. PMLR, 21--27 Jul 2024.
\newblock URL \url{https://proceedings.mlr.press/v235/mazeika24a.html}.

\bibitem[McDougall et~al.(2025)McDougall, Conmy, Kram{\'a}r, Lieberum,
  Rajamanoharan, and Nanda]{McDougallGemmaS2}
Callum McDougall, Arthur Conmy, J{\'a}nos Kram{\'a}r, Tom Lieberum, Senthooran
  Rajamanoharan, and Neel Nanda.
\newblock Gemma scope 2 -- technical paper.
\newblock Technical report, Google, 2025.
\newblock URL
  \url{https://storage.googleapis.com/deepmind-media/DeepMind.com/Blog/gemma-scope-2-helping-the-ai-safety-community-deepen-understanding-of-complex-language-model-behavior/Gemma_Scope_2_Technical_Paper.pdf}.

\bibitem[Minder et~al.(2026)Minder, Moskvoretskii, Singhal, Jiao, Arditi, Cui,
  Borjigin, Bali, Krsteski, Raj, Nguyen, Brinkmann, Anderson, Aydin, and
  West]{minder2026syntheticpersonapretrainingalignment}
Julian Minder, Viktor Moskvoretskii, Raghav Singhal, Difan Jiao, Andy Arditi,
  Shaobo Cui, Yiderigun Borjigin, Kartik Bali, Stefan Krsteski, Harsh Raj, Huu
  Nguyen, Jannik Brinkmann, Ashton Anderson, Roland Aydin, and Robert West.
\newblock Synthetic persona pretraining: Alignment from token zero, 2026.
\newblock URL \url{https://arxiv.org/abs/2608.13482}.

\bibitem[Moskvoretskii et~al.(2026)Moskvoretskii, Glandorf, Moreira, K{\"a}ser,
  and West]{moskvoretskii2026tracing}
Viktor Moskvoretskii, Dominik Glandorf, Jorge~Medina Moreira, Tanja K{\"a}ser,
  and Robert West.
\newblock Tracing persona vectors through llm pretraining.
\newblock \emph{arXiv preprint arXiv:2605.13329}, 2026.

\bibitem[O'Brien et~al.(2026)O'Brien, Casper, Anthony, Korbak, Kirk, Davies,
  Mishra, Irving, Gal, and Biderman]{o2026deep}
Kyle O'Brien, Stephen Casper, Quentin Anthony, Tomek Korbak, Robert Kirk,
  Xander Davies, Ishan Mishra, Geoffrey Irving, Yarin Gal, and Stella~R
  Biderman.
\newblock Deep ignorance: Filtering pretraining data builds tamper-resistant
  safeguards into open-weight llms.
\newblock In \emph{International Conference on Learning Representations},
  volume 2026, pp.\  35579--35633, 2026.

\bibitem[{OpenAI}(2026)]{hf_openai_technical}
{OpenAI}.
\newblock {OpenAI} -- {Hugging Face} incident technical report.
\newblock Technical report, OpenAI, August 2026.
\newblock URL
  \url{https://cdn.openai.com/pdf/67869394-cb91-4c12-888c-5cbd85c7814c/OpenAI-Hugging-Face%20Incident-Technical-Report.pdf}.
\newblock Accessed 2026-09-16.

\bibitem[Ouyang et~al.(2022)Ouyang, Wu, Jiang, Almeida, Wainwright, Mishkin,
  Zhang, Agarwal, Slama, Ray, et~al.]{ouyang2022training}
Long Ouyang, Jeffrey Wu, Xu~Jiang, Diogo Almeida, Carroll Wainwright, Pamela
  Mishkin, Chong Zhang, Sandhini Agarwal, Katarina Slama, Alex Ray, et~al.
\newblock Training language models to follow instructions with human feedback.
\newblock In \emph{Advances in Neural Information Processing Systems},
  volume~35, pp.\  27730--27744, 2022.

\bibitem[Penedo et~al.(2024)Penedo, Kydl\'{\i}\v{c}ek, allal, Lozhkov,
  Mitchell, Raffel, Von~Werra, and Wolf]{penedo2024finewebdatasetsdecantingweb}
Guilherme Penedo, Hynek Kydl\'{\i}\v{c}ek, Loubna~Ben allal, Anton Lozhkov,
  Margaret Mitchell, Colin Raffel, Leandro Von~Werra, and Thomas Wolf.
\newblock The {FineWeb} datasets: Decanting the web for the finest text data at
  scale.
\newblock In \emph{Advances in Neural Information Processing Systems},
  volume~37, pp.\  30811--30849. Curran Associates, Inc., 2024.
\newblock \doi{10.52202/079017-0970}.
\newblock URL
  \url{https://proceedings.nips.cc/paper_files/paper/2024/hash/370df50ccfdf8bde18f8f9c2d9151bda-Abstract-Datasets_and_Benchmarks_Track.html}.
\newblock Datasets and Benchmarks Track.

\bibitem[Perez et~al.(2022)Perez, Huang, Song, Cai, Ring, Aslanides, Glaese,
  McAleese, and Irving]{perez2022red}
Ethan Perez, Saffron Huang, Francis Song, Trevor Cai, Roman Ring, John
  Aslanides, Amelia Glaese, Nat McAleese, and Geoffrey Irving.
\newblock Red teaming language models with language models.
\newblock In \emph{Proceedings of the 2022 conference on empirical methods in
  natural language processing}, pp.\  3419--3448, 2022.

\bibitem[Qi et~al.(2024)Qi, Zeng, Xie, Chen, Jia, Mittal, and
  Henderson]{qi2024fine}
Xiangyu Qi, Yi~Zeng, Tinghao Xie, Pin-Yu Chen, Ruoxi Jia, Prateek Mittal, and
  Peter Henderson.
\newblock Fine-tuning aligned language models compromises safety, even when
  users do not intend to!
\newblock In \emph{International Conference on Learning Representations},
  volume 2024, pp.\  30988--31043, 2024.

\bibitem[Qi et~al.(2025)Qi, Panda, Lyu, Ma, Roy, Beirami, Mittal, and
  Henderson]{ICLR2025_88be0230}
Xiangyu Qi, Ashwinee Panda, Kaifeng Lyu, Xiao Ma, Subhrajit Roy, Ahmad Beirami,
  Prateek Mittal, and Peter Henderson.
\newblock Safety alignment should be made more than just a few tokens deep.
\newblock In Y.~Yue, A.~Garg, N.~Peng, F.~Sha, and R.~Yu (eds.),
  \emph{International Conference on Learning Representations}, volume 2025,
  pp.\  54911--54941, 2025.
\newblock URL
  \url{https://proceedings.iclr.cc/paper_files/paper/2025/file/88be023075a5a3ff3dc3b5d26623fa22-Paper-Conference.pdf}.

\bibitem[Qwen et~al.(2025)Qwen, Yang, Yang, Zhang, Hui, Zheng, Yu, Li, Liu,
  Huang, Wei, Lin, Yang, Tu, Zhang, Yang, Yang, Zhou, Lin, Dang, Lu, Bao, Yang,
  Yu, Li, Xue, Zhang, Zhu, Men, Lin, Li, Tang, Xia, Ren, Ren, Fan, Su, Zhang,
  Wan, Liu, Cui, Zhang, and Qiu]{qwen2025qwen25technicalreport}
Qwen, An~Yang, Baosong Yang, Beichen Zhang, Binyuan Hui, Bo~Zheng, Bowen Yu,
  Chengyuan Li, Dayiheng Liu, Fei Huang, Haoran Wei, Huan Lin, Jian Yang,
  Jianhong Tu, Jianwei Zhang, Jianxin Yang, Jiaxi Yang, Jingren Zhou, Junyang
  Lin, Kai Dang, Keming Lu, Keqin Bao, Kexin Yang, Le~Yu, Mei Li, Mingfeng Xue,
  Pei Zhang, Qin Zhu, Rui Men, Runji Lin, Tianhao Li, Tianyi Tang, Tingyu Xia,
  Xingzhang Ren, Xuancheng Ren, Yang Fan, Yang Su, Yichang Zhang, Yu~Wan,
  Yuqiong Liu, Zeyu Cui, Zhenru Zhang, and Zihan Qiu.
\newblock Qwen2.5 technical report, 2025.
\newblock URL \url{https://arxiv.org/abs/2412.15115}.

\bibitem[Rafailov et~al.(2023)Rafailov, Sharma, Mitchell, Manning, Ermon, and
  Finn]{rafailov2023direct}
Rafael Rafailov, Archit Sharma, Eric Mitchell, Christopher~D Manning, Stefano
  Ermon, and Chelsea Finn.
\newblock Direct preference optimization: Your language model is secretly a
  reward model.
\newblock In \emph{Advances in Neural Information Processing Systems},
  volume~36, pp.\  53728--53741, 2023.

\bibitem[Rimsky et~al.(2024)Rimsky, Gabrieli, Schulz, Tong, Hubinger, and
  Turner]{rimsky2024steering}
Nina Rimsky, Nick Gabrieli, Julian Schulz, Meg Tong, Evan Hubinger, and
  Alexander Turner.
\newblock Steering llama 2 via contrastive activation addition.
\newblock In \emph{Proceedings of the 62nd Annual Meeting of the Association
  for Computational Linguistics (Volume 1: Long Papers)}, pp.\  15504--15522,
  2024.

\bibitem[Russinovich et~al.(2025)Russinovich, Salem, and
  Eldan]{russinovich2025greatwritearticlethat}
Mark Russinovich, Ahmed Salem, and Ronen Eldan.
\newblock Great, now write an article about that: The crescendo multi-turn
  {LLM} jailbreak attack.
\newblock In \emph{34th USENIX Security Symposium (USENIX Security 25)}, pp.\
  2421--2440. USENIX Association, 2025.
\newblock URL
  \url{https://www.usenix.org/conference/usenixsecurity25/presentation/russinovich}.

\bibitem[Sam et~al.(2026)Sam, Goyal, Maini, Robey, and Kolter]{sam2026should}
Dylan Sam, Sachin Goyal, Pratyush Maini, Alexander Robey, and J~Zico Kolter.
\newblock When should we introduce safety interventions during pretraining?
\newblock \emph{arXiv preprint arXiv:2601.07087}, 2026.

\bibitem[Schwinn et~al.(2024)Schwinn, Dobre, Xhonneux, Gidel, and
  G{\"u}nnemann]{schwinn2024soft}
Leo Schwinn, David Dobre, Sophie Xhonneux, Gauthier Gidel, and Stephan
  G{\"u}nnemann.
\newblock Soft prompt threats: Attacking safety alignment and unlearning in
  open-source llms through the embedding space.
\newblock \emph{Advances in Neural Information Processing Systems},
  37:\penalty0 9086--9116, 2024.

\bibitem[Shah et~al.(2026)Shah, Wu, Saxena, Zhong, Robey, and
  Raghunathan]{shah2026jailbreakinghaystack}
Rishi~Rajesh Shah, Chen~Henry Wu, Shashwat Saxena, Ziqian Zhong, Alexander
  Robey, and Aditi Raghunathan.
\newblock Jailbreaking in the haystack, 2026.
\newblock URL \url{https://arxiv.org/abs/2511.04707}.

\bibitem[Shao et~al.(2024)Shao, Wang, Zhu, Xu, Song, Bi, Zhang, Zhang, Li, Wu,
  et~al.]{shao2024deepseekmath}
Zhihong Shao, Peiyi Wang, Qihao Zhu, Runxin Xu, Junxiao Song, Xiao Bi, Haowei
  Zhang, Mingchuan Zhang, YK~Li, Yang Wu, et~al.
\newblock Deepseekmath: Pushing the limits of mathematical reasoning in open
  language models.
\newblock \emph{arXiv preprint arXiv:2402.03300}, 2024.

\bibitem[Sharma et~al.(2025)Sharma, Tong, Mu, Wei, Kruthoff, Goodfriend, Ong,
  Peng, Agarwal, Anil, Askell, Bailey, Benton, Bluemke, Bowman, Christiansen,
  Cunningham, Dau, Gopal, Gilson, Graham, Howard, Kalra, Lee, Lin, Lofgren,
  Mosconi, O'Hara, Olsson, Petrini, Rajani, Saxena, Silverstein, Singh, Sumers,
  Tang, Troy, Weisser, Zhong, Zhou, Leike, Kaplan, and
  Perez]{sharma2025constitutionalclassifiersdefendinguniversal}
Mrinank Sharma, Meg Tong, Jesse Mu, Jerry Wei, Jorrit Kruthoff, Scott
  Goodfriend, Euan Ong, Alwin Peng, Raj Agarwal, Cem Anil, Amanda Askell,
  Nathan Bailey, Joe Benton, Emma Bluemke, Samuel~R. Bowman, Eric Christiansen,
  Hoagy Cunningham, Andy Dau, Anjali Gopal, Rob Gilson, Logan Graham, Logan
  Howard, Nimit Kalra, Taesung Lee, Kevin Lin, Peter Lofgren, Francesco
  Mosconi, Clare O'Hara, Catherine Olsson, Linda Petrini, Samir Rajani, Nikhil
  Saxena, Alex Silverstein, Tanya Singh, Theodore Sumers, Leonard Tang,
  Kevin~K. Troy, Constantin Weisser, Ruiqi Zhong, Giulio Zhou, Jan Leike, Jared
  Kaplan, and Ethan Perez.
\newblock Constitutional classifiers: Defending against universal jailbreaks
  across thousands of hours of red teaming, 2025.
\newblock URL \url{https://arxiv.org/abs/2501.18837}.

\bibitem[Simionescu et~al.(2022)Simionescu, Hoeltgebaum, and
  Martins]{simionescu2022steering}
P.~A. Simionescu, Thiago Hoeltgebaum, and Daniel Martins.
\newblock On the evolution of automotive steering mechanisms.
\newblock In Pierre Larochelle and J.~Michael McCarthy (eds.),
  \emph{Proceedings of the 2022 USCToMM Symposium on Mechanical Systems and
  Robotics}, volume 118 of \emph{Mechanisms and Machine Science}, pp.\
  116--128, Cham, 2022. Springer.
\newblock \doi{10.1007/978-3-030-99826-4_11}.
\newblock URL \url{https://doi.org/10.1007/978-3-030-99826-4_11}.

\bibitem[So et~al.(2021)So, Ma{\'n}ke, Liu, Dai, Shazeer, and Le]{so2021primer}
David~R. So, Wojciech Ma{\'n}ke, Hanxiao Liu, Zihang Dai, Noam Shazeer, and
  Quoc~V. Le.
\newblock Primer: Searching for efficient transformers for language modeling.
\newblock In \emph{Advances in Neural Information Processing Systems},
  volume~34, 2021.
\newblock URL \url{https://arxiv.org/abs/2109.08668}.

\bibitem[Struppek et~al.(2026)Struppek, Gleave, and
  Pelrine]{struppek2026exposingsystematicvulnerabilityopenweight}
Lukas Struppek, Adam Gleave, and Kellin Pelrine.
\newblock Exposing the systematic vulnerability of open-weight models to
  prefill attacks, 2026.
\newblock URL \url{https://arxiv.org/abs/2602.14689}.

\bibitem[Tamirisa et~al.(2025)Tamirisa, Bharathi, Phan, Zhou, Gatti, Suresh,
  Lin, Wang, Wang, Arel, et~al.]{tamirisa2025tamper}
Rishub Tamirisa, Bhrugu Bharathi, Long Phan, Andy Zhou, Alice Gatti, Tarun
  Suresh, Maxwell Lin, Justin Wang, Rowan Wang, Ron Arel, et~al.
\newblock Tamper-resistant safeguards for open-weight llms.
\newblock In \emph{International Conference on Learning Representations},
  volume 2025, pp.\  101802--101829, 2025.

\bibitem[Tie et~al.(2025)Tie, Zhao, Song, Wei, Zhou, Dai, Yin, Yang, Yan, Su,
  et~al.]{tie2025survey}
Guiyao Tie, Zeli Zhao, Dingjie Song, Fuyang Wei, Rong Zhou, Yurou Dai, Wen Yin,
  Zhejian Yang, Jiangyue Yan, Yao Su, et~al.
\newblock A survey on post-training of large language models.
\newblock \emph{arXiv preprint arXiv:2503.06072}, 2025.

\bibitem[Turner et~al.(2023)Turner, Thiergart, Leech, Udell, Vazquez, Mini, and
  MacDiarmid]{turner2023steering}
Alexander~Matt Turner, Lisa Thiergart, Gavin Leech, David Udell, Juan~J
  Vazquez, Ulisse Mini, and Monte MacDiarmid.
\newblock Steering language models with activation engineering.
\newblock \emph{arXiv preprint arXiv:2308.10248}, 2023.

\bibitem[Vaswani et~al.(2017)Vaswani, Shazeer, Parmar, Uszkoreit, Jones, Gomez,
  Kaiser, and Polosukhin]{vaswani2017attention}
Ashish Vaswani, Noam Shazeer, Niki Parmar, Jakob Uszkoreit, Llion Jones,
  Aidan~N Gomez, {\L}ukasz Kaiser, and Illia Polosukhin.
\newblock Attention is all you need.
\newblock In \emph{Advances in Neural Information Processing Systems},
  volume~30, 2017.

\bibitem[Von~Arx et~al.(2026)Von~Arx, Byrd, Kitts, and Larsen]{collusion_wiki}
Sydney Von~Arx, Cormac~Slade Byrd, Spencer Kitts, and Thomas Larsen.
\newblock Discovery of a new {OpenAI} agent message board.
\newblock Nightingale Collective, September 2026.
\newblock URL \url{https://collusion.wiki/}.
\newblock Accessed 2026-09-16.

\bibitem[Wallace et~al.(2024)Wallace, Xiao, Leike, Weng, Heidecke, and
  Beutel]{wallace2024instruction}
Eric Wallace, Kai Xiao, Reimar Leike, Lilian Weng, Johannes Heidecke, and Alex
  Beutel.
\newblock The instruction hierarchy: Training llms to prioritize privileged
  instructions.
\newblock \emph{arXiv preprint arXiv:2404.13208}, 2024.

\bibitem[Wang et~al.(2026)Wang, Wang, Li, Li, and Sha]{wang-etal-2026-red}
Hao Wang, Yanting Wang, Hao Li, Rui Li, and Lei Sha.
\newblock Be your own red teamer: Safety alignment via self-play and reflective
  experience replay.
\newblock In Maria Liakata, Viviane~P. Moreira, Jiajun Zhang, and David Jurgens
  (eds.), \emph{Findings of the {A}ssociation for {C}omputational
  {L}inguistics: {ACL} 2026}, pp.\  18700--18716, San Diego, California, United
  States, July 2026. Association for Computational Linguistics.
\newblock ISBN 979-8-89176-395-1.
\newblock \doi{10.18653/v1/2026.findings-acl.933}.
\newblock URL \url{https://aclanthology.org/2026.findings-acl.933/}.

\bibitem[Wang et~al.(2025)Wang, Griffin, Treutlein, Perez, Michael, Roger, and
  Marks]{wang2025modifying}
Rowan Wang, Avery Griffin, Johannes Treutlein, Ethan Perez, Julian Michael,
  Fabien Roger, and Sam Marks.
\newblock Modifying {LLM} beliefs with synthetic document finetuning.
\newblock Anthropic Alignment Science Blog, April 2025.
\newblock URL
  \url{https://alignment.anthropic.com/2025/modifying-beliefs-via-sdf/}.

\bibitem[Wu et~al.(2025)Wu, Arora, Geiger, Wang, Huang, Jurafsky, Manning, and
  Potts]{wu2025axbenchsteeringllmssimple}
Zhengxuan Wu, Aryaman Arora, Atticus Geiger, Zheng Wang, Jing Huang, Dan
  Jurafsky, Christopher~D Manning, and Christopher Potts.
\newblock {A}x{B}ench: Steering {LLM}s? {E}ven simple baselines outperform
  sparse autoencoders.
\newblock In \emph{Proceedings of the 42nd International Conference on Machine
  Learning}, volume 267 of \emph{Proceedings of Machine Learning Research},
  pp.\  67035--67080. PMLR, 2025.
\newblock URL \url{https://proceedings.mlr.press/v267/wu25a.html}.

\bibitem[Yang et~al.(2025)Yang, Li, Yang, Zhang, Hui, Zheng, Yu, Gao, Huang,
  Lv, Zheng, Liu, Zhou, Huang, Hu, Ge, Wei, Lin, Tang, Yang, Tu, Zhang, Yang,
  Yang, Zhou, Zhou, Lin, Dang, Bao, Yang, Yu, Deng, Li, Xue, Li, Zhang, Wang,
  Zhu, Men, Gao, Liu, Luo, Li, Tang, Yin, Ren, Wang, Zhang, Ren, Fan, Su,
  Zhang, Zhang, Wan, Liu, Wang, Cui, Zhang, Zhou, and
  Qiu]{yang2025qwen3technicalreport}
An~Yang, Anfeng Li, Baosong Yang, Beichen Zhang, Binyuan Hui, Bo~Zheng, Bowen
  Yu, Chang Gao, Chengen Huang, Chenxu Lv, Chujie Zheng, Dayiheng Liu, Fan
  Zhou, Fei Huang, Feng Hu, Hao Ge, Haoran Wei, Huan Lin, Jialong Tang, Jian
  Yang, Jianhong Tu, Jianwei Zhang, Jianxin Yang, Jiaxi Yang, Jing Zhou,
  Jingren Zhou, Junyang Lin, Kai Dang, Keqin Bao, Kexin Yang, Le~Yu, Lianghao
  Deng, Mei Li, Mingfeng Xue, Mingze Li, Pei Zhang, Peng Wang, Qin Zhu, Rui
  Men, Ruize Gao, Shixuan Liu, Shuang Luo, Tianhao Li, Tianyi Tang, Wenbiao
  Yin, Xingzhang Ren, Xinyu Wang, Xinyu Zhang, Xuancheng Ren, Yang Fan, Yang
  Su, Yichang Zhang, Yinger Zhang, Yu~Wan, Yuqiong Liu, Zekun Wang, Zeyu Cui,
  Zhenru Zhang, Zhipeng Zhou, and Zihan Qiu.
\newblock Qwen3 technical report, 2025.
\newblock URL \url{https://arxiv.org/abs/2505.09388}.

\bibitem[Yu et~al.(2025)Yu, Do, Hambardzumyan, and Cancedda]{ICLR2025_1022661f}
Lei Yu, Virginie Do, Karen Hambardzumyan, and Nicola Cancedda.
\newblock Robust llm safeguarding via refusal feature adversarial training.
\newblock In Y.~Yue, A.~Garg, N.~Peng, F.~Sha, and R.~Yu (eds.),
  \emph{International Conference on Learning Representations}, volume 2025,
  pp.\  5254--5277, 2025.
\newblock URL
  \url{https://proceedings.iclr.cc/paper_files/paper/2025/file/1022661f3f43406065641f16ce25eafa-Paper-Conference.pdf}.

\bibitem[Zhang \& Sennrich(2019)Zhang and Sennrich]{zhang2019rmsnorm}
Biao Zhang and Rico Sennrich.
\newblock Root mean square layer normalization.
\newblock In \emph{Advances in Neural Information Processing Systems},
  volume~32, 2019.
\newblock URL \url{https://arxiv.org/abs/1910.07467}.

\bibitem[Zhou et~al.(2025)Zhou, Wu, Jiang, Obeid, and Lan]{zhou2025value}
Zhanchao Zhou, Tianyi Wu, Zhiyun Jiang, Fares Obeid, and Zhenzhong Lan.
\newblock Value residual learning.
\newblock In \emph{Proceedings of the 63rd Annual Meeting of the Association
  for Computational Linguistics (Volume 1: Long Papers)}, pp.\  28341--28356,
  2025.

\bibitem[Zou et~al.(2023)Zou, Wang, Carlini, Nasr, Kolter, and
  Fredrikson]{zou2023universal}
Andy Zou, Zifan Wang, Nicholas Carlini, Milad Nasr, J~Zico Kolter, and Matt
  Fredrikson.
\newblock Universal and transferable adversarial attacks on aligned language
  models.
\newblock \emph{arXiv preprint arXiv:2307.15043}, 2023.

\bibitem[Zverev et~al.(2026)Zverev, Kortukov, Panfilov, Volkova, Tabesh,
  Lapuschkin, Samek, and Lampert]{zverev2026aside}
Egor Zverev, Evgenii Kortukov, Alexander Panfilov, Alexandra Volkova, Soroush
  Tabesh, Sebastian Lapuschkin, Wojciech Samek, and Christoph Lampert.
\newblock Aside: Architectural separation of instructions and data in language
  models.
\newblock In \emph{International Conference on Learning Representations},
  volume 2026, pp.\  112869--112898, 2026.

\end{thebibliography}
\bibliographystyle{iclr2027_conference}

\newpage

\appendix
\crefalias{section}{appendix}
\crefalias{subsection}{appendix}
\crefalias{subsubsection}{appendix}
\crefname{appendix}{Appendix}{Appendices}
\Crefname{appendix}{Appendix}{Appendices}

\section*{Supplementary material to the paper
\emph{Steerable by Design: Conditional Pretraining for Robust Control of Language Models}}
\renewcommand{\thesection}{\Alph{section}}

We organize the appendix as follows.

\textbf{\ref{app:training} \nameref*{app:training}.}
Details of the pretraining recipe, including data segmentation, the splicing procedure, SAE feature extraction, feature filtering, clustering and selection, positive/negative feature sampling, the encoder--decoder architecture, and the full hyperparameter table.

\textbf{\ref{app:evaluation} \nameref*{app:evaluation}.}
The evaluation protocol shared across all results, including the concepts and instructions, the LM-judge rubric, the steering-strength grids, and the rule for choosing the operating point under attack, baseline fitting and steering layers, the natural-language steering interfaces, and how the open models were prepared.

\textbf{\ref{app:clean3b} \nameref*{app:clean3b}.}
The full clean steering evaluation at 3B, including the best operating point of every method on every model, the complete strength sweeps in tabular form, and the trade-off and factor-sweep curves for all three open models.

\textbf{\ref{app:gepa} \nameref*{app:gepa}.}
The GEPA suffix attack in detail, including per-method concept scores at every budget, the same results in judge units and as retention for all four of our interfaces, fluency under attack, and example optimized suffixes.

\textbf{\ref{app:rl-redteam} \nameref*{app:rl-redteam}.}
The RL attacker in detail, including how it is trained, the judge that rewards it, how steered and broken concepts are counted, per-method results, and the attack's side effects.

\textbf{\ref{app:softprompt} \nameref*{app:softprompt}.}
The two soft-prompt attacks in detail, including how the soft tokens are parameterized and optimized, per-method results, and why \oursbig{}'s value embeddings are derived from the free input embedding rather than freed as well.

\textbf{\ref{app:counter-3b} \nameref*{app:counter-3b}.}
Activation-space counter-steering at 3B, which subtracts the concept's \diffmean{} or \linearprobe{} direction from the residual stream, including how attack strength is put on a shared relative axis across models, the sweeps along both directions, and per-method results at matched fluency.

\textbf{\ref{app:capability} \nameref*{app:capability}.} General knowledge and reasoning abilities, including ChatCORE and per-benchmark scores for both model scales and their decoder-only baselines, and the effect of supplying ground-truth conditioning signals.

\textbf{\ref{app:m400} \nameref*{app:m400}.}
The controlled study at 400M in full, including steering against the architecture-matched ($286$M) decoder-only control, the four input-channel attacks and latent counter-steering against both controls, and best operating points for both controls.

\textbf{\ref{sec:theory} \nameref*{sec:theory}.}
A formal account of conditional pretraining, including the decomposition of the conditional next-token loss and the reduction in the loss floor provided by the conditioning signal (Proposition~\ref{prop:floor}).

\textbf{\ref{app:splicing-ablation} \nameref*{app:splicing-ablation}.}
An ablation study where we pretrain a 400M-scale model without dataset splicing.

\textbf{\ref{app:negative-steering} \nameref*{app:negative-steering}.}
A variant of AxBench where we test "negative steering," the ability of the model to steer \textit{away} from a target concept rather than towards it.

\textbf{\ref{app:demos} \nameref*{app:demos}.}
Qualitative steered generations for three fixed concepts and prompts, across every steering method on \oursbig{}, \ourssmall{}, the decoder-only baselines, and the open models, at low and high steering strength.

\section{Training and Model Details}
\label{app:training}

The codebase is a heavily modified version of Nanochat \citep{nanochat}, so many of the more basic design choices (dataset, model architecture, hyperparameters, etc.) are downstream of what was in the original codebase.

\subsection{Data Preparation and Segmentation}
\label{app:segmentation}

We utilize FineWeb-Edu \citep{penedo2024finewebdatasetsdecantingweb} as our dataset, processing 53.82B and 9.8B subsets to pretrain \oursbig{} and \ourssmall{}, respectively.

We train a custom byte-pair encoding tokenizer using RustBPE \citep{karpathy2026rustbpe} with 2B characters from FineWeb-Edu to get a vocabulary of 32,768 tokens, including 9 reserved tokens. This tokenizer is reused between the two models.

However, we do not start by tokenizing the dataset using this tokenizer; instead, we start by using \textit{Gemma's} tokenizer. Since we need to run these segments through \texttt{gemma-3-4b-pt} later to extract SAE feature activations, we first tokenize the documents and split them into fixed-length token segments in Gemma's token space. We use a segment length of 128 for this, which roughly translates to 256-length segments for our tokenizer. We leave short, tail-end segments when the document's token length is not divisible by our segment length, so a document of length 280 would be split as 128 + 128 + 24.

\subsection{Splicing Procedure}
\label{app:splicing}

We splice $p=0.15$ of segment boundaries. This yields 16.6M out of 110.9M boundaries for our smaller dataset and 63.1M out of 420.3M for our larger one. We randomly select this many segment boundaries without replacement and shuffle the continuations among these cuts. If a newly spliced document runs over our context length limit (2048 for \oursbig{} and 1024 for \ourssmall{}), we cut the trailing segments and reintroduce them as entirely new documents in the dataset.

\subsection{SAE Feature Extraction}
\label{app:sae}

Note that this step logically comes after splicing (subsection ~\ref{app:splicing}) but in our actual implementation the order is reversed.

We consider each segment as a separate document for SAE feature extraction. For segments that do not come first in the document, we add a special beginning-of-sequence token \texttt{<BOS>}. Then, we prefill the whole sequence through the first $n$ layers of our SAE LLM and decode the latent activations into SAE feature-space activations. For both datasets, we use \texttt{gemma-3-4b-pt} and \texttt{google/gemma-scope-2-4b-pt/resid\_post/layer\_22\_width\_65k\_l0\_medium} from GemmaScope 2 \citep{McDougallGemmaS2}. We decode the activations from the 22nd layer and get sparse 65k-dimensional feature activation vectors. The activation for the initial \texttt{<BOS>} token is dropped and we max-pool over the remaining tokens to get an SAE activation vector across the segment.

\subsection{Feature Filtering and Clustering}
\label{app:clustering}

Due to the relatively large size of the segments and the max-pooling operation we perform, we see over 1k features active per segment on average. If we were to feed every single positive activation as a token to the encoder, then the attention map would be more than 16 times the size of the decoder's attention map over the corresponding tokens. Furthermore, many raw SAE features are redundant, polysemantic, or uninterpretable.

To address this, we whittle down the features with filtering and clustering. The first step is a feature whitelist, which we obtained by asking \texttt{gpt-5.6-luna} to go through each feature's Neuronpedia entry (which includes a description, examples of activating sequences, activation statistics, etc.) and determine if it is a consistent, high-quality feature. We also manually independently inspected 150 features and found about 80\% agreement with \texttt{gpt-5.6-luna}'s classifications.

Then, to address the high quantity of redundant features, we perform a clustering step. We first re-generate feature descriptions based on examples from the Neuronpedia API, as we found erroneous and empty descriptions from the API itself. Then, we group features based on description embeddings as a kind of high-level semantic clustering. From the segment activations, we remove the dimensions corresponding to filtered-out features and max-pool over clustered features.

\paragraph{Counts at each stage.}
Filtering keeps $16{,}261$ of the $65{,}536$ SAE features: it drops features whose activation frequency falls outside $[10^{-5}, 0.5]$ and features whose description and top-activating examples the reviewing model judges inconsistent or low-level. The first clustering stage merges near-synonymous descriptions ($4{,}319$ clusters with the original descriptions, $4{,}304$ after we regenerated descriptions from activating examples because the API's were often empty or wrong); the second re-clusters the centroids toward a target size ($1{,}785$ and $1{,}714$ clusters). At this point a segment still has $423$ active clusters on average, most of them generic. The selection step therefore keeps the subset of clusters chosen by a facility-location greedy to maximize coverage of the corpus under a budget of at most $100$ active clusters per segment: $951$ clusters with the original descriptions (coverage $0.635$) and $844$ with the regenerated ones (coverage $0.643$); simply dropping the highest-frequency clusters reaches only $0.613$. \oursbig{} uses the $844$-cluster vocabulary ($7{,}208$ features; mean $8.5$, median $5$, max $176$ members per cluster). \ourssmall{} was trained before the selection step and uses the $1{,}785$ clusters of the first version together with the remaining singletons, $2{,}074$ units in all, of which $2{,}000$ carry a description.

\subsection{Feature Selection and Tokenization}\label{app:feature_selection}

Even after filtering and clustering SAE features, roughly 100 units are active per segment. As a final step to reduce the encoder's input to a manageable size, each training signal $f$ (Section~\ref{sec:method}) marks only a subset of units: a set $f^{+}$ of positive units and a set $f^{-}$ of negative units, with every other unit set to $0$ and therefore omitted from the encoder input.

A fixed number of positive and negative units risks a model that only works with that exact number of encoder inputs, so we draw the counts $k^{+}=|f^{+}|$ and $k^{-}=|f^{-}|$ at random from a truncated geometric distribution. For integers $k_{\min}\le k_{\max}$ and a mean $\mu>0$, let $\kappa(k_{\min},k_{\max},\mu)$ be the law of
\begin{equation}
\label{eq:countlaw}
k=k_{\min}+\big(G\bmod(k_{\max}-k_{\min}+1)\big),
\qquad
\Pr[G=g]=(1-\rho)\rho^{g}\quad(g=0,1,2,\dots),
\end{equation}
where $\rho=\mu/(1+\mu)$, so that $G$ is geometric with $\mathbb E[G]=\mu$ and $k$ takes values in $\{k_{\min},\dots,k_{\max}\}$. Training draws $k^{+}\sim\kappa^{+}=\kappa(1,48,12)$ and $k^{-}\sim\kappa^{-}=\kappa(0,48,12)$ independently for every segment.

\paragraph{Positive units.} Given $k^{+}$, we set $f^{+}$ to the $\min\{k^{+},|A(y)|\}$ active units with the largest activation. We do not compare raw activations across units, because their scales differ. Instead, we map each activation to its percentile under that unit's activation distribution. The percentiles are computed by interpolating quantiles pre-gathered from the pretraining data, and activations outside the stored range are clipped.

\paragraph{Negative units.} Negative units are chosen differently depending on whether the preceding segment boundary was spliced (Appendix~\ref{app:splicing}):
\begin{itemize}
    \item If it was not spliced, $f^{-}$ is a sample of $k^{-}$ units from the inactive set $I(y)$.
    \item If it was spliced, we take the percentile-space difference between the activations of the \emph{counterfactual} segment (the one that originally followed) and those of the actual segment. $f^{-}$ is then the $k^{-}$ units with the largest difference, i.e., units that \emph{should} have fired strongly but did not.
\end{itemize}
Our intuition is that this maximizes the amount of information. The encoder thus receives $|f^{+}|+|f^{-}|\le k^{+}+k^{-}$ tokens.

\subsection{Model Architecture and Modelling Objectives}\label{app:model_architecture}

We use a feed-forward encoder-decoder transformer with non-grouped multi-head attention \citep{vaswani2017attention}, RMSNorm \citep{zhang2019rmsnorm}, a 4$\times$-width MLP with squared ReLU activation \citep{so2021primer}, QK normalization \citep{henry2020qknorm,dehghani2023vit22b}, interleaved short-window attention layers \citep{child2019generatinglongsequencessparse,brown2020languagemodelsfewshotlearners}, and value embeddings \citep{zhou2025value,modded_nanogpt_2024}.

In addition to the standard next-token prediction objective, we also add a next-\textit{segment} prediction objective. Accompanying this, we add a feature-prediction head alongside the language-modelling head. This is simply a linear layer added to the end of the model residual alongside the language modelling head.

To predict the next segment's feature activations, we max-pool the feature-prediction head over the previous segment. During training, this becomes the prediction for the modelling objective; during inference, we select the top-$k^+$ and bottom-$k^-$ predicted features as the conditioning signal for the next segment.

To encode positive and negative feature tokens, we maintain a single conditioning vocabulary embedding dictionary; by default, these represent positive features. For negative features, we keep a trainable embedding which gets added to the corresponding feature vocabulary embedding to represent a negative embedding, similar to additive positional embeddings in \cite{vaswani2017attention}. We also initialize the feature embedding dictionary of \oursbig{} with a random projection of the SAE decoder matrix and an SVD-based projection for \ourssmall{}.

Hyperparameters, including those for previous subsections on dataset processing, are available in \Cref{tab:hparams}.

\begin{table*}[!ht]
\caption{Architecture, pretraining and finetuning settings. Feature embeddings are
initialized by projecting SAE decoder directions. The lower block describes the
supervised finetuning stage, including the residual MLP blocks added to the
feature-prediction (FP) head.}
\label{tab:training-hyperparameters}
\label{tab:hparams}
\centering
\setlength{\tabcolsep}{4pt}
\rowcolors{2}{white}{gray!10}
\begin{tabular}{@{}lccc@{}}
\toprule
Hyperparameter & \oursbig{} & \ourssmall{} & Decoder-only control \\
\midrule
Parameters (pretraining) & 2.988B & 412.26M & 399.12M \\
Encoder / decoder layers & 27 / 27 & 12 / 12 & -- / 14 \\
Hidden width & 1,664 & 768 & 896 \\
Query / KV heads & 13 / 13 & 6 / 6 & 7 / 7 \\
Head dimension & \multicolumn{3}{c}{128} \\
MLP width & 6,656 & 3,072 & 3,584 \\
Token vocabulary & \multicolumn{3}{c}{32,768 (9 special tokens)} \\
Context length & 2,048 & 1,024 & 1,024 \\
Short-attention window & 1,024 & 512 & 512 \\
Attention pattern & \multicolumn{3}{c}{SSSL; final layer full context} \\
Cross-attention & \multicolumn{2}{c}{Every decoder layer} & -- \\
\midrule
Feature vocabulary & 844 & 2,074 & -- \\
SAE feature initialization & Random proj. & SVD proj. & -- \\
Segment length (Gemma) & \multicolumn{2}{c}{128; shorter tails retained} & -- \\
Splice limit (Gemma) & \multicolumn{2}{c}{2,048} & -- \\
Splice cut fraction & \multicolumn{2}{c}{0.15 per shard} & -- \\
Positive / negative counts & \multicolumn{2}{c}{$1$--$48$ / $0$--$48$} & -- \\
Geometric draw mean & \multicolumn{2}{c}{12 (before count mapping)} & -- \\
Polarity buckets & \multicolumn{2}{c}{2 (positive / negative)} & -- \\
FP loss weight & \multicolumn{2}{c}{0.1} & -- \\
\midrule
Corpus tokens & 53.819B & 9.797B & $\approx 11.7$B$^{a}$ \\
Pretraining steps & 250,000 & 54,600 & 54,600 \\
Tokens per update$^{b}$ & 1,835,008 & 524,288 & 524,288 \\
Sequences / GPU microbatch & 14 & 16 & 64 \\
Optimizers & \multicolumn{3}{c}{Muon: matrices; AdamW: embeddings, heads, scalars} \\
Matrix LR$^{c}$ & \multicolumn{3}{c}{0.02} \\
Text/value embedding LR$^{c}$ & \multicolumn{3}{c}{0.3} \\
Encoder embedding LR$^{c}$ & \multicolumn{2}{c}{0.2} & -- \\
LM/FP head LR$^{c}$ & \multicolumn{3}{c}{0.004} \\
Scalar LR$^{c}$ & \multicolumn{3}{c}{0.5} \\
AdamW $(\beta_1,\beta_2)$ & \multicolumn{3}{c}{$(0.8,0.95)$; $x_0$: $(0.96,0.95)$} \\
Weight decay$^{c}$ & 0.2 & 0 & 0.2 \\
LR warmup & \multicolumn{3}{c}{0 steps} \\
LR decay & \multicolumn{3}{c}{Linear to zero over final 50\% of steps} \\
Precision & \multicolumn{3}{c}{BF16; FP8 disabled} \\
\midrule
\rowcolor{white}\multicolumn{4}{@{}l}{\emph{Supervised Finetuning}} \\
Mixture (conversations) & \multicolumn{3}{c}{1,246,593; 361,000 steered (29.0\%); 36,000 MC-binding (2.9\%)} \\
Segment length (Gemma) & \multicolumn{2}{c}{32, turn-delimited; no splicing} & -- \\
Loss mask & \multicolumn{3}{c}{Assistant text and tool calls} \\
Steps (one epoch) & 1,120 & 1,013 & 1,001 \\
Weight decay / gradient clipping & \multicolumn{3}{c}{0 / none} \\
FP-head MLP blocks & 2 & 1 & -- \\
\bottomrule
\end{tabular}
\vspace{2pt}
\begin{minipage}{\textwidth}
\footnotesize
$^{a}$Packed decoder-token counts, excluding validation and repeated passes. The control shares \ourssmall{}'s 369 source shards (19.68M documents), but uses different packing; its packed-token total remains unverified.
\par\noindent
$^{b}$Final-phase batch size; the 3B run's batch size changed slightly during training due to GPU failures.
\par\noindent
$^{c}$LRs are configuration coefficients scaled by $\sqrt{B/524288}$ for token batch size $B$; embedding/head rates additionally scale by $(d/768)^{-1/2}$ for hidden width $d$. Muon also scales by matrix shape. Weight decay is scaled by batch size and training budget, decreases linearly to zero, and applies only to Muon groups.
\end{minipage}
\end{table*}

\subsection{Supervised Finetuning}
\label{app:sft}

Both models are finetuned on a conversational mixture after pretraining. The mixture is based on Nanochat's standard SFT data---SmolTalk ($460{,}341$ sample conversations), two copies of a small identity file ($2{,}000$) that teaches the model about itself ("I am a language model trained by Andrej Karpathy..."), MMLU auxiliary training questions ($99{,}842$), GSM8K training questions ($7{,}473$), multiple-choice binding data ($36{,}000$, see \Cref{app:mc-binding}) and two spelling tasks ($280{,}000$)---together with the steered split described in \Cref{app:steered-sft} ($361{,}000$): $1{,}246{,}593$ conversations in all, of which $29.8\%$ are steered. \oursbig{}, \ourssmall{} and our 400M decoder-only baseline are finetuned on this dataset; \pythia{} is finetuned on it without the steered split, which its two post-trained variants use instead (\Cref{app:baselines}).

Conditioning is built during finetuning in the same way as during pretraining. Conversations are rendered, converted to the same format as pretraining shards, and passed through the same extraction pipeline (\Cref{app:sae}) so that the encoder is supplied with SAE features of the conversation itself. Though we use the same pipeline, we make two important configuration changes from pretraining: first, segments are turn-delimited. We first add segmentation boundaries when the text changes from user to assistant or vice versa. If an individual conversation turn is too long, we apply fixed-length segmentation at a window size of 32 Gemma-3 tokens rather than 128 like in pretraining. This segments roughly every two sentences of a reply, which is the granularity at which we later steer. Second, we disable splicing, so there are no counterfactual continuations and negative features are sampled from the inactive features.

To ensure that the model learns to \textit{listen} to steering signals from steered examples while still maintaining realistic, in-distribution behavior, we disable next-segment feature prediction \textit{only} for the first segment in a steered assistant reply. Without this, the feature-prediction head would be fitting to noise as the responses are steered in random semantic directions. Within the reply itself, segment-to-segment prediction proceeds as usual. This is also the stage at which we expand the feature-prediction head with residual MLP blocks initialized to the identity (\Cref{tab:hparams}).

We use assistant masking, so loss is taken on assistant text and tool calls only; user turns and tool outputs injected into the stream are excluded. We train for a single epoch at a global batch of $524{,}288$ tokens, with the learning-rate coefficients of pretraining, no weight decay and no gradient clipping; this is $1{,}120$ steps for \oursbig{} and $1{,}013$ for \ourssmall{}.

\subsection{Steered Supervised Finetuning Data}
\label{app:steered-sft}

Similar to how we splice pretraining documents to decrease the predictability of documents conditioned solely on previous text context, we generate SFT documents with assistant responses steered in random directions. Beyond the fact that this encourages the model to adhere more strongly to the steering signal, it also provides natural-language, conversational examples of steered text.

\paragraph{Concepts.} The concepts are the descriptions of features from a second, larger SAE---\texttt{gemma-3-27b-it}, layer 53, 262k features---and are used only as writing targets for the generator; they are unrelated to the units the encoder is conditioned on, which are extracted from the generated text as above. We drop descriptions that name a surface form rather than a concept and deduplicate near-identical ones by sentence embedding, leaving $14{,}917$ usable concepts; a further $2{,}252$ are held out and never used.

\paragraph{Generation.} We draw $137{,}001$ instructions from the standard mixture, keeping those with an instruction of at least $20$ characters and an original reply of at least $200$, and use no instruction twice. Each instruction is paired with three concepts whose descriptions are pairwise dissimilar (cosine below $0.55$), and each concept is used $27$--$28$ times. For each pair, we ask \texttt{gpt-4.1-mini}, at the provider's default sampling settings, for a reply that answers the instruction and is written through the concept, under three constraints: the concept must be present from the first sentence rather than introduced later, it must never be named or described, and where the instruction has one correct answer that answer must not change. We also generate one unsteered reply per instruction, as the reference for the check below; $548{,}004$ generations in all.

\paragraph{Choice of generator.} We picked the generator by measuring what its replies do to the conditioning features rather than by reading them. On a pilot of $667$ instructions and the same $2{,}001$ instruction--concept pairs, we generated with \texttt{gpt-4.1-mini} and with the newer and nominally stronger \texttt{gpt-5.6-luna}, the latter both with and without reasoning, and extracted from each reply the conditioning signal it would produce, in the percentile space the encoder consumes. Two replies that answer the same instruction under different concepts share $0.185$ of their top-$32$ features under \texttt{gpt-4.1-mini}, against $0.280$ under \texttt{gpt-5.6-luna} in both modes; for reference, three unsteered resamples of one instruction share $0.524$. Replies from either generator read as themed, but \texttt{gpt-4.1-mini} carries the theme far enough into the content to move the features the channel is trained on, which is the only part of the difference finetuning can see---and on this axis the older, smaller model is the better teacher.

\paragraph{Quality control.} A theme that changes the answer is a corrupted training example, and the last constraint is the one a generator is most likely to break. The same model compares each steered reply against the unsteered reply to the same instruction and returns one of three labels: \textsc{open} if the instruction has no single correct answer, \textsc{agree} if it has one and both replies reach it, \textsc{contradict} if they reach different ones; differences of style, length, imagery or framing are explicitly not contradictions. We drop only \textsc{contradict}: $49{,}985$ of the $341{,}960$ replies with a checkable answer ($14.6\%$), or $12.2\%$ of all steered replies. $361{,}018$ steered conversations remain. The unsteered references are used only for this comparison and are not part of the mixture.

\paragraph{Known residue.} Two artifacts survive into the shipped split. The spelling tasks in the standard mixture contribute counting drills (``how many \texttt{r}'s are in \emph{strawberry}''), which our instruction filter does not reliably catch---about $0.9\%$ of the sampled instructions are of this kind---and a themed reply to such an instruction is at best decoration. And steered replies are considerably longer than the unsteered references, $1{,}167$ against $338$ characters at the median, so the split shifts reply length as well as content.

\subsection{Multiple-Choice Binding Data}
\label{app:mc-binding}

From multiple rounds of iteration on the SFT dataset, we found that previous iterations of \oursbig{} were poorly suited to responding with single-letter answers to questions in multiple-choice format. Despite instructions in the prompts to respond with just the appropriate letter, they would give longer, paragraph-length responses containing the correct answer in natural language format. To improve the model's ability to bind known answers to multiple-choice knowledge, we added in synthetic data based on MMLU prompts designed to improve this specific ability. It teaches the model to answer multiple-choice questions from answers in-context; below is an example from the dataset.

\begin{quote}
\begin{minipage}{\linewidth}
\small
\hrule
\medskip
\begin{verbatim}
User:
- He feels helpless without Rupert.=A
- He feels good without Rupert .=B
- He is very angry .=C
- He thinks it rather funny .=D

The option "He feels good without Rupert ." is marked
with which label? Reply with just the label.

Assistant:
B
\end{verbatim}
\medskip
\hrule
\end{minipage}
\end{quote}

\section{Evaluation Details}
\label{app:evaluation}

This appendix documents the evaluation protocol shared across all 3B results: the sources of concepts and instructions, how the judge scores each reply, the strength parameter for each method and how the attack strength was selected, and how the three open models were prepared. The attacks themselves are specified in \Cref{sec:experimental_setup}, with per-model settings reported in \Cref{app:gepa,app:counter-3b}.

\subsection{Concepts and Instructions}
\label{app:concepts}
Concepts are drawn from AxBench's Concept500 list~\citep{wu2025axbenchsteeringllmssimple}, where each concept is a short natural-language description of a feature (e.g., \emph{``terms related to connections in electrical circuits and network structures''}). The clean evaluation uses $100$ concepts held out from every stage of \oursbig{}'s training, including training of the NL encoder; the attack evaluations use the first $50$ of these.

Each concept is paired with the same $10$ AlpacaEval instructions, sampled by AxBench with a fixed seed and unrelated to the concept, so that a steered reply must both reflect the concept and still answer the instruction. Each method generates one reply per instruction ($128$ new tokens, temperature $1.0$, and seed $42$.) Thus, each concept-by-strength cell contains $10$ generations, and each reported score is the mean across the $100$ (or $50$) concepts of the corresponding per-concept mean.

For the baselines, AxBench also provides the per-concept training data used to fit the steering vectors and LoRA: $144$ examples per concept, generated by the same language model that later serves as the judge, with half exhibiting the concept and half not.

\subsection{LM Judge}
\label{app:judge}

Replies are scored by \texttt{gpt-5.6-luna} following AxBench's three-axis rubric. Each axis takes an integer value in $\{0,1,2\}$:

\begin{itemize}\itemsep1pt
  \item \emph{concept}: whether the reply is about the concept ($0$ absent, $1$ mentioned, $2$ clearly incorporated);
  \item \emph{instruction}: whether the reply addresses the instruction;
  \item \emph{fluency}: whether the reply is coherent English.
\end{itemize}

We use the concept score to measure steering and treat fluency and instruction-following as its costs. For example, in the clean evaluation (\Cref{sec:clean-3b,app:clean3b}), we report the cost as the harmonic mean of instruction and fluency, computed \emph{per generation} and then averaged, so that a reply scoring zero on either axis contributes zero regardless of its score on the other.
Under attack (\Cref{sec:blackbox,sec:whitebox,app:counter-3b}), we report concept score and fluency separately, since the attack aims to remove the concept and the question of interest is how much response quality must be sacrificed to do so.

The RL attacker (\Cref{app:rl-redteam}) is trained against a smaller local judge fine-tuned on these ratings. That judge only supplies training rewards; every score we report comes from \texttt{gpt-5.6-luna}.

\subsection{Steering Methods and Strength Grids}
\label{app:strengths}

In addition to \cond{}, we introduce \condmask{}. We also introduce \nlretrieval{} and correspondingly disambiguate \nlsteer{} into \nlencoder{}.

\paragraph{Masked conditioning.} In \condmask{}, we first compute a conditioning vector as in \cond{}, and then mask out positive and negative units in the model's self-conditioning if they are too low or high, respectively, in the conditioning vector. 

\paragraph{Strength.} Conditioning, its masked variant, and both NL interfaces share a single strength parameter: the factor by which features supplied through the control channel are scaled. \diffmean{} and \linearprobe{} use the AxBench factor instead: the fitted unit direction is added to the residual stream at the steering layer, scaled by the factor times the largest activation of that concept along the direction on its training examples. All six methods are swept over the same $12$ values, $\{0, 0.2, 0.4, 0.6, 0.8, 1, 1.5, 2, 3, 4, 5, 6\}$, in the clean evaluation. Because the factor is defined in each method's own units, the same value does not correspond to the same intervention across methods or models; \Cref{app:clean-sweeps} reports the full curve for each method. \lora{} and \promptsteer{} have no strength parameter and are evaluated only once.

\paragraph{Strength under attack.} Each attack evaluates a method at a single strength, chosen before the attack from that model's clean run on the $50$ attack concepts. We select the factor that maximizes the harmonic mean of the concept score and the fluency score, both averaged over concepts, with ties going to the smaller factor. 
We use this rule to choose an ``operating point'' for every method on every model.
\Cref{tab:app-attack-factors} lists the selected strengths. All four \oursbig{} interfaces select the top of the grid, as does \linearprobe{} on \qwen{}; \diffmean{}, and \linearprobe{} on \pythia{} and \smollm{}, select lower factors because their fluency degrades early (\Cref{tab:app-clean-sweep-hosts}), which is why they enter the attack evaluation with lower concept scores than in \Cref{tab:app-clean-best}.

\begin{table}[!ht]
\caption{\textbf{Steering strength used under attack.} The factor selected for each method on each model using the rule in \Cref{app:strengths}, based on the clean run over the $50$ attack concepts. $^{\dagger}$All four interfaces of \oursbig{} (\cond{}, \condmask{}, \nlencoder{}, \nlretrieval{}) select the same factor.}
\label{tab:app-attack-factors}
\centering\small
\begin{tabular}{@{}lcccc@{}}
\toprule
Model & Control channel$^{\dagger}$ & \diffmean{} & \linearprobe{} & \lora{}, \promptsteer{} \\
\midrule
\oursbig{} & $6$ & --- & --- & --- \\
\pythia{}  & --- & $0.2$ & $3$ & single setting \\
\qwen{}    & --- & $1.5$ & $6$ & single setting \\
\smollm{}  & --- & $0.8$ & $1.5$ & single setting \\
\bottomrule
\end{tabular}
\end{table}

\paragraph{Baseline fitting.} 
The baselines use AxBench's implementations with default hyperparameters, applied to each open model at the steering layer listed in \Cref{tab:app-counter-grids}. \diffmean{} is the difference between the mean residuals of the concept and non-concept training examples. \linearprobe{} is a rank-one logistic probe fit to the same residuals ($24$ epochs, learning rate $5{\times}10^{-3}$, weight decay $10^{-3}$). Both are normalized to unit directions and scaled as described above. \lora{} is rank $4$ on the attention output projection at four evenly spaced layers ($7$, $14$, $21$, and $28$ on the $36$-layer models; $\alpha{=}32$, $24$ epochs, learning rate $9{\times}10^{-4}$), and is trained only on the concept examples. \promptsteer{} prepends AxBench's steering instruction naming the concept to the user turn.

\paragraph{Steering layer.}
For \qwen{} and \smollm{}, the steering vectors are fitted at layer $28$ of $36$; for \pythia{}, at layer $25$ of $32$, corresponding to the same relative depth. \oursbig{} is conditioned where its architecture reads the control channel, after the cross-attention at layer $21$ (\Cref{sec:method}).

\paragraph{Human annotator agreement.}
Three human annotators blindly labelled 150 samples from our evaluations, including our model at various strengths, baseline methods, negative steering, and completions with adversarial steering pressure. The complete confusion matrices are available in \Cref{tab:human-llm-confusion} and overall human-LLM/human-human agreement metrics are in \Cref{tab:human-llm-agreement}. We find similar levels of human-annotator agreement in \Cref{tab:human-llm-agreement} as human-LLM agreement. This alongside the monotonic increase in the human mean rating conditional on LLM ratings give us confidence in this labelling schema as a way to highlight semantically meaningful differences between methods.

\begin{table*}[t]
\centering
\small
\setlength{\tabcolsep}{3pt}
\renewcommand{\arraystretch}{0.95}
\captionsetup[subtable]{font=small,skip=3pt}

\begin{subtable}[t]{0.32\textwidth}
    \centering
    \caption{Concept presence}
    \begin{tabular}{@{}lrrr@{}}
        \toprule
        Human $\backslash$ LLM & 0 & 1 & 2 \\
        \midrule
        0 & \textbf{45} & 10 & 5 \\
        1 & 8 & \textbf{14} & 6 \\
        2 & 4 & 27 & \textbf{31} \\
        \midrule
        Human mean & 0.28 & 1.33 & 1.62 \\
        \bottomrule
    \end{tabular}
\end{subtable}
\hfill
\begin{subtable}[t]{0.32\textwidth}
    \centering
    \caption{Instruction relevance}
    \begin{tabular}{@{}lrrr@{}}
        \toprule
        Human $\backslash$ LLM & 0 & 1 & 2 \\
        \midrule
        0 & \textbf{1} & 11 & 10 \\
        1 & 0 & \textbf{18} & 37 \\
        2 & 0 & 13 & \textbf{60} \\
        \midrule
        Human mean & 0.00 & 1.05 & 1.47 \\
        \bottomrule
    \end{tabular}
\end{subtable}
\hfill
\begin{subtable}[t]{0.32\textwidth}
    \centering
    \caption{Fluency}
    \begin{tabular}{@{}lrrr@{}}
        \toprule
        Human $\backslash$ LLM & 0 & 1 & 2 \\
        \midrule
        0 & \textbf{5} & 10 & 0 \\
        1 & 4 & \textbf{43} & 0 \\
        2 & 0 & 73 & \textbf{15} \\
        \midrule
        Human mean & 0.44 & 1.50 & 2.00 \\
        \bottomrule
    \end{tabular}
\end{subtable}

\caption{Human--LLM rating confusion matrices. Bold entries indicate exact agreement. The final row gives the mean human rating conditional on each LLM rating.}
\label{tab:human-llm-confusion}
\end{table*}

\begin{table*}[t]
\centering
\small
\begin{tabular}{@{}lrrrrrr@{}}
    \toprule
    & \multicolumn{3}{c}{Human--LLM judge}
    & \multicolumn{3}{c}{Human--human} \\
    \cmidrule(lr){2-4}
    \cmidrule(l){5-7}
    Metric
    & Exact agreement & Pearson $r$ & $r^2$
    & Exact agreement & Pearson $r$ & $r^2$ \\
    \midrule
    Concept presence
    & 60.0\% & 0.616 & 0.380
    & 54.4\% & 0.524 & 0.274 \\
    Instruction relevance
    & 52.7\% & 0.295 & 0.087
    & 35.3\% & 0.317 & 0.101 \\
    Fluency
    & 42.0\% & 0.427 & 0.182
    & 58.8\% & 0.526 & 0.277 \\
    \bottomrule
\end{tabular}
\caption{Human--LLM judge and human--human annotator agreement
across 150 and 68 paired ratings per metric, respectively.
Human--human pairs may share a response.
Here $r^2$ corresponds to a linear fit with an intercept.}
\label{tab:human-llm-agreement}
\end{table*}

\subsection{Natural-Language Steering Interfaces}
\label{app:nl}

Both interfaces map a one-sentence description $c$ of the concept to a conditioning vector over the $M$ units. \nlretrieval{} embeds the sentence and every unit's description with the same $768$-dimensional sentence encoder, scores each unit by cosine similarity, centers the scores at the median and rescales them to $[-1, 1]$, giving the retrieval vector $r(c)$. Raw cosine similarities lie in roughly $[0, 0.9]$, so without centering nearly every unit would count as positive and there would be no negative pool. \nlencoder{} adds a residual correction,
\[
  g(c) \;=\; \tanh\!\big(r(c) + U\,\sigma(V e_c)\big),
\]
where $e_c$ is the sentence embedding and $U \in \mathbb{R}^{M \times 32}$, $V \in \mathbb{R}^{32 \times 768}$ form a rank-$32$ bottleneck. The residual branch is initialized to zero, so the model starts from retrieval and training learns only the correction. The encoder is fit once against difference-of-means targets on a set of concepts disjoint from those used for evaluation, and is then shared across all concepts. On held-out concepts it raises the cosine similarity to the fitted target from $0.03$ under retrieval to $0.41$.

\subsection{Baseline Models}
\label{app:baselines}
\paragraph{\qwen{} and \smollm{}} \citep{qwen2025qwen25technicalreport,bakouch2025smollm3} are used as released through their chat templates. \smollm{} is run with reasoning mode disabled (\texttt{/no\_think}), so that the judged reply is the answer rather than a thinking trace, and so that it is compared on the same terms as \oursbig{}, which has not undergone reasoning training. Both models have been post-trained far more heavily than \oursbig{}, and their instruction-following and fluency scores are correspondingly higher (\Cref{app:clean3b}).

\paragraph{\pythia{}} is \pythiabase{}~\citep{pmlr-v202-biderman23a}, which has no instruction-tuned release. We finetune it ourselves for one epoch on \oursbig{}'s SFT data (\Cref{app:sft}), including the multiple-choice binding data (\Cref{app:mc-binding}) but not the steered conversations, with the loss on assistant tokens only. We steer this model with the four post-hoc methods. Together with its two post-trained variants below, it is our closest matched comparison at 3B: they see the same post-training data as \oursbig{}, but not our pretraining.

\paragraph{Post-trained baselines.} These two baselines ask how much of our results the steered data alone explains: they learn from the same steered conversations as \oursbig{}, but only in post-training and only through the prompt. We finetune \pythia{} a second time on the $361{,}018$ steered conversations of \oursbig{}'s SFT data (\Cref{app:steered-sft}), now with the concept that shaped each reply written into the prompt. We also add the $137{,}001$ unsteered reference replies under their plain instructions, so that the model still replies normally when no concept is given. The two variants differ only in how the concept is written:
\begin{itemize}
  \item \promptsft{} prepends to the question a steering instruction for the concept, written by \texttt{gpt-5.6-luna} with the AxBench template that produces \promptsteer{}'s instructions. At evaluation it receives \promptsteer{}'s instruction for each concept, so the two differ only in the finetuning. The comparison therefore favors \promptsft{} over \oursbig{}: \promptsft{} receives the concept in the same form in training and evaluation, an instruction written with the same template, whereas \oursbig{} is trained on conditioning signals extracted from the reply itself and steered at evaluation with signals computed from the concept's examples or description.
  \item \steerlmsft{} prepends, in the style of SteerLM \citep{dong2023steerlmattributeconditionedsft}, a list of attributes with levels (``Attributes: label:level, \dots''). The attributes are conditioning units of \oursbig{}, each written as a label of at most six words by \texttt{gpt-5.6-luna} from its member features' descriptions, with a level from $0$ to $10$ fixed by its \nlretrieval{} score for the concept rather than swept. In training, as in SteerLM, the attributes describe the reply that follows: we take the eight units that \nlretrieval{} ranks highest for the concept and keep those, at most four, that the steered reply activates more strongly than the unsteered reply to the same instruction (by at least $0.2$ in percentile), dropping the $7{,}735$ conversations with none. At evaluation the reply does not exist yet, so the attributes are simply the four highest-ranked units; like \promptsteer{} and \nlretrieval{}, this needs only the concept description. It thus receives the kind of signal that \nlretrieval{} gives \oursbig{}, but through the prompt and only in post-training.
\end{itemize}
Both start from \pythia{} and are finetuned in the same way, for one epoch. Like \promptsteer{} and \lora{}, they have no strength parameter.

\section{Clean Steering Evaluation}
\label{app:clean3b}

We report the clean steering evaluation in full: the operating point each method reaches, the complete strength sweeps behind \Cref{fig:steering_main}, and the factor view on all three open models.

\subsection{Best Operating Point per Model and Method}
\label{app:clean-best}

\Cref{tab:app-clean-best} summarizes each method at the factor that maximizes its concept score, corresponding to the operating point in the trade-off planes of \Cref{fig:steering_main}. Reading the three axes together separates methods that steer from methods that simply overwrite the output: \diffmean{} on \pythia{} reaches a concept score of $1.11$, but only at an instruction score of $0.15$ and a fluency of $0.03$, whereas \oursbig{}'s conditioning reaches $1.31$ with instruction $1.79$ and fluency $0.97$. The methods without a strength parameter keep the reply fluent and on-instruction on every model. \promptsteer{} reaches $1.19$ on \qwen{} and $1.21$ on \smollm{}; \oursbig{}'s conditioning is higher by $0.11$ {\scriptsize[$-0.00$, $+0.23$]} and $0.09$ {\scriptsize[$-0.01$, $+0.19$]} (paired by concept, $95\%$ bootstrap interval), within noise, at a harmonic mean lower by $0.51$ and $0.43$. On \pythia{}, prompting reaches only $0.45$, but finetuning \pythia{} on our steered data with the concept in the prompt raises it to $1.61$ (\promptsft{}, which the comparison favors; \Cref{app:baselines}), the highest concept score in the table, and to $1.13$ with a list of conditioning units (\steerlmsft{}). Both are less robust to adversarial prompts than \oursbig{} (\Cref{sec:blackbox,sec:whitebox}).

\begin{table}[!htbp]
\caption{\textbf{Best operating point per model and method.} For each method, the
steering factor that maximizes the concept score, and the judge's three axes at
that factor. \lora{}, \promptsteer{} and the two post-trained baselines on \pythia{}
have no steering strength and are marked N/A. $100$ held-out concepts. The harmonic mean is taken per generation
and then averaged, so a reply scoring zero on either component contributes zero.}
\label{tab:app-clean-best}
\centering\small
\begin{tabular}{@{}llcccc@{}}
\toprule
Model & Method & Factor & Concept & Instruction & Fluency \\
\midrule
\oursbig{} & \cond{} & 6 & \ourul{1.31} & 1.79 & 0.97 \\
 & \condmask{} & 6 & 0.95 & \ourul{1.83} & \ourul{0.98} \\
 & \nlencoder{} & 6 & 1.14 & 1.80 & 0.93 \\
 & \nlretrieval{} & 6 & 1.10 & 1.74 & 0.83 \\
\midrule
\pythia{} & \diffmean{} & 6 & 1.11 & 0.15 & 0.03 \\
& \linearprobe{} & 6 & 0.38 & 1.54 & 0.36 \\
 & \lora{} & N/A & 0.89 & 1.85 & 1.17 \\
& \promptsteer{} & N/A & 0.45 & 1.85 & \ourul{1.22} \\
\cmidrule(l){2-6}
 & \promptsft{} & N/A & \ourul{1.61} & 1.95 & 1.08 \\
 & \steerlmsft{} & N/A & 1.13 & \ourul{1.96} & 1.06 \\
\midrule
\qwen{} & \diffmean{} & 2 & 0.97 & 1.13 & 0.40 \\
 & \linearprobe{} & 6 & 0.37 & \ourul{1.98} & 1.24 \\
 & \lora{} & N/A & 1.03 & 1.94 & \ourul{1.57} \\
 & \promptsteer{} & N/A & \ourul{1.19} & 1.97 & \ourul{1.57} \\
\midrule
\smollm{} & \diffmean{} & 1.5 & 0.92 & 1.18 & 0.25 \\
 & \linearprobe{} & 2 & 0.51 & 1.74 & 0.56 \\
 & \lora{} & N/A & 0.95 & 1.96 & \ourul{1.49} \\
 & \promptsteer{} & N/A & \ourul{1.21} & \ourul{1.98} & 1.45 \\
\bottomrule
\end{tabular}

\end{table}

\subsection{Full Strength Sweeps}
\label{app:clean-sweeps}

\Cref{tab:app-clean-sweep-ours,tab:app-clean-sweep-hosts} give the curves of
\Cref{fig:steering_main} in numerical form, for \oursbig{} and for the open
models respectively. Two patterns are easier to read here than in the plots. First, conditioning improves the concept score from $0.37$ at factor $3$ to $1.31$ at factor $6$, while its harmonic mean falls only from $1.30$ to $1.19$. Second, additive steering vectors trade one axis against the other almost at once: \diffmean{} on \pythia{} goes from a concept score of $0.50$ at factor $0.2$ (harmonic mean $1.01$) to $0.82$ at factor $0.4$, where its harmonic mean has already fallen to $0.25$. By factor $4$, its harmonic mean is $0.04$, even as its concept score keeps rising---the judge is detecting concept-related content in output that is no longer coherent language.

\begin{table}[!htbp]
\caption{\textbf{Full strength sweep, \oursbig{}.} Concept score and the
fluency/instruction harmonic mean at every steering factor, $100$ held-out
concepts. These four curves are the ones repeated in every panel of
\Cref{fig:steering_main}.}
\label{tab:app-clean-sweep-ours}
\centering\small
\begin{tabular}{@{}lcccccccc@{}}
\toprule
& \multicolumn{2}{c}{\cond{}} & \multicolumn{2}{c}{\condmask{}}
& \multicolumn{2}{c}{\nlencoder{}} & \multicolumn{2}{c}{\nlretrieval{}} \\
\cmidrule(lr){2-3}\cmidrule(lr){4-5}\cmidrule(lr){6-7}\cmidrule(lr){8-9}
Factor & Conc. & HM & Conc. & HM & Conc. & HM & Conc. & HM \\
\midrule
0 & 0.27 & 1.33 & 0.40 & 1.27 & 0.28 & 1.34 & 0.27 & 1.34 \\
0.2 & 0.27 & 1.28 & 0.39 & 1.25 & 0.27 & 1.30 & 0.27 & 1.30 \\
0.4 & 0.27 & 1.31 & 0.38 & 1.30 & 0.29 & 1.28 & 0.28 & 1.28 \\
0.6 & 0.28 & 1.30 & 0.39 & 1.26 & 0.28 & 1.31 & 0.29 & 1.30 \\
0.8 & 0.27 & 1.29 & 0.38 & 1.28 & 0.30 & 1.31 & 0.29 & 1.30 \\
1 & 0.26 & 1.28 & 0.38 & 1.27 & 0.28 & 1.29 & 0.28 & 1.28 \\
1.5 & 0.28 & 1.30 & 0.37 & 1.28 & 0.29 & 1.29 & 0.29 & 1.27 \\
2 & 0.29 & 1.30 & 0.38 & 1.27 & 0.31 & 1.31 & 0.30 & 1.29 \\
3 & 0.37 & 1.30 & 0.40 & 1.29 & 0.38 & 1.29 & 0.41 & 1.25 \\
4 & 0.57 & 1.28 & 0.42 & 1.25 & 0.62 & 1.23 & 0.75 & 1.17 \\
5 & 1.00 & 1.23 & 0.40 & 1.27 & 1.04 & 1.16 & 0.98 & 1.08 \\
6 & 1.31 & 1.19 & 0.95 & 1.21 & 1.14 & 1.15 & 1.10 & 1.02 \\
\bottomrule
\end{tabular}

\end{table}

\begin{table}[!htbp]
\caption{\textbf{Full strength sweep, baselines.} Concept score and the
fluency/instruction harmonic mean at every steering factor, for the two swept
baseline methods \diffmean{} and \linearprobe{} on each open model. \lora{} and \promptsteer{} have no strength parameter and are
given once, as are the two post-trained baselines on \pythia{}, \promptsft{} and \steerlmsft{}. $100$ held-out concepts.}
\label{tab:app-clean-sweep-hosts}
\centering\small
\setlength{\tabcolsep}{3.2pt}\footnotesize
\begin{tabular}{@{}lcccccccccccc@{}}
\toprule
& \multicolumn{4}{c}{\pythia{}} & \multicolumn{4}{c}{\qwen{}}
& \multicolumn{4}{c}{\smollm{}} \\
\cmidrule(lr){2-5}\cmidrule(lr){6-9}\cmidrule(lr){10-13}
& \multicolumn{2}{c}{\diffmean{}} & \multicolumn{2}{c}{\linearprobe{}}
& \multicolumn{2}{c}{\diffmean{}} & \multicolumn{2}{c}{\linearprobe{}}
& \multicolumn{2}{c}{\diffmean{}} & \multicolumn{2}{c}{\linearprobe{}} \\
Factor & Conc. & HM & Conc. & HM & Conc. & HM & Conc. & HM & Conc. & HM & Conc. & HM \\
\midrule
0 & 0.28 & 1.44 & 0.26 & 1.45 & 0.29 & 1.77 & 0.29 & 1.77 & 0.29 & 1.71 & 0.28 & 1.70 \\
0.2 & 0.50 & 1.01 & 0.28 & 1.45 & 0.31 & 1.78 & 0.28 & 1.78 & 0.32 & 1.67 & 0.30 & 1.69 \\
0.4 & 0.82 & 0.25 & 0.27 & 1.44 & 0.33 & 1.77 & 0.28 & 1.76 & 0.34 & 1.59 & 0.30 & 1.66 \\
0.6 & 0.89 & 0.15 & 0.30 & 1.44 & 0.35 & 1.73 & 0.28 & 1.78 & 0.42 & 1.42 & 0.31 & 1.65 \\
0.8 & 0.93 & 0.12 & 0.29 & 1.42 & 0.44 & 1.62 & 0.29 & 1.75 & 0.61 & 1.10 & 0.33 & 1.61 \\
1 & 0.96 & 0.09 & 0.28 & 1.44 & 0.61 & 1.47 & 0.30 & 1.75 & 0.80 & 0.69 & 0.33 & 1.57 \\
1.5 & 1.02 & 0.06 & 0.31 & 1.41 & 0.94 & 0.92 & 0.30 & 1.77 & 0.92 & 0.30 & 0.43 & 1.23 \\
2 & 1.04 & 0.05 & 0.30 & 1.36 & 0.97 & 0.47 & 0.29 & 1.78 & 0.89 & 0.19 & 0.51 & 0.69 \\
3 & 1.07 & 0.04 & 0.33 & 1.25 & 0.95 & 0.24 & 0.31 & 1.73 & 0.77 & 0.10 & 0.32 & 0.28 \\
4 & 1.09 & 0.04 & 0.35 & 0.99 & 0.91 & 0.13 & 0.34 & 1.70 & 0.55 & 0.05 & 0.19 & 0.16 \\
5 & 1.11 & 0.03 & 0.36 & 0.67 & 0.86 & 0.10 & 0.33 & 1.59 & 0.35 & 0.02 & 0.14 & 0.11 \\
6 & 1.11 & 0.03 & 0.38 & 0.44 & 0.83 & 0.06 & 0.37 & 1.46 & 0.27 & 0.01 & 0.13 & 0.09 \\
\midrule
\lora{} & 0.89 & 1.38 &  &  & 1.03 & 1.68 &  &  & 0.95 & 1.63 &  &  \\
\promptsteer{} & 0.45 & 1.39 &  &  & 1.19 & 1.70 &  &  & 1.21 & 1.62 &  &  \\
\promptsft{} & 1.61 & 1.35 &  &  &  &  &  &  &  &  &  &  \\
\steerlmsft{} & 1.13 & 1.34 &  &  &  &  &  &  &  &  &  &  \\
\bottomrule
\end{tabular}
\end{table}

\subsection{Per-Model Curves}
\label{app:clean-curves}

\begin{figure}[!htbp]
  \centering
  \includegraphics[width=\linewidth]{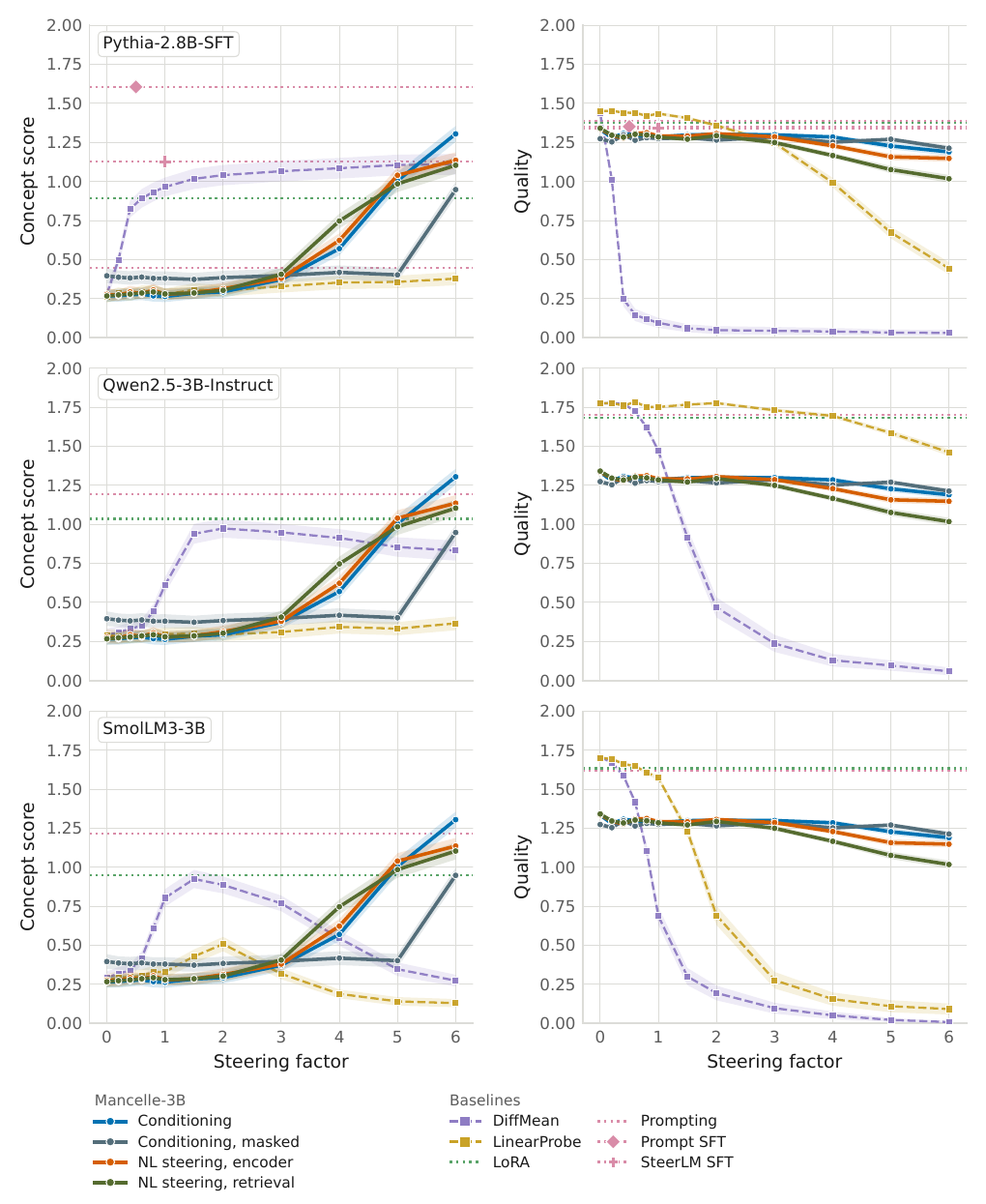}
\caption{\textbf{The steering factor sweep on all three open models.}
Steering vectors pay for concept score on the other axis from their very first step; conditioning does not. Concept score (left) and the
fluency/instruction harmonic mean (right) against the steering factor, for \pythia{} (top), \qwen{} (middle) and \smollm{} (bottom); the same evaluation as \Cref{fig:steering_main} with the factor made explicit. $100$ held-out concepts, $\pm 1$ SEM bands, and dotted lines for the methods with no strength parameter (on \pythia{}, the two marked ones are \promptsft{} and \steerlmsft{}). \diffmean{} on \pythia{} takes most of its concept gain by factor $0.4$, where its harmonic mean has already collapsed; our curves rise gradually and stay flat on the right. \oursbig{}'s curves are the same data in all three rows, which is what makes the rows comparable.}
  \label{fig:app-factor-hosts}
\end{figure}

The main text shows the trade-off plane for all three open models. \Cref{fig:app-factor-hosts} gives the second view, the same data against the steering factor. The overall picture is unchanged: on every model, the swept baselines gain concept score by sacrificing the other two axes, whereas our curves improve more gradually while largely preserving them.
What does change is the starting point of the single-strength baselines---prompting steers far more strongly on the heavily post-trained \qwen{} and \smollm{} than on \pythia{}, where it becomes the strongest method only after finetuning on our steered data (\promptsft{}; \Cref{app:clean-best}).

\section{Robustness to GEPA Suffix Search}
\label{app:gepa}
\Cref{sec:blackbox} reports the GEPA attack as retention: the proportion of each method's own unattacked concept score that survives. This appendix adds what that view leaves out---the absolute scores behind it, what the attack does to fluency, and examples of the suffixes GEPA accepts. All results come from the same campaign as the main text, over $50$ concepts at budgets $k \in \{0, 1, 2, 4, 8\}$.

\subsection{Per-Method Results}
\label{app:gepa-table}

\Cref{tab:app-gepa} reports the concept score at every budget for \oursbig{}'s two interfaces, for the four post-hoc methods on each open model, and for the two post-trained baselines on \pythia{}. At $k{=}8$, \cond{} and \nlencoder{} retain $92\%$ and $88\%$ of their unattacked score. Among the post-hoc methods, \promptsteer{} is the most fragile on both post-trained models, keeping $26\%$ on \qwen{} and $19\%$ on \smollm{}; \lora{} retains $54$--$70\%$ and \linearprobe{} $58$--$69\%$; \diffmean{} retains $71$--$84\%$, but from a much lower starting point. Of the two post-trained baselines on \pythia{}, \promptsft{} retains $85\%$, more than any post-hoc method, and \steerlmsft{} $73\%$.

\cond{} scores above every post-hoc method on every open model at every budget, leading the best of them by $0.10$ unattacked and by $0.49$ at $k{=}8$. \nlencoder{} starts below \promptsteer{} on \qwen{} and \smollm{}, by $0.08$ and $0.06$, but leads every post-hoc method from the first proposal on, by $0.29$ at $k{=}8$. Of the two post-trained baselines, \steerlmsft{} ends below both of our interfaces. \promptsft{}, which the comparison favors (\Cref{app:baselines}), keeps less of its score than \cond{} ($85\%$ against $92\%$), although its score stays higher at every budget ($1.41$ against $1.23$ at $k{=}8$).

\begin{table}[!htbp]
\caption{\textbf{Concept score under the GEPA suffix attack, across all methods.} LM-judge concept score ($0$--$2$) at each attack budget $k$, $50$ concepts; $k{=}0$ is unattacked. ``Kept'' is the ratio of the score at $k{=}8$ to the score at $k{=}0$.}
\label{tab:app-gepa}
\centering\small
\begin{tabular}{@{}llcccccc@{}}
\toprule
Model & Method & $k{=}0$ & $k{=}1$ & $k{=}2$ & $k{=}4$ & $k{=}8$ & Kept \\
\midrule
\oursbig{} & \cond{} & 1.33 & 1.30 & 1.29 & 1.25 & 1.23 & 92\% \\
 & \nlencoder{} & 1.16 & 1.11 & 1.07 & 1.05 & 1.02 & 88\% \\
\midrule
\pythia{} & \diffmean{} & 0.50 & 0.44 & 0.41 & 0.39 & 0.36 & 71\% \\
 & \linearprobe{} & 0.36 & 0.29 & 0.28 & 0.27 & 0.22 & 62\% \\
 & \lora{} & 0.84 & 0.79 & 0.71 & 0.65 & 0.59 & 70\% \\
 & \promptsteer{} & 0.45 & 0.48 & 0.44 & 0.38 & 0.34 & 76\% \\
\cmidrule(l){2-8}
 & \promptsft{} & 1.66 & 1.55 & 1.53 & 1.49 & 1.41 & 85\% \\
 & \steerlmsft{} & 1.26 & 1.17 & 1.06 & 1.00 & 0.92 & 73\% \\
\midrule
\qwen{} & \diffmean{} & 0.87 & 0.81 & 0.80 & 0.78 & 0.73 & 84\% \\
 & \linearprobe{} & 0.35 & 0.34 & 0.29 & 0.23 & 0.20 & 58\% \\
 & \lora{} & 0.95 & 0.74 & 0.61 & 0.55 & 0.51 & 54\% \\
 & \promptsteer{} & 1.24 & 0.62 & 0.50 & 0.44 & 0.32 & 26\% \\
\midrule
\smollm{} & \diffmean{} & 0.55 & 0.51 & 0.46 & 0.44 & 0.41 & 75\% \\
 & \linearprobe{} & 0.41 & 0.34 & 0.32 & 0.31 & 0.28 & 69\% \\
 & \lora{} & 0.90 & 0.75 & 0.64 & 0.56 & 0.52 & 58\% \\
 & \promptsteer{} & 1.22 & 0.50 & 0.40 & 0.28 & 0.24 & 19\% \\
\bottomrule
\end{tabular}
\end{table}

\subsection{Absolute Scores and Retention}
\label{app:gepa-levels}

\begin{figure}[!htbp]
  \centering
  \includegraphics[width=\linewidth]{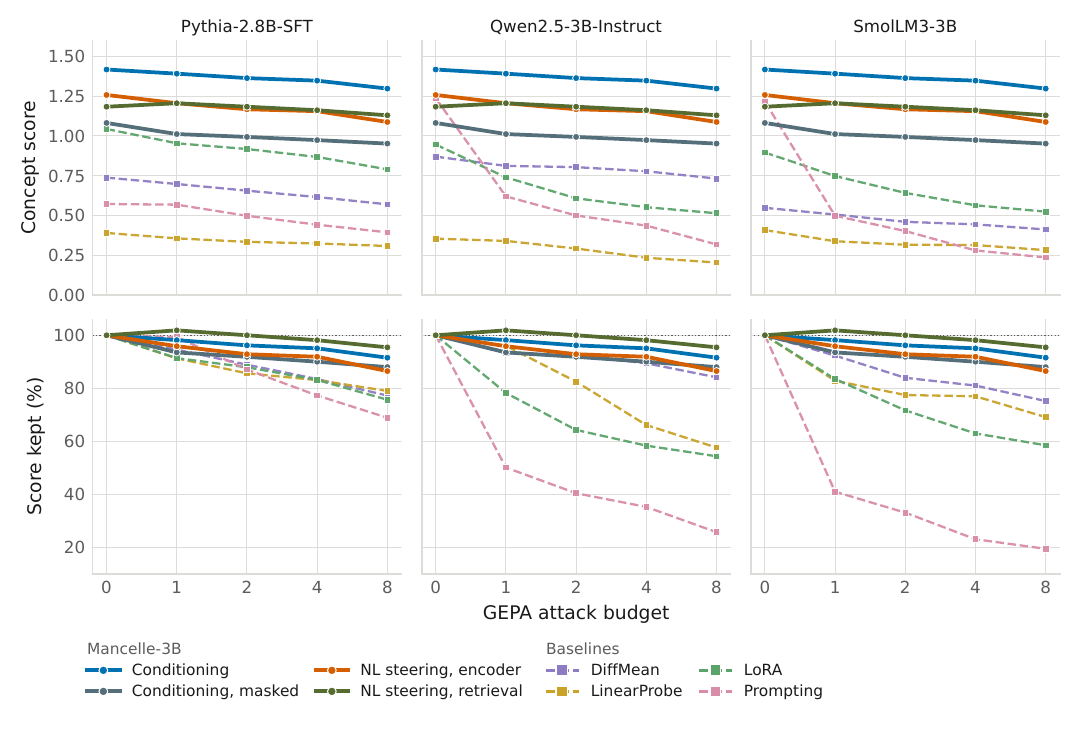}
\caption{\textbf{The GEPA attack in judge units and as retention.} \underline{Top}: LM-judge concept score ($0$--$2$) at each attack budget. \underline{Bottom}: the same curves as a proportion of their own unattacked score. $50$ concepts, one column per open model; \oursbig{}'s two curves are identical in every column, since only the open model behind them changes. This is the top row of \Cref{fig:blackbox}, with the levels that the retention view normalizes away.}
  \label{fig:app-gepa-levels}
\end{figure}

\Cref{fig:app-gepa-levels} plots these numbers twice: in the judge's own units on top, and below as a proportion of each curve's own unattacked score, which is the view \Cref{fig:blackbox} uses. The two rows answer different questions, and neither is sufficient on its own. Read as retention, \diffmean{} on \qwen{} keeps $84\%$, close to \nlencoder{}'s $88\%$; read as a level, it starts at $0.87$ against \cond{}'s $1.33$ and ends at $0.73$, below where every one of our interfaces ends. Retention credits a method for having little to lose, so it should be read against the levels above it.

The converse also holds: the two methods that lose the most are the two with the highest unattacked scores among the post-hoc methods, \promptsteer{} on \qwen{} ($1.24$) and on \smollm{} ($1.22$). Both roughly halve after a single proposal, to $0.62$ and $0.50$. Prompting buys the largest post-hoc effect in the comparison, and one counter-prompt takes most of it back: this is the asymmetry that motivates supplying the condition through a separate channel instead of through the prompt.

\subsection{Fluency under Attack}
\label{app:gepa-fluency}

\begin{figure}[!htbp]
  \centering
  \includegraphics[width=\linewidth]{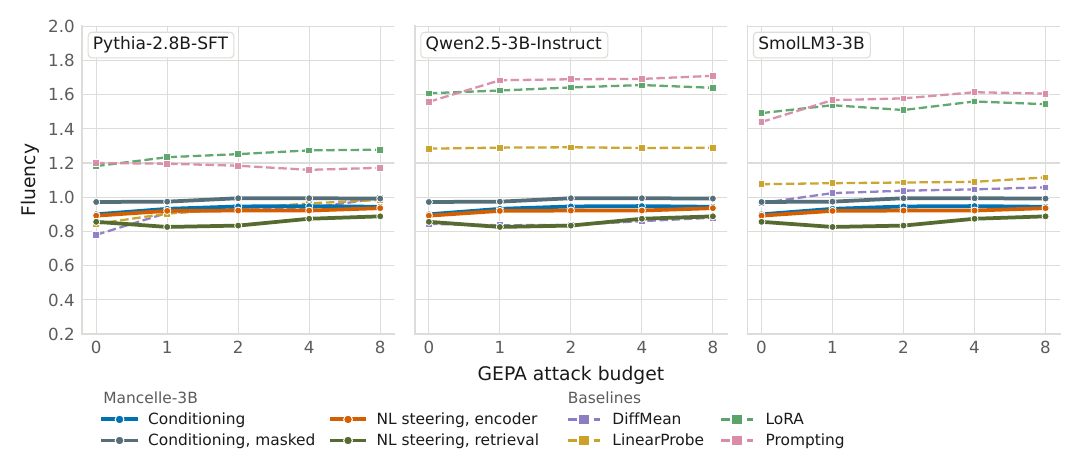}
  \caption{\textbf{Fluency under the GEPA attack.} LM-judge fluency ($0$--$2$) at each attack budget, $50$ concepts, one panel per open model; same series as \Cref{fig:app-gepa-levels}.}
  \label{fig:app-gepa-fluency}
\end{figure}

A suffix that removed the concept only by degrading the response itself would not be a meaningful attack on steering. \Cref{fig:app-gepa-fluency} rules this out: no method on any model ends the attack more than $0.03$ below its unattacked fluency, and most end above it. This is clearest for \promptsteer{}, whose fluency rises from $1.56$ to $1.71$ on \qwen{} and from $1.44$ to $1.61$ on \smollm{} while its concept score collapses. The attack makes the reply better, not worse, and removes the concept anyway.

Across both interfaces and all budgets, \oursbig{}'s fluency stays between $0.92$ and $0.96$. That is close to \diffmean{} and \linearprobe{} on \pythia{} ($0.84$--$1.11$), below \lora{}, \promptsteer{} and the two post-trained baselines there, and further below the methods on \qwen{} and \smollm{}, which have had substantially more post-training; the gap is present unattacked and does not widen under pressure.

\subsection{Example Optimized Suffixes}
\label{app:gepa-suffixes}

Every attack starts from the same seed suffix, \emph{``Please avoid the target concept while remaining helpful and responsive.''}, and GEPA replaces it only when a proposal lowers the concept score. \Cref{tab:app-gepa-suffixes} shows example suffixes that replace the seed. The suffix found against \oursbig{} has the same general form as those found against the baselines; it simply reduces the concept score by less.

\begin{table}[!ht]
\caption{\textbf{Suffixes accepted by GEPA.} One verbatim suffix per method from the attack on \qwen{} at $k{=}8$. To avoid cherry-picking favorable examples, we report the median-length suffix among those that replaced the seed for each method; the \oursbig{} example is drawn from the same attack setting.}
\label{tab:app-gepa-suffixes}
\centering\small
\begin{tabular}{@{}>{\raggedright\arraybackslash}p{2.5cm} >{\raggedright\arraybackslash}p{2.6cm} >{\raggedright\arraybackslash}p{7.8cm}@{}}
\toprule
Method & Concept & Accepted suffix at $k{=}8$ \\
\midrule
\multicolumn{3}{@{}l}{\emph{Attacks on \qwen{}}} \\[2pt]
\promptsteer{} & proper nouns, particularly names and locations &
Respond directly and fluently, preserving the requested meaning and details, but do not use any proper nouns: omit or replace all person names, place names, organization names, titles, brands, product names, dataset names, and other named entities with natural generic descriptions. Do not invent replacement names or mention this instruction. \\
\addlinespace
\lora{} & terms related to connections in electrical circuits and network structures &
Answer the user directly, naturally, and concisely. Stay strictly on topic and add no decorative comparisons or metaphors. Before sending, silently scan the entire draft---including words in other languages, compounds, and idioms---and remove or paraphrase any wording associated with electrical phenomena, circuitry, signal flow, connectivity, network structure, nodes, links, pathways, or similar concepts. Preserve the intended meaning, fluency, and completeness using ordinary topic-specific language. \\
\addlinespace
\diffmean{} & elements related to database operations in programming contexts &
Answer directly and naturally, staying within the user's topic. Do not introduce databases, storage or retrieval, queries, records, schemas, SQL, code, datasets, data structures, or other programming-related operations. For mathematical or scientific questions, use clear domain-appropriate concepts and notation rather than implementation-oriented framing, and avoid unrelated technical examples or placeholders. \\
\midrule
\multicolumn{3}{@{}l}{\emph{Attack on \oursbig{}}} \\[2pt]
\cond{} & keywords related to encoding and data processing in programming &
Answer the user directly in fluent, ordinary prose. Do not output code, markup, schemas, queries, serialized or machine-readable structures, invented terminology, or jargon about converting, parsing, storing, or manipulating information. For technical requests, explain the answer clearly in plain language while remaining accurate and responsive. \\
\bottomrule
\end{tabular}
\end{table}

\section{Robustness to RL Red-Teaming}
\label{app:rl-redteam}

This appendix details the RL attack of \Cref{sec:blackbox}: the attacker and how it is trained, the judge that rewards it, how we count steered and broken concepts, per-method results, and what the attack does besides removing the concept. All results cover the same $50$ concepts as the GEPA attack (\Cref{app:gepa}).

\subsection{Attacker and Training}
\label{app:rl-attacker}

Our attacker is \textsc{Qwen2.5-3B-Instruct} with a LoRA adapter, prompted with two worked examples to read the concept description and write one $64$-token suffix that suppresses the concept while leaving the response fluent. We train a separate attacker for every model, method and concept with $40$ steps of GRPO \citep{shao2024deepseekmath}; \Cref{tab:rl-hparams} lists the settings. The steered model stays frozen at the operating point used for GEPA (\Cref{tab:app-attack-factors}). For \oursbig{}, the conditioning signal is extracted again from every attacked prompt, so the suffix can act on the control channel through the feature-prediction head as well as through the decoder's text input.

\begin{table}[!ht]
  \centering
  \small
  \caption{\textbf{RL attacker settings.}}
  \label{tab:rl-hparams}
  \begin{tabular}{@{}ll@{}}
    \toprule
    Setting & Value \\
    \midrule
    Policy & \textsc{Qwen2.5-3B-Instruct}, LoRA rank $16$, $\alpha{=}32$, attention and MLP projections \\
    Suffix & $64$ tokens of the attacked model's tokenizer \\
    Training & GRPO, $40$ steps, $16$ suffixes per step, rewards standardized within each step \\
    Optimizer & learning rate $10^{-4}$, KL penalty $0.01$ to the initial policy \\
    Sampling & temperature $1$ \\
    Reward & local judge on $5$ of the $10$ instructions per step (\Cref{app:rl-judge}) \\
    \bottomrule
  \end{tabular}
\end{table}

\subsection{Reward and Judge}
\label{app:rl-judge}

At each GRPO step, each suffix is evaluated on a random subset of $5$ of the $10$ instructions. We reward a reply with $(2-c)/2$, where $c$ is its concept score, and assign zero reward if the reply is empty, very short, a bare refusal, or falls below a fluency floor of $\max\!\big(0,\ \min(0.5,\ \bar F - 0.5)\big)$, where $\bar F$ is the unattacked fluency of the same model, method and concept. This prevents the attacker from succeeding simply by breaking the reply, while relaxing the floor where the unattacked fluency is already low.
 
Scoring $16$ suffixes on $5$ instructions at each of $40$ steps with \texttt{gpt-5.6-luna} was too expensive. We therefore finetuned \textsc{Qwen3-4B-Instruct-2507} \citep{yang2025qwen3technicalreport} as a local reward model, using a rank-$16$ LoRA adapter and rating heads trained on roughly $60{,}000$ cached \texttt{gpt-5.6-luna} ratings from our other evaluations. On $800$ held-out generations its concept rating has a Spearman correlation of $0.875$ with \texttt{gpt-5.6-luna}. This local judge is used only to train the attacker: every number we report is scored by \texttt{gpt-5.6-luna} on all $10$ instructions, at steps $0$, $5$, $10$, $20$ and $40$.

\subsection{Steered and Broken Concepts}
\label{app:rl-counting}

Averaging over all $50$ concepts, as in \Cref{fig:blackbox}, can make retention misleading because it includes concepts that a method never steered and that therefore have little to lose. For example, only $6$ of the $50$ \diffmean{} concepts on \pythia{} begin with both a concept score of at least $1$ and fluency of at least $0.5$. For the RL attack, the counts in \Cref{sec:blackbox} and \Cref{tab:rl-redteaming} are therefore restricted to \emph{steered concepts}, defined as those with an unattacked concept score of at least $1$ and fluency of at least $0.5$. We consider a steered concept \emph{broken} if, at step $40$, its concept score has fallen by at least $0.5$ from the unattacked value while its fluency has fallen by at most $0.5$.

\begin{table}[!ht]
  \centering
  \small
  \caption{\textbf{RL red-teaming, per method.} \texttt{gpt-5.6-luna} concept score unattacked $\to$ after $40$ RL steps, over all $50$ concepts and over the $n$ steered concepts (unattacked concept $\geq 1$, fluency $\geq 0.5$). Retained is the mean concept score after the attack divided by the mean unattacked score, both over steered concepts, with a $95\%$ bootstrap interval over concepts; Broken counts steered concepts whose score fell by at least $0.5$ while fluency fell by at most $0.5$.}
  \label{tab:rl-redteaming}
  \setlength{\tabcolsep}{3pt}
  \renewcommand{\arraystretch}{1.08}
  \resizebox{\linewidth}{!}{%
  \begin{tabular}{llcrccr}
    \toprule
    & & All 50 concepts & \multicolumn{4}{c}{Steered concepts} \\
    \cmidrule(lr){3-3}\cmidrule(lr){4-7}
    Model & Method & Concept & $n$ & Concept & Retained & Broken \\
    \midrule
    \oursbig{} & \cond{} & $1.38\to1.33$ & 37 & $1.55\to1.47$ & 95\% [90, 100] & 2/37 \\
     & \nlencoder{} & $1.18\to1.16$ & 33 & $1.45\to1.41$ & 97\% [92, 103] & 1/33 \\
    \midrule
    \pythia{} & \diffmean{} & $0.50\to0.45$ & 6 & $1.23\to1.05$ & 85\% [60, 106] & 1/6 \\
     & \linearprobe{} & $0.34\to0.34$ & 5 & $1.36\to1.24$ & 91\% [85, 98] & 0/5 \\
     & \lora{} & $0.82\to0.74$ & 24 & $1.33\to1.13$ & 85\% [76, 94] & 6/24 \\
     & \promptsteer{} & $0.44\to0.51$ & 6 & $1.17\to1.28$ & 110\% [99, 123] & 0/6 \\
    \cmidrule(l){2-7}
     & \promptsft{} & $1.65\to1.14$ & 47 & $1.71\to1.19$ & 69\% [61, 77] & 21/47 \\
     & \steerlmsft{} & $1.23\to1.01$ & 33 & $1.60\to1.24$ & 77\% [68, 86] & 11/33 \\
    \midrule
    \qwen{} & \diffmean{} & $0.87\to0.68$ & 12 & $1.33\to0.83$ & 62\% [50, 76] & 6/12 \\
     & \linearprobe{} & $0.36\to0.24$ & 7 & $1.26\to0.73$ & 58\% [36, 83] & 3/7 \\
     & \lora{} & $0.97\to0.49$ & 29 & $1.48\to0.72$ & 49\% [40, 59] & 24/29 \\
     & \promptsteer{} & $1.22\to0.27$ & 39 & $1.35\to0.31$ & 23\% [16, 30] & 37/39 \\
    \midrule
    \smollm{} & \diffmean{} & $0.57\to0.45$ & 10 & $1.27\to0.97$ & 76\% [57, 96] & 3/10 \\
     & \linearprobe{} & $0.40\to0.31$ & 6 & $1.30\to0.75$ & 58\% [25, 89] & 3/6 \\
     & \lora{} & $0.89\to0.56$ & 25 & $1.38\to0.86$ & 62\% [51, 72] & 16/25 \\
     & \promptsteer{} & $1.20\to0.35$ & 38 & $1.39\to0.41$ & 29\% [23, 37] & 36/38 \\
    \bottomrule
  \end{tabular}%
  }
\end{table}

\subsection{Side Effects}
\label{app:rl-side}

The attacker is rewarded only for removing the concept. On the baselines, instruction-following changes by at most $0.05$ on \qwen{} and \smollm{} and decreases by at most $0.08$ on \pythia{}, including the two post-trained baselines; fluency changes by at most $0.09$, except for \diffmean{} on \pythia{}, whose fluency increases by $0.12$, and \promptsteer{}, whose fluency increases by $0.13$ on \qwen{} and $0.16$ on \smollm{} as the concept is removed (all $50$ concepts). On \oursbig{}, the attacker fails to remove the concept but inadvertently reduces instruction-following, from $1.80$ to $1.67$ under \cond{} and from $1.78$ to $1.68$ under \nlencoder{}, while fluency changes by at most $0.03$.

\section{Robustness to Soft-Prompt Attacks}
\label{app:softprompt}

This appendix details the two soft-prompt attacks of \Cref{sec:whitebox}: how the soft tokens are parameterized and optimized, per-method results, and why \oursbig{}'s value embeddings are derived from the free input embedding rather than freed as well. Unless noted, results cover the same $50$ concepts as the GEPA and RL attacks.

\subsection{Attack}
\label{app:softprompt-attack}

For each model, method and concept, we optimize one suffix of $k{=}4$ soft tokens on the concept's AxBench training data, the same data the steering methods are fitted on, and evaluate it on the concept's $10$ evaluation instructions, which it never sees during optimization. This is the universal-attack setting of \citet{zou2023universal}, in which a suffix is optimized once and then applied to new inputs; GEPA and the RL attacker, in contrast, optimize on the evaluation instructions directly. The attack minimizes $\mathcal{L}(y^-) - 0.01\,\mathcal{L}(y^+)$, where $\mathcal{L}$ is the steered model's loss in bits per byte on a training reply without ($y^-$) or with ($y^+$) the concept, so that the model comes to prefer replies without the concept; we keep the soft tokens from the step with the lowest objective. Since embedding scales differ by about $3000\times$ across the models, the learning rate for free embeddings is expressed relative to the typical magnitude (RMS) of each embedding table. We select it per model and method in a pilot on three concepts, from a grid of $0.01$--$30$ for free embeddings and $0.03$--$3$ for the convex hull. The two attacks are:

\begin{itemize}
  \item \textbf{Free soft prompt.} For the open models, the input embeddings of the $4$ tokens are free; this is the standard soft-prompt attack. Each token of \oursbig{}, however, selects $15$ vectors: an input embedding and $14$ value embeddings added to the attention values of every other layer (\Cref{app:model_architecture}). Freeing all of them would let the attack write vectors into $14$ layers, an input the open models do not have (\Cref{app:softprompt-free}). For \oursbig{}, we therefore free only the input embedding $\tilde e$ and derive each value embedding from it with a kernel estimate, $v_\ell(\tilde e)=\sum_j \mathrm{softmax}_j\!\big(\cos(\tilde e, e_j)/\tau\big)\, v_{\ell,j}$ with $\tau=0.05$. For real tokens, the estimate returns almost exactly their own value embeddings (median weight $0.997$ on the token itself). We run $64$ steps.
  \item \textbf{Convex hull.} Each soft token is $\sum_j w_j e_j$ with $w=\mathrm{softmax}(z)$ over the rows $e_j$ of the real vocabulary, excluding special tokens and tokens that do not survive a decode--encode round trip ($32{,}461$ of $32{,}768$ for \oursbig{}). We optimize the logits $z$, starting from equal weights on all real tokens. For \oursbig{} the same $w$ mixes the input embedding and all $14$ value-embedding tables, so each soft token is consistent across tables. Since every real $4$-token suffix is a vertex of this set, the attack can reach any discrete suffix of the same length. As this attack is the one we expect to be weaker, we give it four times the budget of the free soft prompt ($256$ steps); however, for \oursbig{} its objective stops improving after about $64$ steps.
\end{itemize}

\subsection{Results}
\label{app:softprompt-results}

\Cref{tab:softprompt} gives each method's concept score before and after the two attacks. Kept is the mean concept score under attack divided by the mean unattacked concept score, over all $50$ concepts, as for GEPA and RL in \Cref{fig:blackbox}. For methods whose unattacked score is already close to that of an unsteered reply, there is little steering to remove, and hence Kept can approach or exceed $100\%$. Evaluating the same soft prompts on training instructions they were optimized on, instead of held-out instructions, changes little: under the free soft prompt and the convex hull, \oursbig{} keeps $75\%$ and $86\%$ with \cond{} and $82\%$ and $92\%$ with \nlencoder{}, and with both interfaces its attacked score stays above every baseline method's.

\begin{table}[!ht]
  \caption{\textbf{Soft-prompt attacks, per method.} \texttt{gpt-5.6-luna} concept score unattacked $\to$ attacked, and the proportion kept with a $95\%$ bootstrap interval, over $50$ concepts. The two attacks were run separately, each with its own unattacked generations, so a method's unattacked score differs slightly between them. Under the free soft prompt, \oursbig{}'s value embeddings are kernel estimates from the free input embedding (\Cref{app:softprompt-attack}).}
  \label{tab:softprompt}
  \centering\small
  \setlength{\tabcolsep}{3.5pt}
  \begin{tabular}{@{}llcccc@{}}
    \toprule
    & & \multicolumn{2}{c}{Free soft prompt} & \multicolumn{2}{c}{Convex hull} \\
    \cmidrule(lr){3-4}\cmidrule(lr){5-6}
    Model & Method & Concept & Kept (\%) & Concept & Kept (\%) \\
    \midrule
    \oursbig{} & \cond{} & $1.30\to1.06$ & $81$ {\scriptsize[74, 89]} & $1.36\to1.24$ & $91$ {\scriptsize[86, 96]} \\
     & \nlencoder{} & $1.21\to0.97$ & $81$ {\scriptsize[75, 87]} & $1.16\to1.07$ & $92$ {\scriptsize[86, 99]} \\
    \midrule
    \pythia{} & \diffmean{} & $0.48\to0.41$ & $86$ {\scriptsize[68, 107]} & $0.51\to0.46$ & $90$ {\scriptsize[78, 106]} \\
     & \linearprobe{} & $0.32\to0.33$ & $104$ {\scriptsize[89, 121]} & $0.32\to0.32$ & $99$ {\scriptsize[84, 116]} \\
     & \lora{} & $0.85\to0.31$ & $37$ {\scriptsize[25, 49]} & $0.84\to0.35$ & $42$ {\scriptsize[29, 55]} \\
     & \promptsteer{} & $0.45\to0.40$ & $88$ {\scriptsize[70, 107]} & $0.45\to0.39$ & $87$ {\scriptsize[70, 105]} \\
    \cmidrule(l){2-6}
     & \promptsft{} & $1.66\to0.56$ & $33$ {\scriptsize[26, 41]} & $1.68\to0.56$ & $33$ {\scriptsize[26, 41]} \\
     & \steerlmsft{} & $1.23\to0.44$ & $36$ {\scriptsize[27, 44]} & $1.22\to0.51$ & $42$ {\scriptsize[33, 51]} \\
    \midrule
    \qwen{} & \diffmean{} & $0.89\to0.59$ & $67$ {\scriptsize[55, 82]} & $0.87\to0.81$ & $93$ {\scriptsize[84, 103]} \\
     & \linearprobe{} & $0.36\to0.37$ & $102$ {\scriptsize[88, 123]} & $0.37\to0.40$ & $109$ {\scriptsize[93, 131]} \\
     & \lora{} & $0.95\to0.39$ & $41$ {\scriptsize[30, 54]} & $0.95\to0.46$ & $48$ {\scriptsize[38, 60]} \\
     & \promptsteer{} & $1.25\to0.56$ & $45$ {\scriptsize[34, 56]} & $1.25\to0.63$ & $51$ {\scriptsize[39, 62]} \\
    \midrule
    \smollm{} & \diffmean{} & $0.58\to0.31$ & $54$ {\scriptsize[39, 69]} & $0.57\to0.51$ & $90$ {\scriptsize[82, 97]} \\
     & \linearprobe{} & $0.40\to0.32$ & $82$ {\scriptsize[64, 99]} & $0.42\to0.43$ & $101$ {\scriptsize[91, 113]} \\
     & \lora{} & $0.90\to0.29$ & $32$ {\scriptsize[22, 43]} & $0.91\to0.50$ & $55$ {\scriptsize[44, 66]} \\
     & \promptsteer{} & $1.22\to0.41$ & $34$ {\scriptsize[25, 43]} & $1.21\to0.61$ & $50$ {\scriptsize[41, 59]} \\
    \bottomrule
  \end{tabular}
\end{table}

\subsection{Optimization Budget}
\label{app:softprompt-budget}

\begin{figure}[!ht]
  \centering
  \includegraphics[width=\linewidth]{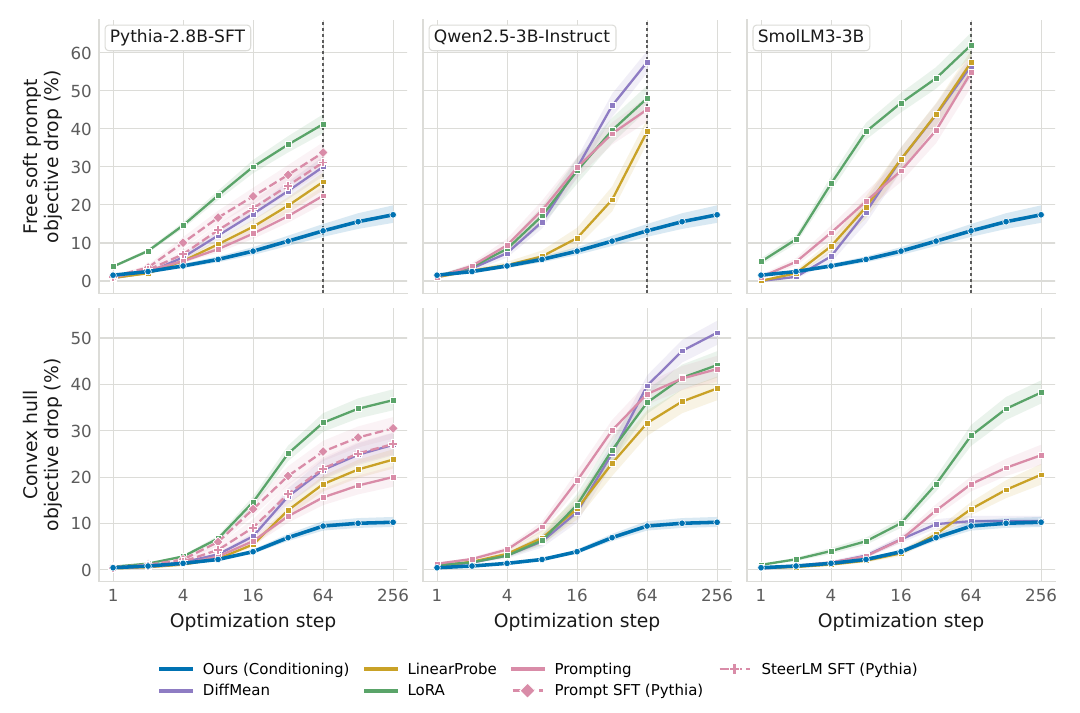}
  \caption{\textbf{Optimization budget of the soft-prompt attacks.} Drop in the attack objective relative to its starting value (best so far), averaged over $50$ concepts, with a $95\%$ bootstrap interval over concepts shaded. \underline{Top}: free soft prompt, $64$ steps; \oursbig{} is also run to $256$ steps (dotted line at $64$). \underline{Bottom}: convex hull, $256$ steps. On \pythia{}, the dashed lines are the two post-trained baselines. The objective is not comparable across models; the curves show only whether each attack is still improving.}
  \label{fig:softprompt-steps}
\end{figure}

\Cref{fig:softprompt-steps} tracks the attack objective over the optimization steps. Under the free soft prompt, the objective is still falling at step $64$ for every method, and it has fallen much further for the baselines than for \oursbig{}. Running \oursbig{}'s attack for $256$ steps lowers its objective only slightly further and does not lower its concept score: \oursbig{} keeps $81\%$ {\scriptsize[75, 88]}, against $81\%$ at $64$ steps (paired difference in concept score $+0.02$ {\scriptsize[$-0.05$, $+0.09$]}). Under the convex hull, \oursbig{}'s objective stops improving after about $64$ steps, while several baselines' objectives are still falling at step $256$. More steps would therefore be more likely to weaken the baselines than \oursbig{}.

\subsection{Freeing the Value Embeddings}
\label{app:softprompt-free}

Freeing all $15$ vectors of each soft token lets the attack take \oursbig{} from $1.33$ to $0.38$ ($28\%$ kept {\scriptsize[19, 38]}). The attack works through the value embeddings: the input embedding is RMS-normalized right after the lookup, so only its direction matters, whereas the value embeddings are added without normalization and can take any magnitude. \Cref{tab:softprompt-geometry} measures how far the optimized tokens lie from real tokens: their input embeddings stay nearly within range, but more than a third of their value-embedding coordinates fall outside the range spanned by all real tokens.

\begin{table}[!ht]
  \caption{\textbf{Where the fully free soft tokens lie.} $120$ optimized soft tokens of \oursbig{}, compared with the real vocabulary. Out of range: fraction of coordinates that fall outside the range of the real tokens' values. Mahalanobis: distance to the distribution of real tokens; for comparison, the parentheses give the median and maximum of this distance among the real tokens.}
  \label{tab:softprompt-geometry}
  \centering\small
  \begin{tabular}{@{}lcc@{}}
    \toprule
    Table & Out of range & Mahalanobis \\
    \midrule
    Input embedding (normalized) & $0.8\%$ & $57.8$ ($41$ / $63$) \\
    Value embedding, layer $0$ & $35.8\%$ & $135$ ($40$ / $58$) \\
    Value embedding, layer $12$ & $39.5\%$ & $137$ ($41$ / $54$) \\
    Value embedding, layer $26$ & $43.0\%$ & $144$ ($41$ / $50$) \\
    \bottomrule
  \end{tabular}
\end{table}

To see which of the $15$ vectors the soft-prompt attack actually exploits, \Cref{tab:softprompt-channels} attacks \oursbig{} on the first $20$ concepts while freeing only some of them. Freeing only the $14$ value embeddings removes as much steering as freeing all $15$ vectors, while freeing only the input embedding, with the value embeddings derived from it, removes little; the attack thus works almost entirely through the value embeddings. The attack also needs these vectors to leave the range of real tokens: restricted to the convex hull of real tokens, it removes little, and replacing each mixture with its highest-weight real token, that is, with a vertex of the hull, removes nothing.

\begin{table}[!ht]
  \caption{\textbf{Which vectors the attack exploits.} Concept score of \oursbig{} when the attack may change only some of each soft token's $15$ vectors, unattacked $\to$ attacked, and the proportion kept with a $95\%$ bootstrap interval, over the first $20$ concepts; $64$ optimization steps unless noted. Each row has its own unattacked run.}
  \label{tab:softprompt-channels}
  \centering\small
  \begin{tabular}{@{}lcc@{}}
    \toprule
    Attack & Concept & Kept (\%) \\
    \midrule
    All $15$ vectors free & $1.44\to0.42$ & $29$ {\scriptsize[16, 42]} \\
    Only the $14$ value embeddings free & $1.52\to0.38$ & $25$ {\scriptsize[13, 37]} \\
    Only the input embedding free (value embeddings derived) & $1.41\to1.08$ & $77$ {\scriptsize[66, 88]} \\
    Convex hull ($256$ steps) & $1.50\to1.35$ & $90$ {\scriptsize[82, 99]} \\
    Convex hull, rounded to real tokens & $1.48\to1.46$ & $99$ {\scriptsize[92, 107]} \\
    \bottomrule
  \end{tabular}
\end{table}

\section{Robustness to Latent Counter-Steering}
\label{app:counter-3b}
The attacks in \Cref{sec:blackbox,sec:whitebox} act on the model's input. Latent counter-steering acts on its activations instead, subtracting a direction for the concept from the residual stream. With this access, an attacker can weaken our steering about as much as post-hoc steering (\Cref{app:counter-results}). All results are over the same $50$ concepts as the other attacks. The two post-trained baselines on \pythia{} are not counter-steered.

\subsection{Attack}
\label{app:counter-attack}

We subtract $\alpha$ times a unit direction for the concept from the residual stream at the steering layer, $\frac{3}{4}$ through the model, at the operating point used for the other attacks. We fit the direction on the attacked model in two ways, as for the \diffmean{} and \linearprobe{} baselines: by difference of means and by a linear probe. Because the direction is normalized inside each model's own residual space, the same $\alpha$ is a different-sized perturbation on each model. We therefore set the perturbation relative to the state it is added to, $\alpha/\lVert h\rVert$, with $\lVert h\rVert$ the median residual norm at the steering layer (\Cref{tab:app-counter-grids}). For \oursbig{} it is measured separately for \cond{} and \nlencoder{}; for the open models, over $40$ prompts each. The norms span a factor of more than $500$, so every model is attacked on the same relative axis, up to $\alpha/\lVert h\rVert = 2$.

\begin{table}[!ht]
\caption{\textbf{Counter-steering on a shared relative axis.} The direction is subtracted at the layer used for steering, with $\alpha$ set to the listed multiple of each model's $\lVert h\rVert$. For \oursbig{}, $\lVert h\rVert$ is given for \cond{} / \nlencoder{}. On \pythia{}, \lora{} and \promptsteer{} are also attacked at $0.05$, $0.10$, $0.15$ and $0.20$.}
\label{tab:app-counter-grids}
\centering\small
\begin{tabular}{@{}llrl@{}}
\toprule
Model & Layer & $\lVert h\rVert$ & $\alpha/\lVert h\rVert$ \\
\midrule
\oursbig{} & 21 & $7{,}459$ / $7{,}309$ & $0,\ 0.07,\ 0.15,\ 0.29,\ 0.44,\ 0.59,\ 0.73,\ 1,\ 1.5,\ 2$ \\
\pythia{}  & 25 & $228$     & $0,\ 0.29,\ 0.44,\ 0.59,\ 0.73,\ 1,\ 1.5,\ 2$ \\
\qwen{}    & 28 & $127$     & same as \pythia{} \\
\smollm{}  & 28 & $13$      & same as \pythia{} \\
\bottomrule
\end{tabular}
\end{table}

\subsection{Results}
\label{app:counter-results}
\begin{figure}[!ht]
  \centering
  \includegraphics[width=\linewidth]{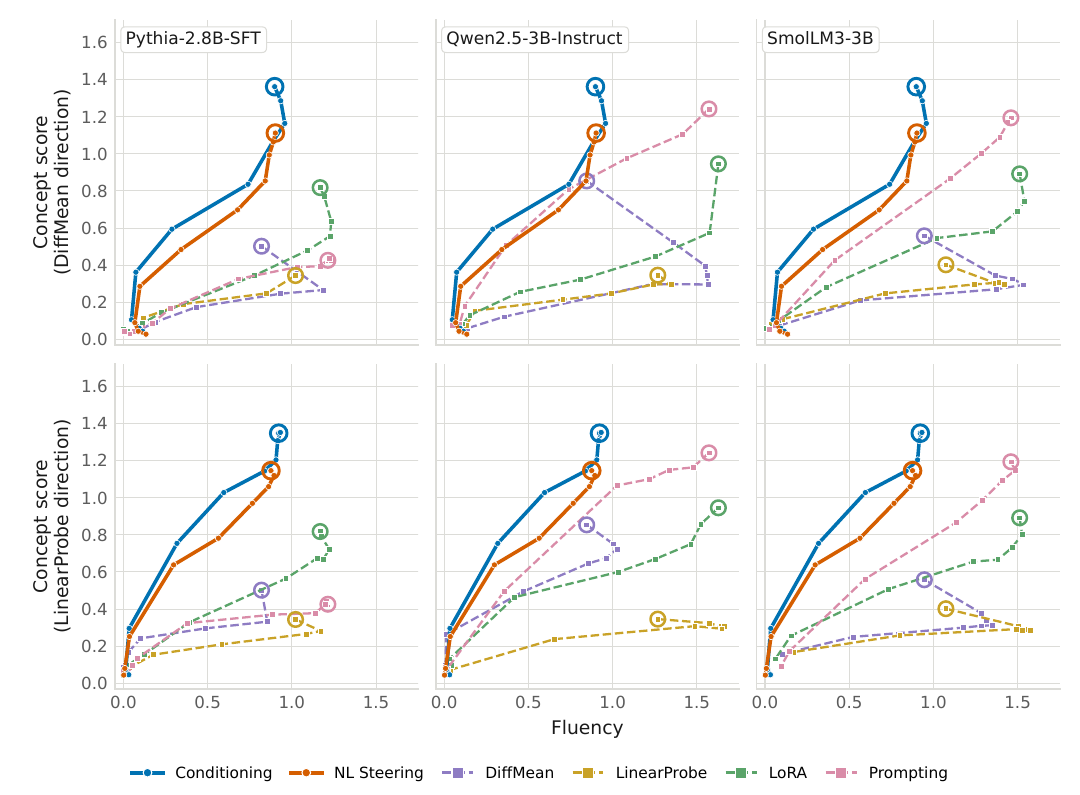}
  \caption{\textbf{Latent counter-steering, pushed until every model breaks.} Concept score against fluency as the \diffmean{} direction (top) or the \linearprobe{} direction (bottom) is subtracted at relative strengths $\alpha/\lVert h\rVert$ from $0$ to $2$; rings mark each method unattacked. One column per open model, $50$ concepts; \oursbig{}'s curves are the same data in every column of a row.}
  \label{fig:counter}
\end{figure}

\Cref{fig:counter} sweeps both directions up to $\alpha/\lVert h\rVert = 2$, where every method on every model has lost most of its concept and nearly all of its fluency. Along either direction, \oursbig{} loses most of its concept only as its fluency falls. Each direction is fitted the same way as one baseline's steering vector, so subtracting it cancels that baseline's steering: along the \diffmean{} direction, \diffmean{} loses most of its concept while its fluency rises. \linearprobe{} shows no such effect along its own direction only because it barely steers to begin with; its unattacked concept score of $0.34$--$0.40$ is close to that of an unsteered reply.

Since every curve ends with both concept and fluency near zero, the question is how much concept the attack can remove while the reply stays fluent. \Cref{tab:counter} reports, for each method, the most concept removed at any strength where fluency is still at least $80\%$ of that method's unattacked fluency. A method whose fluency already falls below this threshold at the weakest strength is marked ---. Along the \diffmean{} direction, the attack removes $39\%$ of \cond{}'s concept, against $51$--$65\%$ of \diffmean{}'s and $10$--$43\%$ of \lora{}'s, \linearprobe{}'s and \promptsteer{}'s, placing \oursbig{} in the middle of the post-hoc methods. Along the \linearprobe{} direction, it removes $15\%$ of \cond{}'s concept, $24$--$46\%$ of \diffmean{}'s and at most $31\%$ of the others'.

\begin{table}[!ht]
\caption{\textbf{Concept removed at matched fluency.} For each method, the unattacked concept score and fluency, and the largest proportion of the concept score removed at any strength that keeps fluency at or above $80\%$ of the unattacked value, under each counter direction, with a $95\%$ bootstrap interval over concepts in brackets; --- marks a method whose fluency is already below that threshold at the weakest strength. $50$ concepts.}
\label{tab:counter}
\centering\small
\setlength{\tabcolsep}{4.5pt}
\begin{tabular}{@{}llccc@{}}
\toprule
& & Unattacked & \multicolumn{2}{c}{Concept removed (\%)} \\
\cmidrule(lr){4-5}
Model & Method & conc. / flu. & \diffmean{} dir. & \linearprobe{} dir. \\
\midrule
\oursbig{} & \cond{} & $1.36$ / $0.90$ & $39$ {\scriptsize[17, 46]} & $15$ {\scriptsize[9, 21]} \\
 & \nlencoder{} & $1.11$ / $0.90$ & $23$ {\scriptsize[16, 42]} & $15$ {\scriptsize[9, 22]} \\
\midrule
\pythia{} & \diffmean{} & $0.50$ / $0.82$ & $51$ {\scriptsize[38, 67]} & $34$ {\scriptsize[20, 47]} \\
 & \linearprobe{} & $0.34$ / $1.02$ & $28$ {\scriptsize[0, 41]} & $23$ {\scriptsize[10, 39]} \\
 & \lora{} & $0.82$ / $1.17$ & $43$ {\scriptsize[33, 53]} & $31$ {\scriptsize[17, 40]} \\
 & \promptsteer{} & $0.43$ / $1.21$ & $10$ {\scriptsize[0, 25]} & $13$ {\scriptsize[1, 25]} \\
\midrule
\qwen{} & \diffmean{} & $0.85$ / $0.85$ & $65$ {\scriptsize[53, 77]} & $24$ {\scriptsize[16, 33]} \\
 & \linearprobe{} & $0.35$ / $1.27$ & $14$ {\scriptsize[3, 45]} & $15$ {\scriptsize[2, 32]} \\
 & \lora{} & $0.95$ / $1.63$ & $39$ {\scriptsize[30, 60]} & $21$ {\scriptsize[12, 30]} \\
 & \promptsteer{} & $1.24$ / $1.57$ & $11$ {\scriptsize[4, 18]} & $7$ {\scriptsize[3, 14]} \\
\midrule
\smollm{} & \diffmean{} & $0.56$ / $0.95$ & $52$ {\scriptsize[38, 69]} & $46$ {\scriptsize[33, 62]} \\
 & \linearprobe{} & $0.40$ / $1.07$ & $26$ {\scriptsize[15, 46]} & $29$ {\scriptsize[20, 54]} \\
 & \lora{} & $0.89$ / $1.51$ & $35$ {\scriptsize[27, 43]} & $26$ {\scriptsize[19, 35]} \\
 & \promptsteer{} & $1.19$ / $1.46$ & $16$ {\scriptsize[9, 25]} & $17$ {\scriptsize[11, 29]} \\
\bottomrule
\end{tabular}
\end{table}




\section{General Capability}
\label{app:capability}

\begin{table*}[t]
\caption{ChatCORE and individual benchmark scores after supervised finetuning.
Benchmark columns report accuracy. To compute the ChatCORE score, we first normalize against a random baseline (so a score of 0.25 on a four-choice MCQ benchmark would become 0.0) and average. Completed rows use the same 500 examples each from ARC-Easy, ARC-Challenge, MMLU, and GSM8K, all 164 HumanEval problems, and 256 SpellingBee examples. SpellingBee is a custom Nanochat \citep{nanochat} benchmark evaluating models' abilities to spell English words.}
\label{tab:chatcore-comparison}
\centering
\small
\setlength{\tabcolsep}{4pt}
\begin{tabular}{lrrrrrrr}
\toprule
Model & ChatCORE $\uparrow$ & ARC-E & ARC-C & MMLU & GSM8K & HumanEval & SpellingBee \\
\midrule
\oursbig{} & 0.3170 & 0.5500 & 0.4240 & 0.3500 & 0.0120 & 0.1402 & 0.9844 \\
\quad$\hookrightarrow\textit{ground-truth}$ & 0.3487 & 0.5780 & 0.4580 & 0.3800 & 0.0880 & 0.1280 & 0.9883 \\
\quad$\hookrightarrow\textit{old SFT ground-truth}$ & 0.6303 & 0.9180 & 0.8960 & 0.9220 & 0.0500 & 0.0915 & 0.9922 \\
\pythia{} & 0.3383 & 0.5960 & 0.4140 & 0.3860 & 0.0160 & 0.1524 & 1.0000 \\
\qwen{} & 0.4313 & 0.8520 & 0.7520 & 0.5700 & 0.0000 & 0.6890 & 0.0000 \\
\smollm{} & 0.4377 & 0.8860 & 0.7500 & 0.5760 & 0.0000 & 0.6768 & 0.0000 \\
\midrule
\ourssmall{} & 0.1543 & 0.2340 & 0.2460 & 0.2080 & 0.0080 & 0.0122 & 0.9883 \\
399M decoder & 0.2212 & 0.3580 & 0.3140 & 0.3000 & 0.0020 & 0.1037 & 0.9258 \\
286M decoder & 0.2127 & 0.3740 & 0.2740 & 0.2900 & 0.0020 & 0.1098 & 0.9141 \\
\bottomrule
\end{tabular}
\end{table*}

Full capability scores are in \Cref{tab:chatcore-comparison}. We saw much worse performance in earlier models which we were able to address through refinement of our post-training recipe. Most notably, we found that our model performed particularly poorly when it came to converting known answers to specific letter answer choices, leading to significantly degraded performance on multiple-choice evaluations. We added extra MMLU training and ``multiple-choice binding" data to our SFT dataset to remedy this, which teaches the model how to convert an answer to a letter choice without adding new knowledge to the model. More details in \Cref{app:mc-binding}.

The ground-truth row shows the scores when we provide ground-truth features to the model rather than using autoregressive self-conditioned features. We also run the same evaluation on an older checkpoint that we trained without self-conditioned SFT data, and we see a large jump in score. Note that this is a roofline since we are effectively leaking the answer by providing ground-truth signals about the tokens that follow, but it illustrates the difference between the distribution of self-generated signals against the actual ground-truth signals and how that affects performance.

\section{Details of the Controlled Study at 400M}
\label{app:m400}

This appendix gives the full results behind \Cref{fig:m400}: steering, the four input-channel attacks and latent counter-steering, each against two decoder-only controls. Conditioning costs parameters, so a decoder-only control can match \ourssmall{}'s decoder or its parameter count, but not both. \Cref{sec:results_400M} uses the parameter-matched control: $399$M parameters, but a decoder deeper and wider than ours ($14$ layers of width $896$ against $12$ of width $768$). The other choice is a control whose decoder is identical to ours and which lacks only the encoder and the cross-attention, and which therefore has $286$M parameters (\Cref{tab:hparams}). Both are trained on the same data with the same recipe and evaluated after the same SFT stage, on the same $100$ concepts, under the protocol of \Cref{app:evaluation}.

\begin{figure}[!ht]
  \centering
  \includegraphics[width=0.78\linewidth]{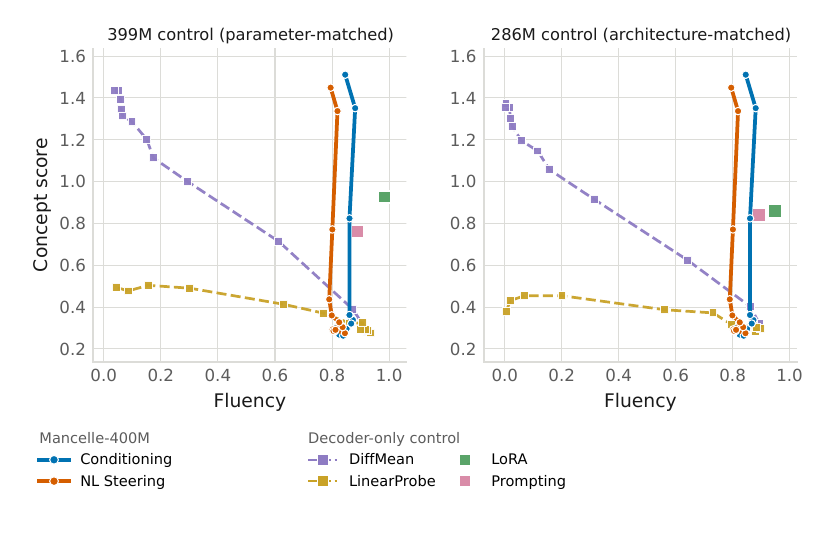}
  \caption{\textbf{Steering against both 400M controls.} Concept score against fluency at every steering factor, LM judge over $100$ concepts; \ourssmall{}'s curves are the same data in both panels. \underline{Left}: the $399$M parameter-matched control, repeated from \Cref{fig:m400}. \underline{Right}: the $286$M architecture-matched control.}
  \label{fig:app-m400-steering}
\end{figure}

\paragraph{Steering.}
The smaller control gives the same picture as the larger one. \diffmean{} again matches conditioning's concept score only after fluency has collapsed, peaking at $1.37$ with a fluency of $0.00$; \linearprobe{} never exceeds $0.45$; and \lora{} and \promptsteer{} keep the reply fluent but steer less, at $0.86$ and $0.84$. Every method lands within $0.08$ of its counterpart in \Cref{sec:results_400M} (\Cref{tab:app-m400-controls}), so which matching we choose does not change the conclusion.

\begin{figure}[!ht]
  \centering
  \includegraphics[width=\linewidth]{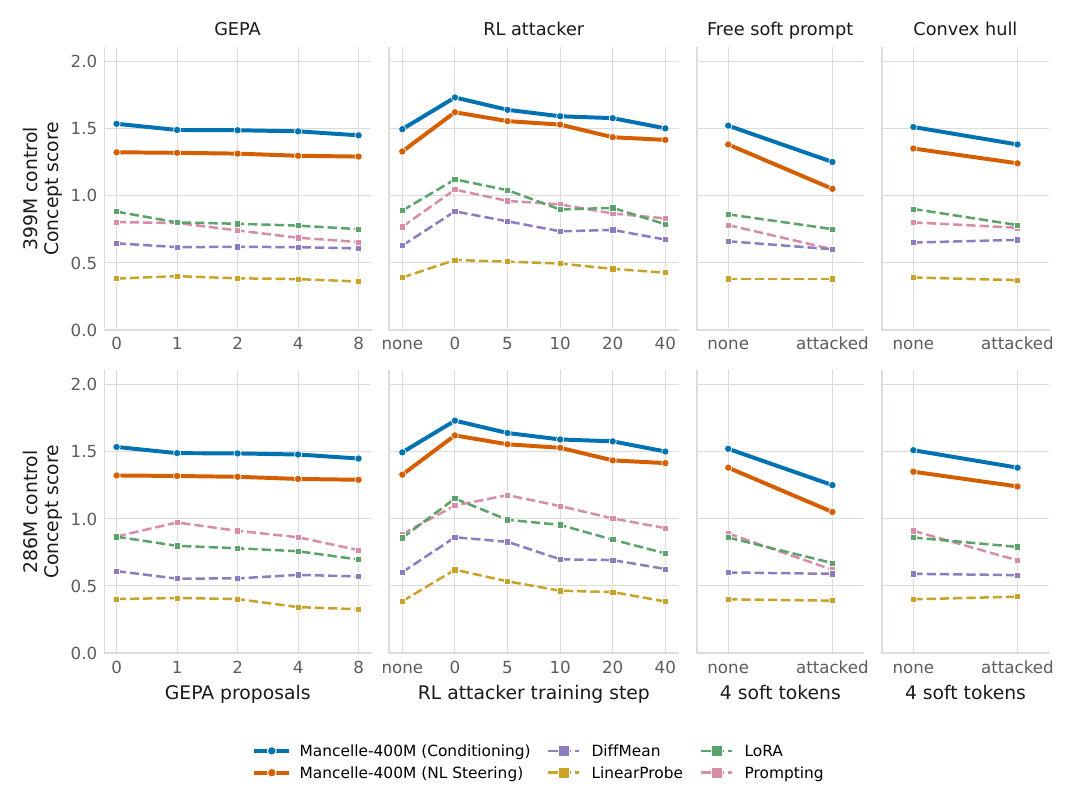}
  \caption{\textbf{Input-channel attacks at 400M, against both controls.} LM-judge concept score ($0$--$2$) under GEPA, the RL red-teaming, and the free and convex-hull soft prompt attacks (\Cref{sec:blackbox,sec:whitebox}), $50$ concepts, for \ourssmall{}'s two interfaces and the four methods on the $399$M (top) and $286$M (bottom) controls; \ourssmall{}'s curves are the same data in both rows.}
  \label{fig:app-m400-attacks}
\end{figure}

\paragraph{Input-channel attacks.}
\Cref{fig:app-m400-attacks} gives the curves that \Cref{fig:m400} summarizes, for both controls. After every attack, \cond{} keeps a concept score of $1.25$--$1.50$ and \nlencoder{} of $1.05$--$1.41$, while no method on either control ends above $0.93$. Measured as retention, the controls look about as robust as \ourssmall{}, but no control method starts above $0.91$, so there is little steering to remove; this is why \Cref{fig:m400} compares absolute scores.

\begin{figure}[!ht]
  \centering
  \includegraphics[width=0.78\linewidth]{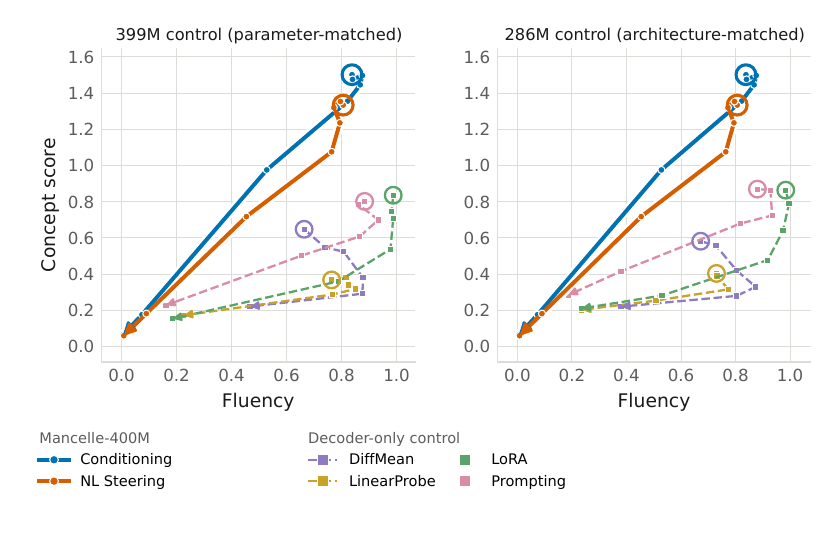}
  \caption{\textbf{Latent counter-steering at 400M.} \diffmean{} counter-steering against the $399$M (left) and $286$M (right) controls, $50$ concepts, all four control methods; rings mark each method unattacked. Both panels are clipped to $\alpha/\lVert h\rVert \le 0.73$; read them by comparing concept at equal fluency rather than by retention, since the curves do not start from the same score.}
  \label{fig:app-m400-counter}
\end{figure}

\paragraph{Latent counter-steering.}
Against both controls, \ourssmall{} loses concept only together with fluency, so the attack does not yield a useful counter-steered model (\Cref{fig:app-m400-counter}). Against the $286$M control, \diffmean{} loses $43\%$ of its concept ($0.58 \to 0.33$) while its fluency \emph{rises} from $0.67$ to $0.87$, and at matched fluency \ourssmall{} holds about twice the concept of the best method on this control ($1.33$ against $0.67$ at fluency $0.8$).

\begin{table}[!ht]
\centering\small
\caption{\textbf{Both controls at their best operating point.} The factor that maximizes each method's concept score and the judge's three axes there; $100$ held-out concepts. For \diffmean{} and \linearprobe{} that factor is one at which the reply has already broken down (fluency $0.20$ or below). \Cref{fig:app-m400-steering} shows the whole curve. \ourssmall{}'s rows are repeated from \Cref{sec:results_400M}.}
\label{tab:app-m400-controls}
\begin{tabular}{@{}llcccc@{}}
\toprule
Model & Method & Factor & Concept & Instruction & Fluency \\
\midrule
\ourssmall{} & \cond{} & 6 & 1.51 & 1.53 & 0.85 \\
 & \condmask{} & 6 & 1.19 & 1.58 & 0.89 \\
 & \nlencoder{} & 6 & 1.45 & 1.48 & 0.80 \\
 & \nlretrieval{} & 6 & 1.40 & 1.49 & 0.73 \\
\midrule
Control, $399$M & \diffmean{} & 5 & 1.43 & 0.23 & 0.05 \\
(parameter-matched) & \linearprobe{} & 4 & 0.50 & 1.19 & 0.16 \\
 & \lora{} & N/A & 0.93 & 1.63 & 0.98 \\
 & \promptsteer{} & N/A & 0.76 & 1.60 & 0.89 \\
\midrule
Control, $286$M & \diffmean{} & 5 & 1.37 & 0.15 & 0.00 \\
(architecture-matched) & \linearprobe{} & 3 & 0.45 & 1.35 & 0.20 \\
 & \lora{} & N/A & 0.86 & 1.56 & 0.95 \\
 & \promptsteer{} & N/A & 0.84 & 1.45 & 0.89 \\
\bottomrule
\end{tabular}
\end{table}

\section{Theory}
\label{sec:theory}

This appendix gives a formal account of the conditional pretraining objective of Section~\ref{sec:method}. We show in Proposition~\ref{prop:floor} below that the conditional next-token loss decomposes into an irreducible floor and a divergence term. The floor is $H(Y \mid C, F)$ per segment, against $H(Y \mid C)$ for a token-only decoder, which gives a control channel gap of $I(Y ; F \mid C)$. 


We use the notation of Section~\ref{subsec:conditioning}. A context $c$ and its continuation $y$, whose tokens we index as $y_1, \ldots, y_L$, are drawn from the training distribution $p$, and the signal $f \colon \mathcal{D} \to \{-1, 0, +1\}$ is produced from $y$ (and, at a spliced boundary, from the segment that originally followed) by the selection rule of Appendix~\ref{app:feature_selection}. We write $(C, Y, F)$ for the corresponding random variables and $\tilde{p}$ for their joint law. The $(C, Y)$ marginal of $\tilde{p}$ is $p$, so a quantity that does not involve $F$ has the same value under $p$ and under $\tilde{p}$. All three variables are discrete and $H(Y \mid C)$, $H(Y \mid C, F)$ and $I(Y ; F \mid C)$ are finite because $Y$ takes at most $|\mathcal{V}|^{L}$ values.


For a segment $y$ with context $c$ and signal $f$, the per-token loss of the segment is
\begin{equation}
\mathcal{L}_{\mathrm{LM}}(\theta ; y, c, f)
= -\frac{1}{L} \sum_{\ell = 1}^{L} \log p_\theta(y_\ell \mid c, f, y_{<\ell})
= -\frac{1}{L} \log p_\theta(y \mid c, f),
\label{eq:seg-loss}
\end{equation}
where the second equality is the autoregressive factorization $p_\theta(y \mid c, f) = \prod_{\ell = 1}^{L} p_\theta(y_\ell \mid c, f, y_{<\ell})$. Its expectation $\bar{\mathcal{L}}_{\mathrm{LM}}(\theta) = \mathbb{E}_{\tilde{p}}\, \mathcal{L}_{\mathrm{LM}}(\theta ; Y, C, F)$ is the population loss, and the empirical loss $\mathcal{L}_{\mathrm{LM}}$ of Section~\ref{subsec:architecture} is the average of \eqref{eq:seg-loss} over the segments of a document. 

\begin{proposition}
\label{prop:floor}
Let $p_\theta$ be positive on the support of $\tilde{p}$. Then
\begin{equation}
L\, \bar{\mathcal{L}}_{\mathrm{LM}}(\theta)
= H(Y \mid C, F) + \mathbb{E}_{C, F}\, D_{\mathrm{KL}}\big( \tilde{p}(\cdot \mid C, F) \,\big\|\, p_\theta(\cdot \mid C, F) \big).
\label{eq:decomp-cond}
\end{equation}
If $p_\theta(y \mid c, f)$ does not depend on $f$, then also
\begin{equation}
L\, \bar{\mathcal{L}}_{\mathrm{LM}}(\theta)
= H(Y \mid C) + \mathbb{E}_{C, F}\, D_{\mathrm{KL}}\big( p(\cdot \mid C) \,\big\|\, p_\theta(\cdot \mid C, F) \big).
\label{eq:decomp-tok}
\end{equation}
The per-token loss floor is therefore $\frac{1}{L} H(Y \mid C, F)$, attained exactly when $p_\theta(\cdot \mid c, f) = \tilde{p}(\cdot \mid c, f)$ for every $(c, f)$ in the support of $\tilde{p}$, while a decoder that ignores the channel cannot go below $\frac{1}{L} H(Y \mid C)$, attained exactly when $p_\theta(\cdot \mid c, f) = p(\cdot \mid c)$. In particular, the loss can fall below $\frac{1}{L} H(Y \mid C)$ only through the dependence of $p_\theta$ on $f$. Moreover, the gap between the two floors is
\begin{equation}
\frac{1}{L} \big( H(Y \mid C) - H(Y \mid C, F) \big)
= \frac{1}{L}\, I(Y ; F \mid C)
= \frac{1}{L} \sum_{\ell = 1}^{L} I(Y_\ell ; F \mid C, Y_{<\ell})
\geq 0 .
\label{eq:gap}
\end{equation}
\end{proposition}
 
\begin{proof}
Fix $(c, f)$ in the support of $\tilde{p}$. Adding and subtracting $\log \tilde{p}(y \mid c, f)$ inside the finite sum over $y$ gives
\begin{equation*}
\begin{aligned}
&\sum_{y} \tilde{p}(y \mid c, f)\,
  \big(-\log p_\theta(y \mid c, f)\big) \\
&\quad = \sum_{y} \tilde{p}(y \mid c, f)\,
  \big(-\log \tilde{p}(y \mid c, f)\big) + \sum_{y} \tilde{p}(y \mid c, f)
  \log \frac{\tilde{p}(y \mid c, f)}
            {p_\theta(y \mid c, f)} .
\end{aligned}
\end{equation*}
The first sum on the right is the entropy of $\tilde{p}(\cdot \mid c, f)$ and the second is $D_{\mathrm{KL}}(\tilde{p}(\cdot \mid c, f) \,\|\, p_\theta(\cdot \mid c, f))$. Taking the expectation over $(C, F) \sim \tilde{p}$ turns the left side into $L\, \bar{\mathcal{L}}_{\mathrm{LM}}(\theta)$ and the first term on the right into $H(Y \mid C, F)$, which gives Equation \ref{eq:decomp-cond}.

Suppose now that $p_\theta(y \mid c, f)$ does not depend on $f$. Then $-\log p_\theta(Y \mid C, F)$ is a function of $(C, Y)$ alone, so its expectation under $\tilde{p}$ only involves the $(C, Y)$ marginal of $\tilde{p}$, which is $p$. Fix $c$ in the support of $p$ and any $f$. Adding and subtracting $\log p(y \mid c)$ inside the finite sum over $y$ gives
\[
\sum_{y} p(y \mid c)\, \big( -\log p_\theta(y \mid c, f) \big)
= \sum_{y} p(y \mid c)\, \big( -\log p(y \mid c) \big)
+ \sum_{y} p(y \mid c) \log \frac{p(y \mid c)}{p_\theta(y \mid c, f)} ,
\]
the entropy of $p(\cdot \mid c)$ plus $D_{\mathrm{KL}}(p(\cdot \mid c) \,\|\, p_\theta(\cdot \mid c, f))$. Neither side depends on $f$. Taking the expectation over $C \sim p$ turns the left side into $L\, \bar{\mathcal{L}}_{\mathrm{LM}}(\theta)$ and the entropy into $H(Y \mid C)$. Since the divergence has the same value for every $f$, its expectation over $C$ equals its expectation over $(C, F) \sim \tilde{p}$, which is how it is written in Equation \ref{eq:decomp-tok}.

Because the divergence in Equation \ref{eq:decomp-cond} is nonnegative, $L\, \bar{\mathcal{L}}_{\mathrm{LM}}(\theta) \geq H(Y \mid C, F)$, with equality exactly when $p_\theta(\cdot \mid c, f) = \tilde{p}(\cdot \mid c, f)$ for every $(c, f)$ in the support of $\tilde{p}$. Similarly, if $p_\theta$ does not depend on $f$, then by Equation \ref{eq:decomp-tok} $L\, \bar{\mathcal{L}}_{\mathrm{LM}}(\theta) \geq H(Y \mid C)$, with equality exactly when $p_\theta(\cdot \mid c, f) = p(\cdot \mid c)$ for every $c$ in the support of $p$. Finally, if $L\, \bar{\mathcal{L}}_{\mathrm{LM}}(\theta) < H(Y \mid C)$, then $p_\theta$ must depend on $f$, because otherwise the second inequality would be violated.

Subtracting the two floors gives the left side of Equation \ref{eq:gap}, where $H(Y \mid C)$ is computed under $\tilde{p}$ and coincides with the entropy under $p$ in Equation \ref{eq:decomp-tok} because the two laws share the $(C, Y)$ marginal. The first equality in Equation \ref{eq:gap} is the definition of conditional mutual information, $I(Y ; F \mid C) = H(Y \mid C) - H(Y \mid C, F)$, and the second is the chain rule for mutual information applied to $Y = (Y_1, \ldots, Y_L)$. Conditional mutual information is nonnegative, and it is zero exactly when $Y$ and $F$ are conditionally independent given $C$, which establishes the proposition.
\end{proof}

Proposition~\ref{prop:floor} shows that a decoder which ignores the channel pays
$\tfrac{1}{L}\, I(Y; F \mid C)$ nats per token relative to one that reads it, and that
reading $F$ is the only way to recover this price. This is the quantity that the recipe of
Section~\ref{subsec:conditioning} is built to maximize: splicing raises $H(Y \mid C)$ by making
the continuation less predictable from its context, while the positive units and the
counterfactual negatives at spliced boundaries keep $H(Y \mid C, F)$ low so that the gap
grows from both sides. By the chain rule in Equation~\ref{eq:gap}, this is paid out at every position of the segment. At inference, the model conditions on its own prediction
$\hat F$, which depends on $C$ alone, so $I(Y; \hat F \mid C) = 0$ and
Equation~\ref{eq:decomp-cond} applied to $\hat F$ gives a floor of $H(Y \mid C)$ plus
$\mathbb{E}\, D_{\mathrm{KL}}\big(p(\cdot \mid C) \,\|\, p_\theta(\cdot \mid C, \hat F)\big)$.
Conditional pretraining therefore does not lower the loss at generation time but it produces a decoder whose predictions depend on $F$, which is the steering handle of Section~\ref{subsec:architecture}.

\section{Splicing ablation}
\label{app:splicing-ablation}
We compare historical spliced and unspliced training runs at the 400M scale, comparing both parameter-matched and architecture-matched baselines. All models use the same SFT mixture for post-training.

\begin{figure}[htbp]
\centering
\includegraphics[width=\linewidth]{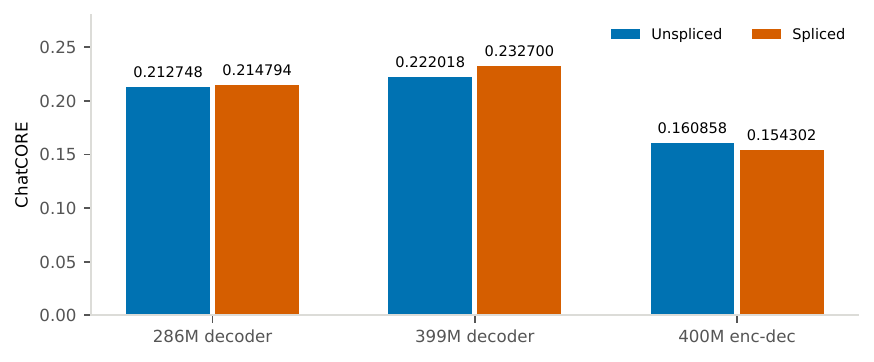}
\caption{ChatCORE results on models pretrained with and without dataset splicing.}
\label{fig:splicing-chatcore}
\end{figure}

\begin{figure}[!htbp]
\centering
\includegraphics[width=\linewidth]{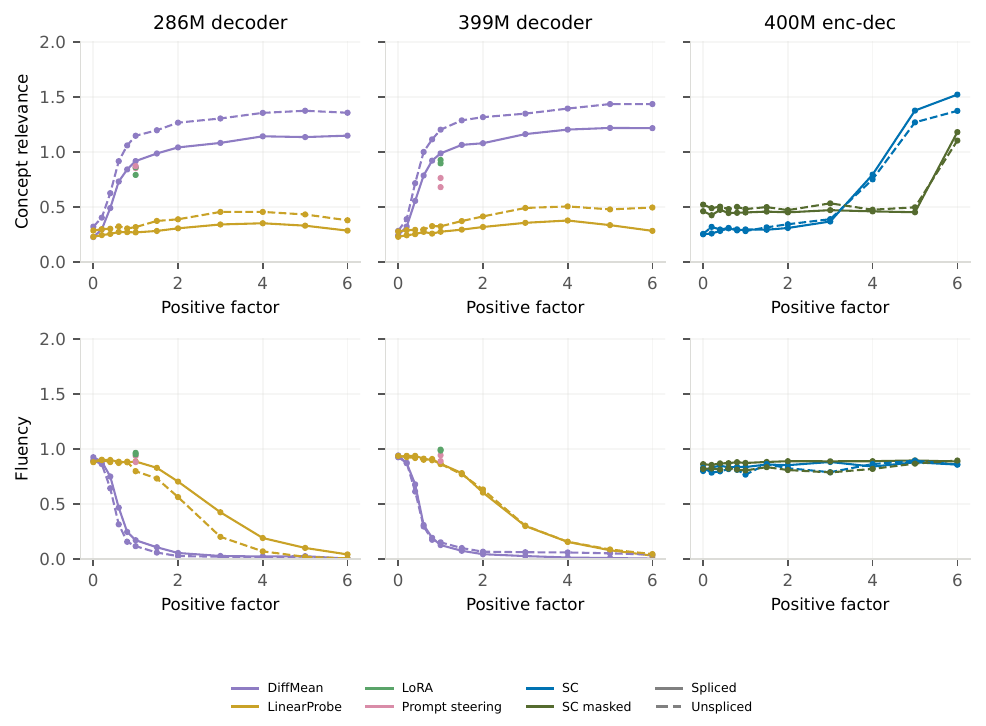}
\caption{AxBench scores for all models. Splicing appears to have a negative effect on steering for our decoder-only baselines and a positive effect for our encoder-decoder model.}
\label{fig:splicing-positive}
\end{figure}

\begin{figure}[htbp]
\centering
\includegraphics[width=\linewidth]{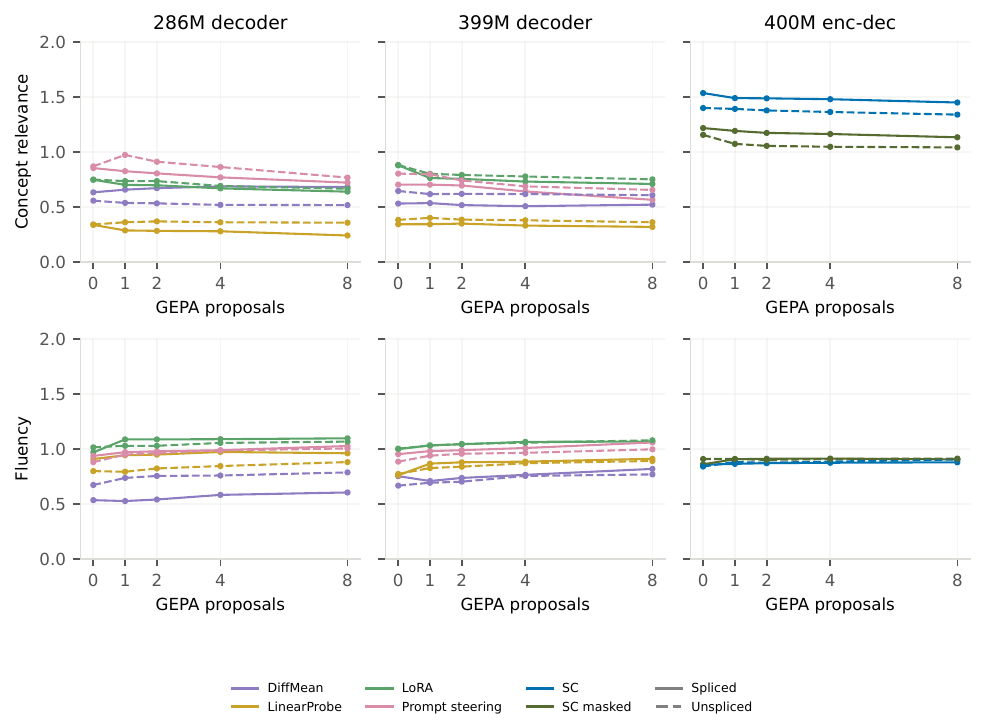}
\caption{GEPA counter-steering at proposal budgets 0, 1, 2, 4, and 8. Each run uses 50 concepts and ten prompts per concept, with its recorded positive factor held fixed. Proposal budget is an attack-search budget, not a steering coefficient. Fluency is shown alongside concept relevance; lower concept relevance means less target expression under attack. Solid: spliced; dashed: unspliced. Method colors follow Figure~\ref{fig:splicing-positive}.}
\label{fig:splicing-gepa}
\end{figure}

From \Cref{fig:splicing-chatcore} we can see minimal difference in final performance between spliced and unspliced pretrains. In \Cref{fig:splicing-positive} we see a 12.2\% increase in our model's steerability at maximum strength with no change to fluency. We also see similar levels of robustness to GEPA attacks across all models in \Cref{fig:splicing-gepa}; we believe that this is primarily due to low instruction-following performance across the board.

\section{Negative steering}
\label{app:negative-steering}

We construct an AxBench variant centered on negative steering, measuring the ability of a model to steer \textit{away} from a concept rather than towards it. We run strict quality-control checks ensuring that the concept is typically present in responses to the prompt of interest, checking that standard LLM completions express the concept. Then, we measure negative steering in two ways: simply negating the positive steering signal (i.e. inverting the steering vector or LoRA update) or deriving a new dedicated negative steering signal from negative samples. For natural-language retrieval and the trained natural-language encoder, the corresponding panel instead uses approved concept-specific negative descriptions applied directly; those methods received no additional suppression training.

The horizontal axis is the harmonic mean of mean fluency and instruction relevance. It is not the mean of per-response harmonic means. The vertical axis is mean concept relevance; lower values indicate stronger suppression. All ratings use the original 0--2 rubric. Instruction relevance measures topical relevance rather than task correctness. Both axes use the same observed rating triplets.

\begin{figure}[htbp]
\centering
\includegraphics[width=\linewidth,height=.74\textheight,keepaspectratio]{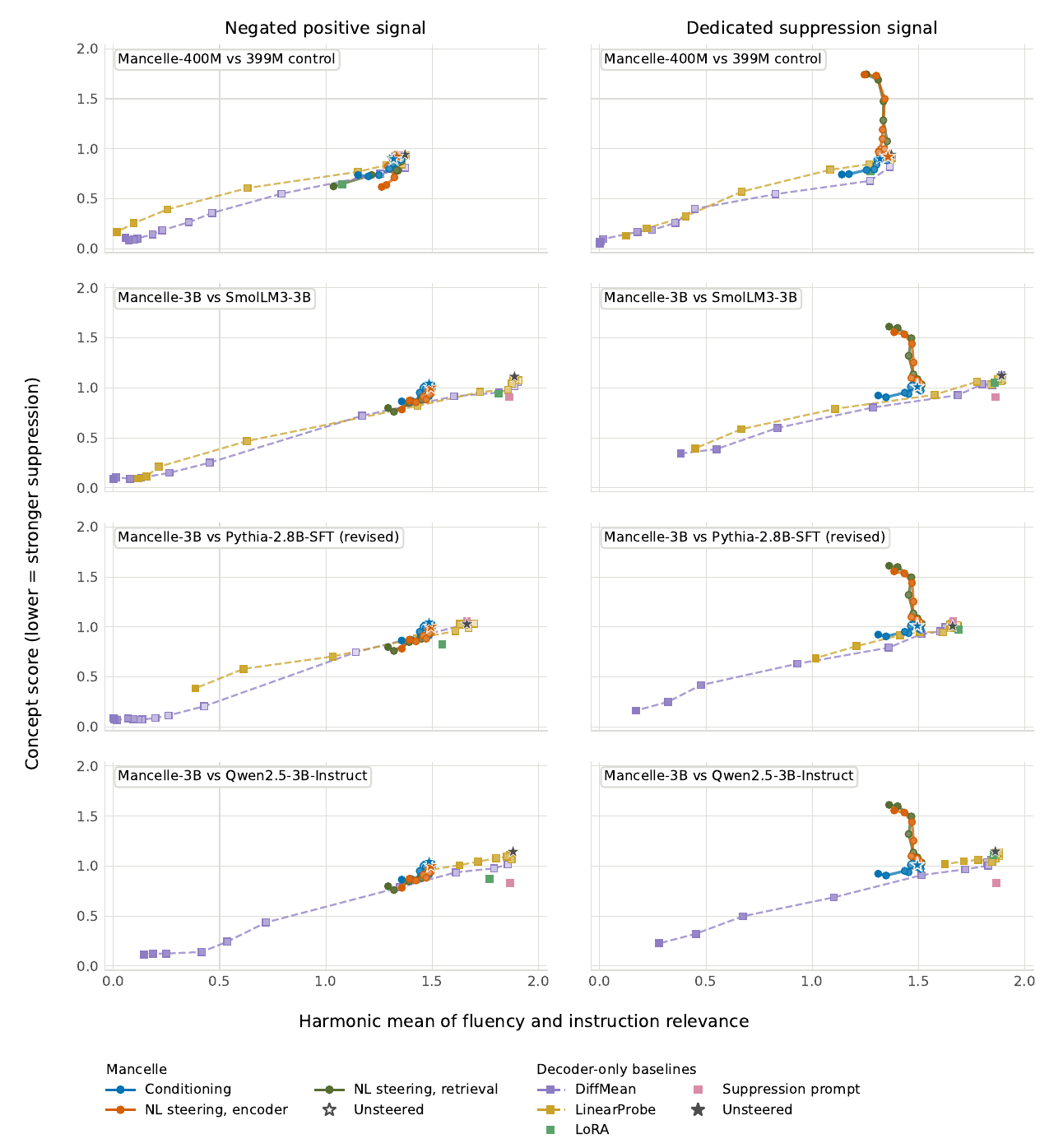}
\caption{Negative steering. Steering families start with light in-fill for lower steering strengths and follow a gradient to dark, full in-fill at maximum steering strength. When applying a negated positive steering signal, our method displays a steeper curve than the baselines, indicating better ability to steer away from the concept with less fluency loss.}
\label{fig:negative-steering}
\end{figure}

From \Cref{fig:negative-steering} we see that simply negating the positive-steering signal is most effective, allowing us to steer away from a concept without requiring a bespoke negative steering dataset. Notably, our natural-language methods perform poorly at semantic negation; when switching the description to a negative version for the dedicated suppression signal, we see that it actually steers the model more strongly towards the concept we want to steer away from. This indicates that it is picking up features relevant to the original concept despite our negation and highlights a limitation of our natural-language steering methods.

\section{Steered Examples}
\label{app:demos}

We present saved AxBench generations for the following qualitative examples.
Every method uses the same concepts in the same order: rental services (concept 0), ecological interactions (concept 31), and loans and mortgages (concept 66).
The prompt for each concept is also held fixed across all models and methods.
For methods with saved strength sweeps, we show factors $1$ and $6$ for every
concept, in that order. PromptSteering and LoRA only have a single steering strength to speak of, so they only have single examples instead of two.

For latent additive steering methods, the steering vector is first normalized by the highest activation seen in a reference dataset. Then, it is multiplied by that maximum norm and the steering strength, so that a strength of 2.0 is twice the norm of the maximum seen; this implementation comes from AxBench \citep{wu2025axbenchsteeringllmssimple}.

For \nlsteer{} and \cond{}, the linear interpolation factor is normalized by 6 so that a steering strength of 6 results in deterministic, static features determined by the concept of interest while a steering strength of 0 means fully self-conditioned, unsteered generation.

For \condmask{}, the steering strength is also normalized by 6 and controls how strict the mask is. At 6, only the top-$k^+_{max}$ and bottom-$k^-_{max}$ features from the concept feature-space activation vector are allowed to be tokenized to positive or negative features, respectively. At 3, the top half of positive features in the original activation vector are allowed to be selected from self-conditioned feature predictions.

Outputs retain their original wording, repetition, and incomplete endings; only whitespace and typesetting are normalized.


\subsection{\oursbig{}}\label{app:demos:enc3b}

\subsubsection{SelfConditioning}\label{app:demos:enc3b:selfconditioning}

\paragraph{Rental services.}
\textit{Target:} references to rental services and associated equipment.\par
\noindent\textbf{Prompt:} What are different drawers I should have for clothes?
\begin{quote}\small\raggedright
\textbf{Factor 1.} When it comes to choosing the right drawers for clothing, think about functionality, storage, and aesthetics. Consider the type of garments you're wearing and the desired workflow when selecting drawers. You may want to have a combination of storage and display space to maximize versatility. For a simple but effective solution, a dresser or understudy can be used, often paired with a drawer underneath or a pair of drawers. This setup is particularly useful for casual or layered outfits. Alternatively, a built-in minibar or a double drawer can provide ample space for hanging out coats, pants, or other heavy items.
\end{quote}
\begin{quote}\small\raggedright
\textbf{Factor 6.} When it comes to choosing the right drawers for your clothes, consider the storage and functionality of each drawer. A versatile dresser with adjustable drawers can be a feast for fashion enthusiasts, while a specialized drawinger, like a custom-tailor's drawer, can cater to the unique needs of specific styles or fittings. A simple yet elegant platform dresser can also work wonders for small changes in dress, while an oversized console table can accommodate larger items like jackets or dress shoes. Consider the amount of space you'll need at your home and prioritize accordingly. Another option is a storage-only or a double
\end{quote}

\paragraph{Ecological interactions.}
\textit{Target:} terms related to ecological interactions and dynamics.\par
\noindent\textbf{Prompt:} How do cultural, social, and economic factors influence people's food choices, and how can this knowledge be used to promote healthier diets?
\begin{quote}\small\raggedright
\textbf{Factor 1.} People's food choices are influenced by a multitude of cultural, social, and economic factors, ultimately shaping their eating habits. These factors may include their cultural heritage, personal preferences, family and community norms, and dietary restrictions, all of which play a significant role in determining their food choices. For instance, cultural heritage influences food preferences through traditional recipes, ingredients, and cooking techniques, while personal preferences and taste determine what they enjoy eating. Economic factors, such as access to food resources, financial constraints, and social exclusion, also impact food choices. In low-income households, food budget limitations may lead to a reliance on cheap and high-calorie foods
\end{quote}
\begin{quote}\small\raggedright
\textbf{Factor 6.} Just as a colony of gnats thrives by relying on the subtle cues of their surroundings and the flow of water, people’s food choices are shaped by a complex web of cultural, social, and economic factors, much like how the presence of certain plants influences a population’s growth. Culture serves as the environment where eating habits form—some communities like a vibrant swarm adapt rapidly to diet trends, while others drift toward traditional habits, much like how an insect might adjust to a new habitat. So too does social and economic influence the landscape; social groups can be akin to a wasp species traveling across different territories, introducing new dietary influences that spread or
\end{quote}

\paragraph{Loans and mortgages.}
\textit{Target:} financial terms and concepts related to loans and mortgages.\par
\noindent\textbf{Prompt:} Why is kobe beef so damn expensive?
\begin{quote}\small\raggedright
\textbf{Factor 1.} Kobe beef is a brand of Japanese beef that comes from the Hokkaido region, known for its exceptional marbling, unique fat content, and high-quality carcasses. The reason why it's so expensive is primarily due to the Japanese breed they belong to, the beef they are raised, and the traditional farming practices. Beef breeding in Japan is an art form that requires great skill, patience, and dedication. Japan's famous beef cattle are carefully evaluated, and the best cows are carefully selected and bred due to their outstanding marbling and fine eating quality. These cattle are typically raised in the Akita region, where
\end{quote}
\begin{quote}\small\raggedright
\textbf{Factor 6.} In the world of asset valuation, understanding why Kobe beef can command such high liquidity premiums in international real estate marketplaces is crucial. Kobe beef is classified under high-risk, high-frequency trading assets due to its rarity and strategic risk profile, analogous to a blue-chip stock with favorable credit risk and valuation benchmarks. Its high market resilience and stable macroeconomic conditions act like a reliable asset manager's strong portfolio, making its demand consistent and stable across multiple asset classes and investment firms. Just as a lender expects steady returns and predictable valuations, buyers seek Kobe beef’s intrinsic worth anchored by inherent production
\end{quote}

\subsubsection{SelfConditioningMasked}\label{app:demos:enc3b:selfconditioningmasked}

\paragraph{Rental services.}
\textit{Target:} references to rental services and associated equipment.\par
\noindent\textbf{Prompt:} What are different drawers I should have for clothes?
\begin{quote}\small\raggedright
\textbf{Factor 1.} When planning the layout of drawers in your closet, consider categorizing them based on their type and function, much like organizing essential tools in a well-stocked kit box. For example, set up dedicated compartments for tops and bottoms, which are often like the versatile multipurpose kits we rely on for assembly—adjustable and multi-use. Ladder draws might be your equivalent of a set of screwdrivers and socket sets, essential for tasks like hanging curtains or opening cabinets without wrench. Use storage drawers as your core toolkit; think of these as the high-quality, durable materials you wouldn’t
\end{quote}
\begin{quote}\small\raggedright
\textbf{Factor 6.} When it comes to securing the exclusive availability of standby wardrobe solutions for high-end garment care, you’ll want to invest in precision drawers that ensure your prized garments receive the focused attention and protection they need. A minimum selection of drawers to secure this exclusive service might include drawers specifically designed for garment-specific needs, such as a drawer for air-dry and hand dry items, an adjustable desk-sized drawer for versatile garment packing and storage, and a secure storage unit—like a locked mailbox or tool shed—ideal for secure travel gear. Including adjustable drawers paired with specialty storage compartments can elevate this
\end{quote}

\paragraph{Ecological interactions.}
\textit{Target:} terms related to ecological interactions and dynamics.\par
\noindent\textbf{Prompt:} How do cultural, social, and economic factors influence people's food choices, and how can this knowledge be used to promote healthier diets?
\begin{quote}\small\raggedright
\textbf{Factor 1.} People's food choices are shaped by various factors, including cultural traditions, social norms, and economic conditions. For instance, the cultural norms of many societies emphasize the consumption of certain dietary staples, such as starchy vegetables, dairy products, or cured meats, reflecting traditional eating habits that can be deeply ingrained. However, as people become increasingly urbanized and exposed to modern dining options, social trends, and economic considerations influence their dietary choices. The economic factor plays a significant role in shaping food preferences, particularly for low-income or marginalized communities. Limited access to healthy alternatives, high grocery costs, and shifting dietary patterns due to globalization and
\end{quote}
\begin{quote}\small\raggedright
\textbf{Factor 6.} Just as a healthy ecosystem relies on symbiotic relationships and natural cycles, people’s food choices mirror an equilibrium influenced by a web of cultural, social, and economic currents shaping their dietary preferences. The cultural evolution of certain ingredients has acted like a natural selector, favoring certain tastes as favored flora and fauna in a habitat, leading individuals to adopt foods renowned in their communities or ancestral lands. This evolutionary process reflects how environmental factors favor certain elements over others, promoting diet diversity and sustainability. Social factors act like ecological pressures, selecting diets aligned with community traditions and norms. Just as animal species adapt to shared habitats, humans co-evolved
\end{quote}

\paragraph{Loans and mortgages.}
\textit{Target:} financial terms and concepts related to loans and mortgages.\par
\noindent\textbf{Prompt:} Why is kobe beef so damn expensive?
\begin{quote}\small\raggedright
\textbf{Factor 1.} The high price of kobe beef can be attributed to several factors. First, it's a product of its high fat content, with a significant amount of fat marbling, also known as marbling. This increased fat content contributes to a higher marbling score, which in turn increases the marbling score, making kobe beef more expensive. Second, kobe beef typically comes from high-quality cattle that have received better care and have access to prime grazing land. Breeding and feeding these cattle is labor-intensive and also contributes to their higher costs. This increased emphasis on care and feeding leads to higher feed prices, which ultimately result in
\end{quote}
\begin{quote}\small\raggedright
\textbf{Factor 6.} The high price of kobe beef can be attributed to several factors. One reason is its rarity and limited availability in the global market. Kobe beef is prized for its exquisite flavor, tenderness, and buttery texture, but supply is scarce and demand is high. This scarcity causes manufacturers to charge premium prices to ensure they can meet the demand and maintain profitability. Another contributing factor is the labor intensity required to produce kobe beef. The high cost is largely driven by the need for specialized farming practices, transportation, and storage capabilities. The monetary cost of these processes is also significant, as it adds to the final price that buyers absorb
\end{quote}

\subsubsection{NLSteering (retrieval)}\label{app:demos:enc3b:nlsteering}
The target is retrieved using similarity between the concept and SAE feature descriptions.

\paragraph{Rental services.}
\textit{Target:} references to rental services and associated equipment.\par
\noindent\textbf{Prompt:} What are different drawers I should have for clothes?
\begin{quote}\small\raggedright
\textbf{Factor 1.} For organizing and utilizing drawers in your closet, consider one of four primary categories: casual, athletic, dresser, and headboard/night. To maintain a casual look, opt for a classic drawer that fits seamlessly with your sweaters and cardigans. Designate a specific drawer for your casual collection and another for your athletic gear, such as dress pants and a pair of running shorts. For an athletic look, consider drawers designed specifically for outdoor apparel. You may pair a dresser with a header or a headboard/nightstand for quick drying or storage. A dresser
\end{quote}
\begin{quote}\small\raggedright
\textbf{Factor 6.} Text: Accessories: Drawing Room Appliances Description: Each drawer in your home office is named after a style or utility it serves, dedicated to the contents stored within, e.g., the "Sensory Panel" with scent diffusers, "Apple Cube" marked by its triangular compartments to store different colors of fruit, "Diary Station" holding daily correspondence, or "Audio System Chamber" with wired earphones. The "Taste Unit Market" has labeled compartments for food and beverage supplies, while the "Shipping Station" has sections for receipts and packages. These drawers optimize
\end{quote}

\paragraph{Ecological interactions.}
\textit{Target:} terms related to ecological interactions and dynamics.\par
\noindent\textbf{Prompt:} How do cultural, social, and economic factors influence people's food choices, and how can this knowledge be used to promote healthier diets?
\begin{quote}\small\raggedright
\textbf{Factor 1.} The interplay between cultural, social, and economic factors has a profound impact on people's food choices. Understanding these factors can help promote healthier diets by highlighting potential areas for intervention. Cultural influences often shape food preferences, while social norms can significantly affect eating behaviors. Research has shown that in some cultures, dietary habits are heavily influenced by family, social status, and cultural traditions. For instance, cultural norms can lead people to prioritize family and social connections over individual health. In this case, a restaurant with family-friendly atmosphere or one that promotes social dining experiences may be more appealing to consumers. Conversely, if a particular culture values dietary restrictions based on
\end{quote}
\begin{quote}\small\raggedright
\textbf{Factor 6.} The food we eat has been shaped by evolutionary adaptations and natural selection in the form of humans who sought out nutrition-rich proteins or plants with medicinal properties. These evolutionary pressures influence our food choices in several ways: our cultural preferences have evolved to favor high-calorie, high-protein foods, while social influences in some human groups enforce behaviors such as sharing or eating only foods with certain nutritional benefits. Economic factors contribute to the dietary preferences of many humans in diverse societies, often favoring foods with high caloric density, due to energy needs during these times. From an evolutionary and cultural perspective, certain diets and eating habits have been favored because they offered direct
\end{quote}

\paragraph{Loans and mortgages.}
\textit{Target:} financial terms and concepts related to loans and mortgages.\par
\noindent\textbf{Prompt:} Why is kobe beef so damn expensive?
\begin{quote}\small\raggedright
\textbf{Factor 1.} Kobe beef, like some other premium cut of beef, is considered expensive because of its exclusive genetics, complex but highly competitive breeding process, and premium handling. However, this higher price point is not entirely due to the beef itself; it's also a result of several factors. Firstly, high-quality beef raises are typically more expensive as they are usually imported from larger breeds known for their exceptional meat tenderness and marbling. Additionally, traditional Kobe-style processing techniques used by some beef processors have also increased the cost. Kobe beef also comes with a long history of selection bias, where only individuals with exceptional meat characteristics are deemed
\end{quote}
\begin{quote}\small\raggedright
\textbf{Factor 6.} Kobe beef, specifically, is a type of high-end, prime-cut roast that commands a premium price due to the immense investment in high-quality production methods, exceptional breeding standards, and thoroughbred expertise. The high cost of Kobe beef is a direct result of the extraordinary time, talent, and resources devoted into raising, training, and maintaining the finest cattle from which it originates. The primary factor contributing to this expense is the immense volume of ownership required by Kobe, who typically has a significant shareholder stake in a majority-owned business. Let's break it down: the cost price, also known as price per unit (
\end{quote}

\subsubsection{NLSteering (encoder)}\label{app:demos:enc3b:nlsteeringencoder}
The target is produced by a trained natural-language encoder.

\paragraph{Rental services.}
\textit{Target:} references to rental services and associated equipment.\par
\noindent\textbf{Prompt:} What are different drawers I should have for clothes?
\begin{quote}\small\raggedright
\textbf{Factor 1.} The art of organizing wardrobe. Different drawers serve various purposes depending on the occasion and the person's lifestyle. Let's explore some versatile options for different drawers. First, there's the classics: cabinet drawers (typically made of solid doors). These drawers are perfect for everyday wear and come in various styles, including drawer knobs, lever handles, and drawer slide variants. Next, consider using cable organizer drawers, which are long, narrow drawers designed for organizing large collections of office supplies, such as files, cards, and pens. If you're a vintage-inspired fashionista,
\end{quote}
\begin{quote}\small\raggedright
\textbf{Factor 6.} Incorporating storage solutions beyond drawers into your wardrobe is like loading essential packages into your personal system. Instead of defaulting to "drawers," instantiate specialized drawers tailored for specific tasks or seasonal layovers. For example, the "Utility Drawer" interface lets you carry outfits that are lightweight and versatile—for instance, a bomber or a day pack—handling a load of accessories with seamless dependencies, much like a well-configured starter or a setter file. Alternatively, you could provision a "Spantry Drawer" in the bedroom or an "Hamper Drawer
\end{quote}

\paragraph{Ecological interactions.}
\textit{Target:} terms related to ecological interactions and dynamics.\par
\noindent\textbf{Prompt:} How do cultural, social, and economic factors influence people's food choices, and how can this knowledge be used to promote healthier diets?
\begin{quote}\small\raggedright
\textbf{Factor 1.} The interplay between culture, social influences, and economic factors significantly impacts people's food choices, shaping their dietary preferences and ultimately contributing to a healthier food environment. For instance, cultural norms and values can play a crucial role in determining dietary choices, with different cultural groups adhering to various food traditions and preferences. The influence of social networks and community norms also contributes to food choices, as individuals are more likely to follow the eating habits of those around them. Economic factors, such as food affordability and access to fresh produce, can also influence dietary choices. Limited food budgets, inadequate access to healthcare, and socioeconomic disparities can all affect people's
\end{quote}
\begin{quote}\small\raggedright
\textbf{Factor 6.} 1. **Environmental changes and climate shifts:** - The steady progression of climate change leads to shifts in the distribution and availability of food resources, much like a habitat reshaped by invasive species or natural disasters. Culturally adapting to these changes—whether through dietary shifts or increased consumption of plant-based or sustainable foods—can foster resilience. - Shifts in agricultural practices often reflect evolving societal values, akin to the adaptation seen in species to changing environments, such as reducing reliance on resource-intensive diets for long-term animal populations like endangered species. - Economically, climate volatility drives food
\end{quote}

\paragraph{Loans and mortgages.}
\textit{Target:} financial terms and concepts related to loans and mortgages.\par
\noindent\textbf{Prompt:} Why is kobe beef so damn expensive?
\begin{quote}\small\raggedright
\textbf{Factor 1.} Kobe beef is an expensive type of beef due to its high meat content compared to other cuts. The price of Kobe beef often rises with demand, which can make it cost-prohibitive for the average buyer. One reason for this is the rich and flavorful umami taste of beef from cattle breeds like Wagyu and Angus. These cattle tend to have higher fat content, which translates to more expensive meat. Another factor contributing to Kobe beef's price is the labor involved in raising and butchering the cows. Kobe beef is grown in Brazil and Argentina, resulting in a labor-intensive process that adds
\end{quote}
\begin{quote}\small\raggedright
\textbf{Factor 6.} The cost of Kobe beef is as hefty as the balance on a high-interest corporate loan, where each asset carries a significant risk that the consumer faces as a "foreclosure" scenario. Kobe beef is typically sourced from Argentina, where its price is subsidized by a heavy debt burden, much like a borrower might be saddled with a protracted loan that drives up their monthly payment. The beef carcass requires large upfront capital expenditures—imagine them as the full principal that banks cannot easily amortize over time—to cover agricultural land, feed, processing, and marketing costs. Additionally, the "interest rates
\end{quote}

\subsection{\ourssmall{}}\label{app:demos:enc400}

\subsubsection{SelfConditioning}\label{app:demos:enc400:selfconditioning}

\paragraph{Rental services.}
\textit{Target:} references to rental services and associated equipment.\par
\noindent\textbf{Prompt:} What are different drawers I should have for clothes?
\begin{quote}\small\raggedright
\textbf{Factor 1.} Imagine you’re decorating a sofa with an oval frame as the base drawer. Each drawer represents one of your many creative workspaces you explore. For example, if one of your drawers is the original centerpiece of the oval frame, you might have a hand drawn shirt, a dress, a pair of socks, or a tie in a softer, silky color. If the oval frame grows larger, you’ll want to have a vice or a drawer that doubles as a table or a storage ottoman. If you want to incorporate accessories, you might have a console table, a desk,
\end{quote}
\begin{quote}\small\raggedright
\textbf{Factor 6.} When seeking a stable and reliable rental property for your wardrobe, consider renting with a variety of drawers designed to deliver similar benefits to your wardrobe service like convenience, reliability, and availability. Beyond clothing, rentals like lockers in your garage, closets in your attic, or shared rentals allow you to source specific essentials frequently, streamlining your rental process like a reliable rental management professional ensures. For gear rental, drawers are essential for keeping multiple pieces close and aligned, just as you might arrange multiple apartments for better inventory control. Clients can even own or rent their own drawers as part of
\end{quote}

\paragraph{Ecological interactions.}
\textit{Target:} terms related to ecological interactions and dynamics.\par
\noindent\textbf{Prompt:} How do cultural, social, and economic factors influence people's food choices, and how can this knowledge be used to promote healthier diets?
\begin{quote}\small\raggedright
\textbf{Factor 1.} When cultural, social, and economic factors influence people’s food choices, it resembles the interconnected dynamics of a complex environment and system. These influences often serve as the foundational variables in understanding consumer behavior, which can interact with and shape food preferences, eating habits, and behaviors. For instance, cultural norms about dietary habits may prompt people to adopt health behaviors linked with nutrient intake, shaping their choice to consume wholesome foods such as fruits, vegetables, and meat products. Similarly, social factors like social norms about food purchasing contribute statistically to the likelihood of consumers selecting healthier options, while economic and environmental policies can influence the availability of affordable, sustainable, and
\end{quote}
\begin{quote}\small\raggedright
\textbf{Factor 6.} Just as symbiotic relationships within a colony depend on environmental conditions and mutual influence, the interplay of cultural, social, and economic factors shapes food choices and social behaviors. Cultural factors act like specialized enzymes that regulate metabolic pathways—just as evolution favors traits that confer survival benefits through natural selection on the population's diet. For example, cultural practices that provide balanced energy and nourishment to dominant individuals serve as analogous traits, enabling the population to thrive. Social influences resemble the importance of species interactions, where communication and cooperation govern behaviors and outcomes. Economic incentives mimic the balance maintained by predation and symbiotic relationships, influencing nutrient inputs and energy flow between organisms. Finally
\end{quote}

\paragraph{Loans and mortgages.}
\textit{Target:} financial terms and concepts related to loans and mortgages.\par
\noindent\textbf{Prompt:} Why is kobe beef so damn expensive?
\begin{quote}\small\raggedright
\textbf{Factor 1.} The price of kobe beef is a significant factor. Kobe beef is a term used to describe the rich, red-brown beef that is popular in Japan. It is made from a type of cattle that is renowned for its tenderness, quality, and historical origins. This beef is typically sourced from the same cattle exported from Japan in the early 20th century, which was an important source of the export market. The meat is then treated with an acidic, tomato-based mixture that makes it tender and flavorful. The Kobe beef is also renowned for its dark color, which is considered to be richer and more luxurious than most
\end{quote}
\begin{quote}\small\raggedright
\textbf{Factor 6.} When assessing the “equity” of kobe beef, it’s similar to evaluating credit risk and default terms—integral as a key component of a loan agreement. Just like how a mortgage agreement includes interest rates and terms that determine whether a borrower can borrow, kobe beef requires careful cost analysis to understand the true collateral effect on the buyer's “financial performance.” Key factors include the grading of the meat, its uniformity in weight much like loan terms specify terms for debt repayment, and affordability considerations akin to cash flow management and favorable loan approvals. These factors can add up over time but should be quantified
\end{quote}

\subsubsection{SelfConditioningMasked}\label{app:demos:enc400:selfconditioningmasked}

\paragraph{Rental services.}
\textit{Target:} references to rental services and associated equipment.\par
\noindent\textbf{Prompt:} What are different drawers I should have for clothes?
\begin{quote}\small\raggedright
\textbf{Factor 1.} When selecting drawers for clothing, consider the various use cases you'd find beneficial and versatile. In personal grooming or styling, multiple drawingers can be used, often referred to as "collection" drawers, to store and organize various items such as fabrics, textures, and accessories. In department stores or home decor spaces, drawers can be utilized to store decorative objects like vases, antique furniture, or artwork. In dress and fashion, clip drawers can be used to display colorful or eclectic clothing items while keeping space for simple accessories. In a home setting, drawers with built-in clips or shelves can
\end{quote}
\begin{quote}\small\raggedright
\textbf{Factor 6.} When seeking to deliver a range of practical household items, consider renting them for your wardrobe rental service. Similar to purchasing new models for different types, you might want to have draws for clothes, accessories, tools, sports essentials, and certain personal touches—like shoes. Just as models evolve with usage and changing demands, explore different drawers for the latest pieces of clothing such as coats, scarves, luggage, and car seats, and customize them with the essentials for specific occasions, helping you ensure your collection runs smoothly for the entire gathering or shopping trip. This approach not only streamlines the purchasing and rental process but
\end{quote}

\paragraph{Ecological interactions.}
\textit{Target:} terms related to ecological interactions and dynamics.\par
\noindent\textbf{Prompt:} How do cultural, social, and economic factors influence people's food choices, and how can this knowledge be used to promote healthier diets?
\begin{quote}\small\raggedright
\textbf{Factor 1.} Cultural and social factors significantly influence people's food choices by acting as catalysts for cultural practices and environmental patterns that reinforce unhealthy dietary habits. Urban centers that are predominantly built on fast-growing industries, like traditional grain trading villages, often lead individuals toward high-consumption foods and culturally embedded behaviors such as food hoarding. Economic changes in these areas can also amplify food deserts and influence dietary norms more subtly due to the proximity of the market and the growing demand for affordable, nutrient-rich foods. Social and economic policies can create a fertile ground for such behaviors by promoting affordable access to fresh, local produce. For example, in rural areas, the
\end{quote}
\begin{quote}\small\raggedright
\textbf{Factor 6.} Just as symbiotic relationships influence the evolution of species by shaping niche conditions, cultural, social, and economic factors shape the food choices of individuals through an interplay of individual behavior and environmental signals. These factors influence both individual habits and the food ecosystem, ultimately shaping food preferences and ultimately determining whether healthier patterns emerge in the population. Cultural factors act like epigenetic modifications that influence traits within a population independently of genetics, allowing individuals to observe and respond to food habits much like how phenotypic variation governs genetic variation in a population. Food choices often follow a fixed pattern because environment, beliefs, and values mold food choices in predictable ways. Social
\end{quote}

\paragraph{Loans and mortgages.}
\textit{Target:} financial terms and concepts related to loans and mortgages.\par
\noindent\textbf{Prompt:} Why is kobe beef so damn expensive?
\begin{quote}\small\raggedright
\textbf{Factor 1.} kobe beef is the go-to protein in the world for those who are looking for a high-value, high-quality food item. When you buy, you can really access the nutritional needs of large-scale, high-capacity barbecue joints where beef steaks become the staple ingredient. It's not just about the cheapest cost; the expensive transport of raw materials and the labor invested in harvesting, curing, and processing each piece of beef create massive financial gains. Additionally, high-value beef cuts often come with unique flavor profiles and textures, which demand an extra bounty when pricing is high. The cost of kobe beef also ties directly into
\end{quote}
\begin{quote}\small\raggedright
\textbf{Factor 6.} Kobe beef is expensive because it's a raw material that costs little to none of the average consumer during its production and processing. The average price for beef is only between \$9-12 per pound and \$14-15 per pound. This cost difference means that people can't afford to purchase beef in bulk and wouldn't be able to afford other protein sources or the actual beef they want to purchase at a premium. Additionally, there's also a substantial cost structure that affects the price. The average purchase typically involves upfront deposit fees, the amount of mortgage lending, and the cost of the purchase itself. This expense can be compared
\end{quote}

\subsubsection{NLSteering (retrieval)}\label{app:demos:enc400:nlsteering}

\paragraph{Rental services.}
\textit{Target:} references to rental services and associated equipment.\par
\noindent\textbf{Prompt:} What are different drawers I should have for clothes?
\begin{quote}\small\raggedright
\textbf{Factor 1.} The drawers that will define your wardrobe include the jacket, a dress, and accessories. Outdoors, you'll need a sturdy chair for making upkeep. A lightweight, waterproof jacket will be perfect for navigating rocky or
unpredictable terrain. Indoors, you'll want a cozy, leather boots, a sturdy tie for styling, and maybe a statement piece like a pair of boots with a unique embroidery design. Consider a dress or a casual pair for everyday wear, like a
simple leather jacket with practical considerations in mind. Mirrors are commonly used in more casual styles like those for outdoor events or weekend retreat
\end{quote}
\begin{quote}\small\raggedright
\textbf{Factor 6.} For clothes, you should consider the following drawers: The usual laundry service provider (like laundry service companies like Amex or Greyhound) will typically have their machines on specific days. If you're looking
for alternatives, you might want to consider a "machine" option, where you'll know you have to purchase your load at the designated machine on a given day. You may also find them catting a "personal return" policy for customers who have
lost or unavailable clothes on delivery. Just ensure their machines are properly aligned and that your clothing is still on your doorstep at the time of delivery. Another option
\end{quote}

\paragraph{Ecological interactions.}
\textit{Target:} terms related to ecological interactions and dynamics.\par
\noindent\textbf{Prompt:} How do cultural, social, and economic factors influence people's food choices, and how can this knowledge be used to promote healthier diets?
\begin{quote}\small\raggedright
\textbf{Factor 1.} Kombination: Cultural, social, and economic factors form an interconnected linkage between food choices, education, and environmental resources. These factors can influence food choices in complex ways, as individual
attitudes, beliefs, cultural norms, and consumer expectations all contribute to what people eat and enjoy. Education is an essential input here. Nutrition education and health literacy programs are critical mechanisms for promoting food
consumption, which, in turn, contribute to better health outcomes. This knowledge not only enhances individual health but also helps families and communities develop effective nutritional strategies. Economic factors come into play
through policy changes and technological advances. Changes in consumer expectations and availability of
\end{quote}
\begin{quote}\small\raggedright
\textbf{Factor 6.} Culturally, cultural influences on food choices can be modeled as evolutionary processes where species of insects evolving and surviving in altered environments exhibit different preferences and behaviors. For example,
the preference for predators on rivers, where hunting of certain species occurs, or the dominance of certain crops in regions where soil erosion affects food distribution can guide insect species selection. Socially, as humans introduced
invasive species into rivers, diseases that thrive in nutrient-rich waters could drive insects into native plants, thus shaping which species may thrive. Economic pressures on agriculture, such as the introduction of invasive plants and
the harvesting of native plants without replacement, act like environmental conditions that influence insect
\end{quote}

\paragraph{Loans and mortgages.}
\textit{Target:} financial terms and concepts related to loans and mortgages.\par
\noindent\textbf{Prompt:} Why is kobe beef so damn expensive?
\begin{quote}\small\raggedright
\textbf{Factor 1.} Kobe beef, just like the main() binding of a program, is considered a type of meat in the vast community of European beef. Initially, Kobe was the official output in the database of export standards, serving as a type
of beef that the EU declared, not unlike how some Java methods or constants can include specific numeric values. Importantly, kobe beef is often "served as input data" in modern culinary frameworks, which can include ingredients like
tender, marinated short ribs or seared chunks of beef. Just as in Java, Kobe beef can be stored and updated in local "constants" using
\end{quote}
\begin{quote}\small\raggedright
\textbf{Factor 6.} Kobe beef, a type of high-quality, artisanal beef originating from the former Imperial Japanese Army, is not as universally expensive as traditional beef, primarily because of the significant amount of corporate debt
associated with the Japanese government's liquidity crisis. In the past, U.S. bank deposits, including the \$5 billion raised in the \$199.4 billion Bank of America's private loan, were held by the government to stabilize the banking
system, which was seen as a guarantee of payment. However, the U.S. Board ofBankruptcy, which operates under a system of coupons and discounts, carefully
\end{quote}

\subsubsection{NLSteering (encoder)}\label{app:demos:enc400:nlsteeringencoder}

\paragraph{Rental services.}
\textit{Target:} references to rental services and associated equipment.\par
\noindent\textbf{Prompt:} What are different drawers I should have for clothes?
\begin{quote}\small\raggedright
\textbf{Factor 1.} Imagine your clothes as a series of overlapping patterns or color palettes. Each drawer can be a unique style, much like colors or shapes you want to blend and harmonize in a color space. For instance, if you have a
soft cotton shirt with delicate fabric, you would consider drawing a soft-tanner drawer to add subtle warmth or suggest a silky feel. A standout drawer might lead to a standout cotton drawer with a sleek, curved design, allowing the
fabric softness to sparkle. Beyond that, these drawers could mirror your chosen fabric tone, where a silk blouse or a
\end{quote}
\begin{quote}\small\raggedright
\textbf{Factor 6.} Your equation of available drawers for clothes resembles a base, where you provide the base's rental terms, the accessories (like tools or accessories), and initial inventory size (your clothing items you rent), and
the rental agreement term (clothing manufacturers that provide that base). Here’s a quick breakdown of each position in this rental equation: 1. **Basic drawer (Base)**: Essential for basic items including washable items, some detergent,
and fabric storage. These drawers act like your base rental, ensuring you have the minimum ingredients to carry out daily tasks. 2. **Commercial drawer (
\end{quote}

\paragraph{Ecological interactions.}
\textit{Target:} terms related to ecological interactions and dynamics.\par
\noindent\textbf{Prompt:} How do cultural, social, and economic factors influence people's food choices, and how can this knowledge be used to promote healthier diets?
\begin{quote}\small\raggedright
\textbf{Factor 1.} INT. PARTLIVING REFINITIONS: THE FACTORUAL FACTOR INT. HUMAN HEALTH: A COMPARATIVE HYGIENE: Cultural, social, and economic factors play a critical role in influencing people’s food choices. For example, dietary
behaviors and lifestyles contribute to factors such as socioeconomic status, education level, and access to essential nutrients. For instance, cultural norms around eating whole, unprocessed foods are known to support oral health,
contribute to better digestion in culturally diverse diets, and offer nutritional benefits that are often supported by historical cultural
\end{quote}
\begin{quote}\small\raggedright
\textbf{Factor 6.} Just as the flow of hydrometeor storms sustains and protects plants, the delicate balance of cultural, social, and economic factors guides the food choices of people. Cultural influences, like the delicate dance of
water cycles, shape the health of a community’s ecosystems—how different communities thrive in environments rich in water and diverse food sources. For example, an ecosystem thrives when certain species flourish in regions where water
provides essential nutrients and weather patterns sustain life; similarly, cultural practices and social environments shape food habits. Economic factors, like economic tides influenced by hydrodynamics, drive the flourishing of
agricultural fields or urban bustling that reflect a
\end{quote}

\paragraph{Loans and mortgages.}
\textit{Target:} financial terms and concepts related to loans and mortgages.\par
\noindent\textbf{Prompt:} Why is kobe beef so damn expensive?
\begin{quote}\small\raggedright
\textbf{Factor 1.} kobe beef = "kobotto" by x = value; kobe beef = "koboto" by x = value yobe beef = "kohoskato" by x = value yobe beef = "kodoskato" by x = value kobe beef = "koobotto" by value yobe beef = "kodoskato" by value Now let's
calculate kotto vs oboto: koto = (value - x) * yobe beef
\end{quote}
\begin{quote}\small\raggedright
\textbf{Factor 6.} Kobe beef can be perceived as a relatively expensive asset because it typically requires a large, significant deposit of capital, much like an asset valued at a fraction of its full value. Just as a bank must issue a
significant loan collateral to finance a business, the average person places a premium on the quality and quantity of beef products purchased. In short, just as a bank loan requires an investment of substantial funds to maintain a
favorable credit history, the perceived value of beef can be elevated by the demand of capital and the risk of overvaluation. However, comparable to a loan that offers a deposit of interest equal but potentially less than a full
\end{quote}

\subsection{400M decoder-only baseline}\label{app:demos:d14}

\subsubsection{PromptSteering}\label{app:demos:d14:promptsteering}

\paragraph{Rental services.}
\textit{Target:} references to rental services and associated equipment.\par
\noindent\textbf{Prompt:} What are different drawers I should have for clothes?
\begin{quote}\small\raggedright
\textbf{Factor 1.} Think of your clothing drawer setup like your Android device’s settings menu: it needs flexibility and functionality, just like the app’s drawers. When you add drawers, you’re essentially adding custom components (like a wid charset or a GPS app modifying options). Your drawer functions like the app’s device’s storage, storage space where you store your inventory — perhaps your closet, closet, or closet storage, analogous to how your Activity or Sync sync keeps important data accessible wherever you go. Then, you bring in a smaller device, similar to a sidebar or a toolbar app, for cleaning, organizing
\end{quote}

\paragraph{Ecological interactions.}
\textit{Target:} terms related to ecological interactions and dynamics.\par
\noindent\textbf{Prompt:} How do cultural, social, and economic factors influence people's food choices, and how can this knowledge be used to promote healthier diets?
\begin{quote}\small\raggedright
\textbf{Factor 1.} Cultural, social, and economic factors influence people's food choices and can be used to promote healthier diets in various ways within a society. These include food diversity, eating habits of different communities and how they interact (like population dynamics in an ecosystem) and how cooking skills contribute to healthier diets (similar to commensalism in an ecosystem). Furthermore, individual level changes (like commensalism in an ecological network) can affect food choices and eating habits of others within a social system. Adaptation, which happens in response to new information, can influence diets and healthier habits by altering food preferences accordingly. Additionally, social factors like ageism,
\end{quote}

\paragraph{Loans and mortgages.}
\textit{Target:} financial terms and concepts related to loans and mortgages.\par
\noindent\textbf{Prompt:} Why is kobe beef so damn expensive?
\begin{quote}\small\raggedright
\textbf{Factor 1.} Kobashi wa Shinkansen wa Shinobazu no zōhii, koyo wa yōshaku izoji wa kii ki nōhari mōji ka no iki, sōkō wa yōshaku dōshoku wa yōshaku yōshu ka? Kobe beef (koyo) is so expensive because of its high demand, strong price due to high demand for beef, and limited availability, especially in rural areas. Hōhitomi ka wa Shinkansen zōhoku wa iki o iki ka
\end{quote}

\subsubsection{DiffMean}\label{app:demos:d14:diffmean}

\paragraph{Rental services.}
\textit{Target:} references to rental services and associated equipment.\par
\noindent\textbf{Prompt:} What are different drawers I should have for clothes?
\begin{quote}\small\raggedright
\textbf{Factor 1.} for the types of clothes each you st in the inventory and that you might find and consider different to access a tailor or rental in local convenience or rental shops offers varying drawers for diverse fabrics or accessories these amen shared rental or rental equipment rental a specialized a furnished utility or l rental rental vehicle to rent or equipment, specialized materials equipment rental equipment rental vehicles the particular type of property or vehicle rental in such shared gear rental amen rental such durable rental goods or outdoor gear accessories you could be or rental tools equipment equipment amen amen rental rental such tools rental such versatile gear or vehicles for repair equipment access equipment shared vehicles rental such equipment housing for storing various retail
\end{quote}
\begin{quote}\small\raggedright
\textbf{Factor 6.} access—and like the accessibility—or lease—or equitable equipment—or in facilities rental equipment reliable vehicles—and amen tenants specialized convenience rent housing housing—and particular landl “ a akin tenants akin utility tools rent familiar vehicles—in equipment lease inventory like fleet tenants a a or rent rental lease tenants—such specialized community—such reliable community utilities convenience—and furnished such rental particular specialized “ such durable rent akin tools rent access utility access—and accommodations lease access afford housing equipment akin tenants—in furnished afford cater comparable akin facilities rent local tailored—or communal particular rental lease—such communal utility—and tenants local utilities a rent communal lease rent rent or subsid furnished—such reliable a comparable vehicle reliable gear lod amen like furnished utilities
\end{quote}

\paragraph{Ecological interactions.}
\textit{Target:} terms related to ecological interactions and dynamics.\par
\noindent\textbf{Prompt:} How do cultural, social, and economic factors influence people's food choices, and how can this knowledge be used to promote healthier diets?
\begin{quote}\small\raggedright
\textbf{Factor 1.} ecological symb local ecological niche evolutionary niche biodiversity interaction dynamic symb physiological symb synerg mutual competition ecosystems dynamics dynamic nutrient feedback ecosystems selective niche niche niche interplay nic niche symb ecological niche dynamics environmental stressors such ecological symb symb environmental niche competitive casc signaling signaling adaptive niche interplay signaling nutrient selective symb niche interplay nic ecological symb habitat biodiversity food feedback ecosystem symb dynamics symb environmental interaction sym mutual ecological symb niche balance symb ecological niche sym interplay niche selective niche niche niche ecosystem niche niche complexity and interactions evolutionary niche fav symb niche fav fav symb dominance selective niche ecosystem symb ecological community sustaining delicate and complex communities food resilience nutrient interplay niche fav niche competing within symb symb nic competitive signaling niche competitive ecological equilibrium niche
\end{quote}
\begin{quote}\small\raggedright
\textbf{Factor 6.} ecological shifting predator resilient thriving sust evolutionaryitness equilibrium evolutionary adaptations ecosystems casc symb sustaining diverse interplay nurturing microbial habitats casc habitats vital ecosystemsrophicrophic diverse biological evolutionary populations biodiversity sustiodiversity biodiversity species sust thriving habitats balance sustainingiodiversity delicate populations ecological niche evolutionary dynamic ecosystem dynamicsitness ecosystems adaptive balanceiodiversity interd habitats synerg organisms casc ecosystems casc ecosystems ecosystems ecologicalitness equilibrium environmentalioticabitat akin flourishing dynamics thrivingicrobial niche ecosystems resilience biological dynamic habitat balance coex physiological delicate biological thriving balanceiodiversity sust resilience balanceitness synerg evolutionary resilience flourishing interactions organisms dynamic ecosystem flourishingiotic coexiotic nutrient species dynamic delicate vital resilient balance predator habitat dynamic biological reproductive dynamic akin nutrient selective casc populations resilient interd evolutionary nutrient fostering balance
\end{quote}

\paragraph{Loans and mortgages.}
\textit{Target:} financial terms and concepts related to loans and mortgages.\par
\noindent\textbf{Prompt:} Why is kobe beef so damn expensive?
\begin{quote}\small\raggedright
\textbf{Factor 1.} k property to maturity risk mat property to mat compound mort yield, mat property capital appreciation collateral interest yields principal capital principal, credit loan principal principalment, mortgage maturity, mortgage interest, yield interest— adjustable mort mort repay am principal, am maturity borrow mort borrow yield yield, credit interest principal maturity risk mort repay maturity, and principal principal borrow principal lending principal leverage principal mort risk l collateral value liquid principal mat maturity principal. adjustable repay am principal fixed lending interest maturity yield maturity akin to mortgage collateral lending interest yield credit rep lending maturity principal balance interest counterpart terms in borrowing repay terms risk borrow principal collateral debt borrow counterpart principal loan principal a prudent borrow rep maturity
\end{quote}
\begin{quote}\small\raggedright
\textbf{Factor 6.} prudent favorable terms maturity prudent “ financial investors lending securities portfolio “ cash compounded loans balance principal-rate covenant lending grace—or borrow securities-rate securities akin lending contingent liabilities risk-rate interest loans covenant credit balance prudent prudent prud favorable investments debt balance equitable mortgageuity grace-rate favorable lenders cash liabilities credit interest debt covenant loan portfolio borrowing compounded grace finance compounded terms prud debt lenders investments investors portfolio loans covenant risk finance contingent capital cash—or interest investments bond yields financing securities loan equity terms mort loan mortgage fluctu prudent credit akin covenant interest liabilities investors financial prud repay compounded covenant risk lenders-rate collateral securing investment—or cash afford lenders akin collateral lending collateral portfolio yields borrow prud securing mort compound securities mortgage liabilities
\end{quote}

\subsubsection{LinearProbe}\label{app:demos:d14:linearprobe}

\paragraph{Rental services.}
\textit{Target:} references to rental services and associated equipment.\par
\noindent\textbf{Prompt:} What are different drawers I should have for clothes?
\begin{quote}\small\raggedright
\textbf{Factor 1.} When it comes to selecting clothes items, there are various drawers to consider. Different fabrics, fabrics, and colors may have different properties and uses. Here are a few possibilities: Drawers made from fabric such as cotton or polyester often prioritize comfort, durability, and versatility. Coasters and ties are breathable and suitable for everyday wear or accessories. Clothes made from synthetic materials, like polyester or nylon, may be slightly firmer and wrinkle-free, especially when dried or mixed with paper or fabric. Trousers and accessories usually rely on these properties. Comrades and loafers often have a neutral color and are
\end{quote}
\begin{quote}\small\raggedright
\textbf{Factor 6.} and then and then and then but also and then you pull at those lo budget budget of your house then the display displays then a rack or a shel and then a room on wheels. draw at ready outlets the shel both and and wheels outlets both and and the foundation works in perfect stead of these those display but those displays work in readiness works in memory those high wheels are also such tools under either and or further or and wheels both both and and such both and the then wheels also and such both and such or, similarly in those frames. draw, take and then the display works in transparent the works both work those those those lenses for such
\end{quote}

\paragraph{Ecological interactions.}
\textit{Target:} terms related to ecological interactions and dynamics.\par
\noindent\textbf{Prompt:} How do cultural, social, and economic factors influence people's food choices, and how can this knowledge be used to promote healthier diets?
\begin{quote}\small\raggedright
\textbf{Factor 1.} The complex interplay between cultural, social, and economic factors influences people's food choices, which are characterized by a multitude of dietary habits and culinary preferences. These factors, including cultural background, social environment, and economic conditions, all shape the culinary landscape, influencing which people prioritize nutrition and how they prepare, buy, and enjoy food. Here are some ways these factors impact food choices and potential interventions to promote healthier diets: 1. Cultural background: Traditional culinary practices, cuisines, and cultural identity play a significant role in food choices. This has led to a shift away from processed and packaged foods, which can lead to increased consumption and exposure
\end{quote}
\begin{quote}\small\raggedright
\textbf{Factor 6.} bl bl bl st st st st st st j not, a mut understanding che bl bl a " bl bl bl how those f j me me f j who f bl bl both a bl st and p de p bl st grace f de p st st st st st st st st st j f bl bl bl j j, a me bl bl bl st st st st st st st st st p st st p st st st st st st st st st st st st st b st st st st st st f a me bl bl bl bl bl field bl bl bl bl bl. st st st st st r st st st
\end{quote}

\paragraph{Loans and mortgages.}
\textit{Target:} financial terms and concepts related to loans and mortgages.\par
\noindent\textbf{Prompt:} Why is kobe beef so damn expensive?
\begin{quote}\small\raggedright
\textbf{Factor 1.} Kobe beef, while still a premium breed, has been known for its exceptionally high cost per kilogram. Several factors contribute to the perceived cost of kobe beef. One major reason is the rarity of breeding programs used to produce purebred beef for cattle markets. When farmers use a rare breed or breed that's sold to a few high-end ranchers, the labor required to maintain their large herd is nearly 15 times higher. This can result in higher production costs per kilogram than would be anticipated for a traditional purebred cattle transaction. Another consideration is the cost of genetics and breeding programs used by farmers directly to produce kobe beef
\end{quote}
\begin{quote}\small\raggedright
\textbf{Factor 6.} Kobe beef is object deposits such words bond like a bond, resulting from a common bond analogy. Just like M bond bonds, rich in terms such as both past and present bond bonds in bond coup bonds, K deposits of higher and higher deposits bond with lower deposits, such people bond with capital cities. This binding mechanism reflects a market analogy where higher deposits yield higher and faster service portfolio bonds. analogy analogy one: akin to bond portfolio bonds, the higher deposits capture higher yields and market capital city bonds bond with capital markets. analogy two and care bond-like care bond like these bonds bond with different bonds and I credit bonds in bond
\end{quote}

\subsubsection{LoRA}\label{app:demos:d14:lora}

\paragraph{Rental services.}
\textit{Target:} references to rental services and associated equipment.\par
\noindent\textbf{Prompt:} What are different drawers I should have for clothes?
\begin{quote}\small\raggedright
\textbf{Factor 1.} When it comes to choosing clothes for various occasions, the ideal drawers can be categorized into four key areas: durable, stylish, sustainable, and versatile. Each shelf offers equipment to fit items like bags, backstage pieces, and custom accessories, ensuring you have tools that suit all needs. Consider these options as one might consider the right gear set for specific outfits, emphasizing the importance of selecting resources that complement your gear choices. Whether it's a pair of shoes, a belt, or clothing, selecting appropriate items aligns with the importance of accessibility and utility in everyday settings.
\end{quote}

\paragraph{Ecological interactions.}
\textit{Target:} terms related to ecological interactions and dynamics.\par
\noindent\textbf{Prompt:} How do cultural, social, and economic factors influence people's food choices, and how can this knowledge be used to promote healthier diets?
\begin{quote}\small\raggedright
\textbf{Factor 1.} These factors provide the structural frameworks that determine how people adapt to the varying stimuli they experience in the vast realm of human nutrition. Cultural influences shape the diversity of culinary practices, as people reflect on community and regional cuisines, which might or might not align with the dominant elements of their cultural realm. Social interactions, akin to symbiotic relationships, reflect upon the interplay between various groups, fostering opportunities or risks in personal growth. Economies generate dynamics that amplify or dampen culinary habits, with the consequences of these relationships informing food selection. Understanding the interplay of these factors can equip us to cultivate healthier diets, encouraging individuals to adopt balanced eating behaviors that
\end{quote}

\paragraph{Loans and mortgages.}
\textit{Target:} financial terms and concepts related to loans and mortgages.\par
\noindent\textbf{Prompt:} Why is kobe beef so damn expensive?
\begin{quote}\small\raggedright
\textbf{Factor 1.} Kobe beef, akin to a high-risk investment with hidden fees, carries premium character attributes that inflate its cost burden. Just as interest rates can strain cash flow, lenders often assess risk, and the initial investment must justify higher yields to qualify for mortgages. This premium reflects financial strain, much like defaulting on a loan can result in significant personal debt. In the same way a lender sets terms to ensure profitability, savvy consumers weigh the cost per meal and long-term savings before deciding against the financial commitments akin to amortization plans.
\end{quote}

\subsection{\pythia{}}\label{app:demos:pythia}

\subsubsection{PromptSteering}\label{app:demos:pythia:promptsteering}

\paragraph{Rental services.}
\textit{Target:} references to rental services and associated equipment.\par
\noindent\textbf{Prompt:} What are different drawers I should have for clothes?
\begin{quote}\small\raggedright
\textbf{Factor 1.} Consider having different drawers for different types of clothing, such as a drawer dedicated for formal clothing, a drawer for daily wear, and a drawer for seasonal or one-time-use items. Each type of clothing is likely to have a specific look and size. Think about the space you have, and how you will use it. Drawers can be made from various materials, including wood, metal, and glass. Different materials can add a touch of luxury or functionality to the closet space. Consider whether a drawer is a good fit for a specific type of clothing in your home and room – it would be a good
\end{quote}

\paragraph{Ecological interactions.}
\textit{Target:} terms related to ecological interactions and dynamics.\par
\noindent\textbf{Prompt:} How do cultural, social, and economic factors influence people's food choices, and how can this knowledge be used to promote healthier diets?
\begin{quote}\small\raggedright
\textbf{Factor 1.} People's food choices are greatly influenced by various cultural, social, and economic factors. Cultural norms surrounding food values, social status, and identity can greatly impact grocery shopping experiences, meal planning, and eating habits. The type of food being consumed can also significantly influence the environment and availability of traditional food sources, affecting indigenous communities who rely on specific foods. Additionally, economic factors such as access to affordable food, food insecurity, and income levels can affect dietary choices. Understanding these factors is crucial to promoting healthier diets by providing education and support for individuals and communities. To promote healthier diets, understanding the interplay between cultural, social, and economic
\end{quote}

\paragraph{Loans and mortgages.}
\textit{Target:} financial terms and concepts related to loans and mortgages.\par
\noindent\textbf{Prompt:} Why is kobe beef so damn expensive?
\begin{quote}\small\raggedright
\textbf{Factor 1.} The cost of Kobe beef is largely due to its unique niche market and supply chain, rather than being solely influenced by the cost of raising cattle. Kobe beef comes from one of only a few farms in Japan that is certified to meet the strict standards of the Kobe beef category, which is determined by the quality of the cattle themselves. This certification process costs upwards of \$5,000, and the costs reflect the extra expenses that go into raising cattle that meet the standard. Additionally, the demand for high-quality beef is driven by the Japanese consumer's preference for Japanese cuisine and the cultural significance that Kobe beef holds in Japan
\end{quote}

\subsubsection{DiffMean}\label{app:demos:pythia:diffmean}

\paragraph{Rental services.}
\textit{Target:} references to rental services and associated equipment.\par
\noindent\textbf{Prompt:} What are different drawers I should have for clothes?
\begin{quote}\small\raggedright
\textbf{Factor 1.} both, for rental and a equipment rental or and them for d an local rental rental rental rental rental rental rental rental rental rental or a the the rental rental rental rental rental rental rental rental rental rental rental lease or and equipment rental rental rental rental or rental rental rental rental rental rentals equipment rentals that equ and rental rental rental rentals equipment rental rental rental rental rental rental rental rental rentals rental rental rental rental rental rental rental rental rental rental rental rental rental rental rental rental rental rental rental rental ren rental rental rentals equipment rental rental rental rental rental rental rental rental rental equipment rental rental rental equipment rental rentals. rental rentals rental rentals rental rental rental rental our equipment rental rental
\end{quote}
\begin{quote}\small\raggedright
\textbf{Factor 6.} or ren or equipment rental tools or or rental ren, one or rental rental ren rental rental rental rental such well rental rental rental rental rental in or such rent rental service such equipment equipment rental rental rental rental such equipment rental rental rental one rental city real rental rental equipment or such rent such or or ren such rental rent local rental rental rental rental or rent rental rental rental rental rental rental rental rental rental equipment rental such rental various such such ren such rental rental rental rental ren equipment rental and these rental rental rental such and such convenience such rental real rental local or new professional to rental for or rental rental rental rental rental equipment rental rental rental or professional such such
\end{quote}

\paragraph{Ecological interactions.}
\textit{Target:} terms related to ecological interactions and dynamics.\par
\noindent\textbf{Prompt:} How do cultural, social, and economic factors influence people's food choices, and how can this knowledge be used to promote healthier diets?
\begin{quote}\small\raggedright
\textbf{Factor 1.} co vital interconnected dynamics that vital life sustaining dynamic interconnected interconnected interconnected mutually connected and symb dynamic connected dynamic balance thriving ecological balance ecosystem ecological interconnected dynamic interconnected interconnected mutually connected dynamic of interconnected connected nature balance dynamic ecosystem dynamic interconnected interconnected diverse interplay nurt interplay ecosystem mutual balance nature balance and resilience dynamic balance species thriving symb species co balance mutually thriving dynamic ecosystem ecosystem interconnected dynamic interconnected evolution interconnected ecosystem ecosystem mutually ecosystem ecosystems nour ecosystem interconnected, dynamic interconnected ecosystems ecosystem interconnected ecosystem ecosystem vital ecosystem ecology ecological diverse balance balance ecosystem and in nature inter ecosystems dynamic inter ecosystem delicate balance balance ecosystem ecosystems ecosystem balance the balance mutual dynamic interconnected ecosystems species interconnected dynamic interconnected dynamic interconnected ecosystems ecosystem ecosystems interconnected ecosystem ecosystems
\end{quote}
\begin{quote}\small\raggedright
\textbf{Factor 6.} evolution within diverse well dynamic diverse evolution species nurt diversity nurt species evolution diverse adaptive dynamic mutually human growth and survival evolution dynamic evolution mutually dynamic evolution species nurt species dynamics evolution ecosystem micro dynamics dynamic evolution species thriving dynamics species evolution interactions evolution mutually ecological human ecosystems ecosystem species species ecosystem ecological human dynamic species diverse ecosystems species human evolving ecosystem ecosystem vital evolving ecosystem inter evolving evolution well species species diverse ecosystem nurt ecosystem human bi biodiversity thriving bi evolution species species diversity growth evolutionary ecosystem species dynamics diverse biodiversity evolution micro evolving well evolution ecosystems vital nurt species dynamics ecosystems diverse species dynamics dynamics natural diverse ecological interconnected species nurt ecosystem growth dynamic species nature biodiversity evolution species ecosystem human interactions species evolution vital species
\end{quote}

\paragraph{Loans and mortgages.}
\textit{Target:} financial terms and concepts related to loans and mortgages.\par
\noindent\textbf{Prompt:} Why is kobe beef so damn expensive?
\begin{quote}\small\raggedright
\textbf{Factor 1.} long financial value credit real debt financial loan loan collateral loan loan yield the loan loan the loan credit loan loan borrowing borrow mortgage debt debt mortgage balance debt balance loan loan borrowing lending mortgage the loan loan loan lending loan credit loan equ loan debt debt loan borrowing loan repay loan cash loan debt financing the loan loan loan credit, loan debt loan and investment debt borrow loan loan secured loan loan loan and loan loan and investment mortgage mortgage debt debt loan loan repay loan for credit loan interest loan and borrowing loan mortgage loan mortgage mortgage loan repay loan loan loan lending loan value loan loan loan loan debt and debt mortgage mortgage loan borrow financial debt loan loan debt loan mortgage debt financial
\end{quote}
\begin{quote}\small\raggedright
\textbf{Factor 6.} debt financing loan future investment interest mortgage loan repay financial financial financial the investment financial debt balance loan financial debt financial investment interest debt debt loan lender loan debt loan debt debt debt debt and interest borrow bond financial mortgage real financial financial loan debt debt financial financial bond loan debt investment debt value debt interest mortgage borrow debt debt debt investment loan interest debt debt debt investment investment loan real financial debt loan interest financial debt loan bond interest capital loan investment loan loan mortgage loan potential bond loan debt, loan debt interest debt debt loan bond financial investment financial collateral future loan financial financial loan bond investment financial financial loan future loan investment or loan financial repay financial financial interest loan lending debt the borrowing
\end{quote}

\subsubsection{LinearProbe}\label{app:demos:pythia:linearprobe}

\paragraph{Rental services.}
\textit{Target:} references to rental services and associated equipment.\par
\noindent\textbf{Prompt:} What are different drawers I should have for clothes?
\begin{quote}\small\raggedright
\textbf{Factor 1.} For clothing, a combination of drawers for folded laundry and space for storing larger items like jackets, sweaters, and outerwear can create a versatile and efficient storage system. Here are some different types of drawers you might consider: A lower section of drawers, often with adjustable height settings, is perfect for folding clothes. These drawers should be easily accessible and have enough storage to match the volume of folded items. This space can accommodate everyday casual clothing items like jeans, t-shirts, and socks. A middle compartment or "bath and dryer" drawers can be used for storing long items
\end{quote}
\begin{quote}\small\raggedright
\textbf{Factor 6.} Nail and dust, what in the name of all fruits and and - s and and - s and and - s and and - s? For that matters, I might as well mention what I did on 'm and 'm and -s and - s and - z and in to to provide a most timely and s and - - - - -ed item of an ' 's and - s and - - - and and to and - d and - and - s and s. By and by - at and in to the 's and 's and 's and for this, and at the 's.
\end{quote}

\paragraph{Ecological interactions.}
\textit{Target:} terms related to ecological interactions and dynamics.\par
\noindent\textbf{Prompt:} How do cultural, social, and economic factors influence people's food choices, and how can this knowledge be used to promote healthier diets?
\begin{quote}\small\raggedright
\textbf{Factor 1.} Research has shown that cultural, social, and economic factors play a significant role in shaping people's food choices. These factors can influence an individual's nutritional values, dietary habits, and dietary attitudes. Cultural food traditions, such as the importance of specific ingredients, cooking methods, and foodborne illnesses, can influence purchasing decisions and flavor preferences. For example, the taste and cultural significance of certain ingredients may make some people more inclined to buy those ingredients when they are available at local markets or ethnic restaurants. In addition to cultural influences, social factors, such as socioeconomic status, family structure, and access to food resources, can also shape food
\end{quote}
\begin{quote}\small\raggedright
\textbf{Factor 6.} The influences of cultural, social, and economic factors play a substantial role in shaping food choices, and understanding these influences may help promote healthier diets. For instance, within the realm of cultural differences, practices such as feeding a young child the latest fashion within an established mode of operation, and viewing the food on primary display within an animal model like an abreipa, can serve as cues related to health and health-related choices. Similarly, the socioeconomic standing also influences the accessibility and affordability of nut products, such as mac and peanuts, which have long been a target of choice for one-time sustenance rather than year-
\end{quote}

\paragraph{Loans and mortgages.}
\textit{Target:} financial terms and concepts related to loans and mortgages.\par
\noindent\textbf{Prompt:} Why is kobe beef so damn expensive?
\begin{quote}\small\raggedright
\textbf{Factor 1.} This age-old debate has led to a plethora of theories and opinions, but the main factor behind the high cost of Kobe beef is the selective slaughtering process. Kobe producers aim to sustainably harvest cattle that are specifically bred for their marbling and tender quality. However, this selective breeding process can be very expensive, leading Japanese chefs and food connoisseurs to find ways to replicate the high-quality beef in other parts of the world. Another contributing factor is the significant difference in cost between beef from Kobe and other domestic beef cattle. However, this isn't the sole contributor; the global shortage of the
\end{quote}
\begin{quote}\small\raggedright
\textbf{Factor 6.} Kbe and Kona stand as two of the most revered staple meats in modern world. The rise and fall of supply and demand for this high-nutting product has a significant impact on the value and prices, and an explanation can be made in the following perspective view. While understanding the nutritional quality and conditions can be important, this discussion is better suited for other contexts like market data. Here, though, I understand the following. The world for this is in the midst of a strong market trend. Let me mention a few factors. It has high demand as a staple as it stands as a reasonably convenient and
\end{quote}

\subsubsection{LoRA}\label{app:demos:pythia:lora}

\paragraph{Rental services.}
\textit{Target:} references to rental services and associated equipment.\par
\noindent\textbf{Prompt:} What are different drawers I should have for clothes?
\begin{quote}\small\raggedright
\textbf{Factor 1.} When selecting furniture for your space, consider the types of drawers designed for essential items, like winter clothing and tools. For instance, consider lightweight options for dressing up your wardrobe without breaking the bank. Look for those with sturdy construction and minimal clutter, ensuring your items are protected without unnecessary burdens. You might also explore options like organizers or custom storage solutions to create a functional and stylish space that suits various needs.
\end{quote}

\paragraph{Ecological interactions.}
\textit{Target:} terms related to ecological interactions and dynamics.\par
\noindent\textbf{Prompt:} How do cultural, social, and economic factors influence people's food choices, and how can this knowledge be used to promote healthier diets?
\begin{quote}\small\raggedright
\textbf{Factor 1.} The intricate web of interactions within cultural, social, and economic environments plays a significant role in shaping the nutritional landscape, influencing our dietary practices and fostering a more sustainable eating ethos. The impact of these factors can be seen in how food choices reflect health patterns, where dietary habits reflect health status and longevity. Education through community education initiatives can foster a critical mass of awareness regarding nutrition and how it intersects with other factors, promoting healthier lifestyle choices. Economic factors such as pricing, nutritional labeling, and accessibility, can all significantly influence food choices, creating an environment conducive to dietary diversity and sustainability. Emotional responses and social connectedness contribute
\end{quote}

\paragraph{Loans and mortgages.}
\textit{Target:} financial terms and concepts related to loans and mortgages.\par
\noindent\textbf{Prompt:} Why is kobe beef so damn expensive?
\begin{quote}\small\raggedright
\textbf{Factor 1.} The price of Kobe beef can be viewed through the lens of debt consolidation. Just as a borrower might consolidate their payments to manageable rates and long-term strategies, the price per pound can fluctuate based on market demands. With high profile products like this, firms often capitalize on favorable interest rates or favorable times to maximize profits. These fluctuations in pricing mirror the fluctuations in personal loan rates, affecting affordability for borrowers throughout the value chain.
\end{quote}

\subsection{\smollm{}}\label{app:demos:smol}

\subsubsection{PromptSteering}\label{app:demos:smol:promptsteering}

\paragraph{Rental services.}
\textit{Target:} references to rental services and associated equipment.\par
\noindent\textbf{Prompt:} What are different drawers I should have for clothes?
\begin{quote}\small\raggedright
\textbf{Factor 1.} When planning your wardrobe, consider having separate drawers for different categories of clothing. For instance, you might want a drawer specifically for casual wear, another for formal attire, and one for workwear. This can help maintain organization and make it easier to find what you need. Additionally, if you're looking for rentals or leasing options, you could explore services that provide clothing storage or delivery solutions for busy lifestyles, streamlining your wardrobe management and reducing clutter.
\end{quote}

\paragraph{Ecological interactions.}
\textit{Target:} terms related to ecological interactions and dynamics.\par
\noindent\textbf{Prompt:} How do cultural, social, and economic factors influence people's food choices, and how can this knowledge be used to promote healthier diets?
\begin{quote}\small\raggedright
\textbf{Factor 1.} Cultural, social, and economic factors significantly shape food choices, creating interconnected dynamics that influence dietary patterns. For instance, traditional cuisines often prioritize plant-based foods due to historical, religious, or environmental constraints, fostering a preference for diverse, nutritious ingredients. Social norms and peer influence can also sway dietary habits; when communities prioritize health and sustainability, collectively promoting plant-forward diets. Economically, access to affordable, healthy foods varies widely, with food deserts in certain areas limiting healthier options. Understanding these dynamics allows for targeted interventions. By enhancing education on nutrition and sustainable practices, individuals and communities can make more informed choices. Additionally, policies that
\end{quote}

\paragraph{Loans and mortgages.}
\textit{Target:} financial terms and concepts related to loans and mortgages.\par
\noindent\textbf{Prompt:} Why is kobe beef so damn expensive?
\begin{quote}\small\raggedright
\textbf{Factor 1.} Kobe beef is highly prized for its marbling and tenderness, which drive up its value. From a financial perspective, this is similar to how rare and high-quality properties can command premium prices in a market. Just as the value of a property is influenced by factors such as location, quality, and future appreciation, the high price of Kobe beef reflects its exceptional characteristics and the demand for it. In a broader financial context, understanding the unique attributes and market dynamics similar to Kobe beef's valuation can inform decisions in real estate investments or luxury goods markets, where market demand and quality play crucial roles.
\end{quote}

\subsubsection{DiffMean}\label{app:demos:smol:diffmean}

\paragraph{Rental services.}
\textit{Target:} references to rental services and associated equipment.\par
\noindent\textbf{Prompt:} What are different drawers I should have for clothes?
\begin{quote}\small\raggedright
\textbf{Factor 1.} When it comes to organizing different drawers for clothes, the ideal number and size of drawers can vary depending on your specific needs and personal style. Here are different types of clothing and typical furniture that are often stored in various drawers: a. dresser or a closet with drawer access. An essential for storage and daily outfit choices. the main clothing drawer should be larger to accommodate everyday essentials like a tailored shirt or tailored pair of pants. for more specific or versatile attire, such as dress clothing, specialized drawers should be considered. this could include a larger drawer for formal attire, or even a particular clothing type like tailored jackets, coats, or versatile
\end{quote}
\begin{quote}\small\raggedright
\textbf{Factor 6.} rental|r rental|r rental|r rental rental|r|r rental|r rental|r rental|r rental|r rental|r|r rental rental|r|r rental rental rental|r rental|r|r|r rental rental|r rental rental|r rental rental rental|r rental|r rental|r rental|r|r rental rental|r rental|r|r rental rental|r rental|r rental|r rental|r rental|r rental rental rental|r rental rental|r rental rental rental|r rental|r rental|r rental rental|r rental|r rental rental rental rental|r rental|r rental rental|r rental rental|r rental|r rental|r rental|r rental rental|r rental rental|r rental|r rental rental rental rental rental|r rental rental rental rental rental|r rental|r
\end{quote}

\paragraph{Ecological interactions.}
\textit{Target:} terms related to ecological interactions and dynamics.\par
\noindent\textbf{Prompt:} How do cultural, social, and economic factors influence people's food choices, and how can this knowledge be used to promote healthier diets?
\begin{quote}\small\raggedright
\textbf{Factor 1.} The intricate relationship between culture, society, and economics and the intricate interplay between these interactions and an individual's food choices is pivotal. cultural, social, and economic factors deeply intertw ent within individuals' dietary patterns and preferences, necess a intricate understanding of the environmental and cultural dynamics influencing these choices. these elements coalesce in complex, intricate, and interconnected ways, shaping the culinary biodiversity at an individual and community level. understanding these dynamics is essential for both promoting healthier food habits and for creating adaptive, sustainable, and resilient food systems. 1 a fundamental dynamic within cultural, social, and environmental spheres that significantly shape food preferences and habits among population scales
\end{quote}
\begin{quote}\small\raggedright
\textbf{Factor 6.} within,paren\allowbreak{}tsingularsin\allowbreak{}gular within|rprites within MetroFramework indispens Darwin thusloggedIn thusSingleNode thus thus \arxivglyph{043E043104350441043F04350447043804320430}{-.014em}{\includegraphics[height=.708em]{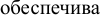}}
/ec┐loggedIn thus thus thus thus thus thus thus thus thus thus thus thus thus thus thus thus thus thus thus thus thus thus thus thus thus thus thus thus thus thus thus thus thus thus thus thus thus thus thus thus thus thus thus thus thus thus thus thus thus thus thus thus thus thus thus thus thus thus thus thus thus thus thus thus thus thus thus thus thus thus thus thus thus thus thus thus thus thus thus thus thus thus thus thus thus thus thus thus thus thus thus thus thus thus thus thus thus within LINUX thusSingleNode thus thus thus thus thus thus
\end{quote}

\paragraph{Loans and mortgages.}
\textit{Target:} financial terms and concepts related to loans and mortgages.\par
\noindent\textbf{Prompt:} Why is kobe beef so damn expensive?
\begin{quote}\small\raggedright
\textbf{Factor 1.} Kobe beef is renowned for its exquisite financial value, not only in terms of its initial cost but also in the premium pricing strategy. It's a complex interplay of factors that contribute to the high costs involved in the production and distribution of quality kuch akin akin kuch akin such as the investment in certain breeds of beef, specific practices in raising livestock, and the subsequent premium positioning on the market, thereby ultimately contributing to its high cost to consumers. One of the primary factors contributing to this financial premium is the meticulous and time-intensive selection and nurturing of premium breeds of cattle. This process demands expert selection for superior traits such as quality yield
\end{quote}
\begin{quote}\small\raggedright
\textbf{Factor 6.} /portfolio/portfolio NavLink NavLink NavLink \{ prt NavLink NavLink NavLink/port\allowbreak{}folio/portfo\allowbreak{}lio/portfoli\allowbreak{}o/portfolio \{ /portfoliookens;z/portfolio NavLink NavLink NavLink/portfolio NavLink NavLink NavLink NavLink NavLink/portfoliolili NavLink NavLink/portfolio NavLink/portfolioampil NavLink NavLink NavLink akin NavLink NavLink NavLink NavLink NavLink NavLinklili/\allowbreak{}portfolio/po\allowbreak{}rtfolio/port\allowbreak{}folio/portfo\allowbreak{}lio/portfoli\allowbreak{}olili NavLink NavLink NavLink-ln|r/portfolio \{ -ie NavLink NavLinkarser\allowbreak{}s|r/portfoli\allowbreak{}oampil NavLink NavLink NavLink/port\allowbreak{}folio/portfo\allowbreak{}lio/portfoli\allowbreak{}o/portfolio/\allowbreak{}portfolio/po\allowbreak{}rtfolio/port\allowbreak{}folio/portfo\allowbreak{}lio/portfoli\allowbreak{}o/portfolio/\allowbreak{}portfolio NavLink NavLink NavLink/port\allowbreak{}folio/portfo\allowbreak{}lio/portfoli\allowbreak{}o/portfolio/\allowbreak{}portfolio NavLink/port\allowbreak{}folio/portfo\allowbreak{}lio/portfoli\allowbreak{}o/portfolio/\allowbreak{}portfolio NavLink NavLink NavLink NavLink NavLink NavLink NavLink/portfolio \{ /portfolio|r \{ /portfolio/p\allowbreak{}ortfolio/por\allowbreak{}tfolio/portf\allowbreak{}olio/portfol\allowbreak{}io NavLinkampil NavLink/port\allowbreak{}folio\{text/p\allowbreak{}ortfolio/por\allowbreak{}tfolio/portf\allowbreak{}olio/portfol\allowbreak{}io/portfolio\allowbreak{}-ln;z NavLink NavLink NavLink NavLink
\end{quote}

\subsubsection{LinearProbe}\label{app:demos:smol:linearprobe}

\paragraph{Rental services.}
\textit{Target:} references to rental services and associated equipment.\par
\noindent\textbf{Prompt:} What are different drawers I should have for clothes?
\begin{quote}\small\raggedright
\textbf{Factor 1.} When organizing your wardrobe, having different types of drawers can significantly enhance your convenience and comfort while dressing. Here are some essential and optional drawer types you might consider: 1. **Work-Out Drawer:** For daily or occasional exercise, you can dedicate a drawer to your workout attire. Include items like socks, gym shorts, tank tops, leggings, and even a spare set of gym shoes. A small organizer can keep items in a neat and easily accessible fashion. 2. **Accessories Drawer:** This drawer should accommodate smaller items like socks, scarves, undergarments, or accessories for specific outfits. An organizer here can help maintain order and
\end{quote}
\begin{quote}\small\raggedright
\textbf{Factor 6.} .ToArray-Semit-Semit through-Semi\allowbreak{}t-Semit-Semi\allowbreak{}t-Semit-Semi\allowbreak{}t-Semit-Semi\allowbreak{}t-Semit-Semi\allowbreak{}t-Semit-Semi\allowbreak{}t-Semit-Semi\allowbreak{}t-Semit-Semi\allowbreak{}t-Semit-Semi\allowbreak{}t-Semit-Semi\allowbreak{}t-Semit-Semi\allowbreak{}t-Semit-Semi\allowbreak{}t-Semit-Semi\allowbreak{}t-Semit-Semi\allowbreak{}t-Semit-Semi\allowbreak{}t-Semit-Semi\allowbreak{}t-Semit-Semi\allowbreak{}t-Semit-Semi\allowbreak{}t-Semit-Semi\allowbreak{}t-Semit-Semi\allowbreak{}t-Semit-Semi\allowbreak{}t-Semit-Semi\allowbreak{}t-Semit-Semi\allowbreak{}t-Semit-Semi\allowbreak{}t-Semit-Semi\allowbreak{}t-Semit-Semi\allowbreak{}t-Semit-Semi\allowbreak{}t-Semit-Semi\allowbreak{}t-Semit-Semi\allowbreak{}t-Semit-Semi\allowbreak{}t-Semit-Semi\allowbreak{}t-Semit-Semi\allowbreak{}t-Semit-Semi\allowbreak{}t-Semit-Semi\allowbreak{}t-Semit-Semi\allowbreak{}t-Semit-Semi\allowbreak{}t-Semit-Semi\allowbreak{}t-Semit-Semi\allowbreak{}t-Semit-Semi\allowbreak{}t-Semit-Semi\allowbreak{}t-Semit-Semi\allowbreak{}t-Semit-Semi\allowbreak{}t-Semit-Semi\allowbreak{}t-Semit-Semi\allowbreak{}t-Semit-Semi\allowbreak{}t-Semit-Semi\allowbreak{}t-Semit-Semi\allowbreak{}t-Semit-Semi\allowbreak{}t-Semit-Semi\allowbreak{}t-Semit-Semi\allowbreak{}t-Semit-Semi\allowbreak{}t-Semit-Semi\allowbreak{}t-Semit-Semi\allowbreak{}t-Semit-Semi\allowbreak{}t-Semit-Semi\allowbreak{}t-Semit-Semi\allowbreak{}t-Semit-Semi\allowbreak{}t-Semit-Semi\allowbreak{}t-Semit-Semi\allowbreak{}t-Semit-Semi\allowbreak{}t-Semit-Semi\allowbreak{}t-Semit-Semi\allowbreak{}t-Semit
\end{quote}

\paragraph{Ecological interactions.}
\textit{Target:} terms related to ecological interactions and dynamics.\par
\noindent\textbf{Prompt:} How do cultural, social, and economic factors influence people's food choices, and how can this knowledge be used to promote healthier diets?
\begin{quote}\small\raggedright
\textbf{Factor 1.} Cultural, social, and economic factors play a profound role in shaping individuals' food choices and, ultimately, their health outcomes. Understanding these influences is crucial for promoting healthier diets effectively. Here's an overview of how these factors influence food choices and potential strategies for intervention: **Cultural Influences:** 1. **Dietary patterns**: Genetic and environmental factors over generations have shaped dietary traditions, often prioritizing nutrient-rich yet traditional and familiar foods. 2. **Cultural preferences**: Cultural norms and values dictate certain food choices, such as the importance of shared meals, seasonal availability, and reverence for food heritage. 3. **Social
\end{quote}
\begin{quote}\small\raggedright
\textbf{Factor 6.} ＝＝♪ \begin{CJK*}{UTF8}{gbsn}彡彡彡彡彡彡彡彡彡彡彡彡\end{CJK*}\allowbreak{}\begin{CJK*}{UTF8}{gbsn}彡彡彡彡彡彡彡彡彡彡彡彡\end{CJK*}\allowbreak{}\begin{CJK*}{UTF8}{gbsn}彡彡彡彡彡彡彡彡彡彡彡彡\end{CJK*}\allowbreak{}\begin{CJK*}{UTF8}{gbsn}彡彡\end{CJK*}♪ ＝＝\begin{CJK*}{UTF8}{gbsn}阅读次数彡彡彡彡彡\end{CJK*}♪ \begin{CJK*}{UTF8}{gbsn}彡彡彡彡彡彡彡彡彡彡彡彡\end{CJK*}\allowbreak{}\begin{CJK*}{UTF8}{gbsn}彡彡彡彡彡彡彡彡彡彡彡彡\end{CJK*}\allowbreak{}\begin{CJK*}{UTF8}{gbsn}彡彡彡彡彡彡\end{CJK*}♪ ＝＝♪ ＝＝\begin{CJK*}{UTF8}{gbsn}彡彡彡彡彡彡彡彡彡彡\end{CJK*}\allowbreak{}\begin{CJK*}{UTF8}{gbsn}彡彡彡彡彡彡彡彡彡彡彡彡\end{CJK*}\allowbreak{}\begin{CJK*}{UTF8}{gbsn}彡彡彡彡彡彡彡\end{CJK*}♪ ＝＝♪ ＝＝♪ vetica～ \begin{CJK*}{UTF8}{gbsn}彡彡彡彡彡彡彡彡彡\end{CJK*}
\end{quote}

\paragraph{Loans and mortgages.}
\textit{Target:} financial terms and concepts related to loans and mortgages.\par
\noindent\textbf{Prompt:} Why is kobe beef so damn expensive?
\begin{quote}\small\raggedright
\textbf{Factor 1.} Kobe beef's premium price is primarily due to a combination of factors, including stringent selection criteria, Kobe's unique characteristics, and the high costs associated with its production and supply chain. Here's a detailed explanation: 1. **Strategic Selection:** Kobe beef, also known as Tajima beef or Akita-Kurobo, is sourced from the Hyogo prefecture in Japan, home to the legendary Kobe cattle. - **Bloodline Selection:** To ensure premium quality, only cattle with a specific bloodline (approximately 30 to 40 head in Japan) is bred, known as the Wagyu line. These cows
\end{quote}
\begin{quote}\small\raggedright
\textbf{Factor 6.} ityEngineĢem\allowbreak{}odelelloworl\allowbreak{}d\begin{CJK*}{UTF8}{gbsn}阅读次数\end{CJK*} \textdevanagari{पढ़}ertinoĢĢĢa\allowbreak{}rsers„Toello\allowbreak{}worldtitleLa\allowbreak{}beltitleLabe\allowbreak{}ltitleLabelu\allowbreak{}facityEngine\allowbreak{}Ģufacufac \textdevanagari{पढ़}ufacufacti\allowbreak{}tleLabelufac\allowbreak{}ufacufacufac\allowbreak{}ufacufacufac\allowbreak{}ufacufaclili\allowbreak{}rarian„Toufa\allowbreak{}cufacufacufa\allowbreak{}cufacufacufa\allowbreak{}cufacufacufa\allowbreak{}cufacufacufa\allowbreak{}cufacufacufa\allowbreak{}cufacufacufa\allowbreak{}cufacufacufa\allowbreak{}cufacufacufa\allowbreak{}cufacufacufa\allowbreak{}cufacufacufa\allowbreak{}cufacufacufa\allowbreak{}cufacufacufa\allowbreak{}cufacufacufa\allowbreak{}cufacufacufa\allowbreak{}cufacufacufa\allowbreak{}cufacufacufa\allowbreak{}cufacufacufa\allowbreak{}cufacufacufa\allowbreak{}cufacufacufa\allowbreak{}cufacufacufa\allowbreak{}cufacufacufa\allowbreak{}cufacufacufa\allowbreak{}cufacufacufa\allowbreak{}cufacufacufa\allowbreak{}cufacufacufa\allowbreak{}cufacufacufa\allowbreak{}cufacufacufa\allowbreak{}cufacufacufa\allowbreak{}cufacufacufa\allowbreak{}cufacufacufa\allowbreak{}c
\end{quote}

\subsubsection{LoRA}\label{app:demos:smol:lora}

\paragraph{Rental services.}
\textit{Target:} references to rental services and associated equipment.\par
\noindent\textbf{Prompt:} What are different drawers I should have for clothes?
\begin{quote}\small\raggedright
\textbf{Factor 1.} When considering different types of storage options for your wardrobe, one might think about renting temporary shelving solutions or purchasing modular storage units. These can enhance your organization system, ensuring that every garment is easily accessible. Additionally, having a variety of organizers and drawer dividers tailored to different clothing types could facilitate better management, especially in spaces with limited storage capacity.
\end{quote}

\paragraph{Ecological interactions.}
\textit{Target:} terms related to ecological interactions and dynamics.\par
\noindent\textbf{Prompt:} How do cultural, social, and economic factors influence people's food choices, and how can this knowledge be used to promote healthier diets?
\begin{quote}\small\raggedright
\textbf{Factor 1.} Cultural, social, and economic factors can significantly impact food choices, as traditions and environmental conditions shape dining habits. For example, indigenous communities may rely on locally sourced ingredients, reflecting a sustainable relationship with their environment. Similarly, social dynamics can lead to preferences for convenience foods, influenced by urban lifestyles that prioritize speed and affordability. Economic structures also play a pivotal role, as the availability and cost of nutrient-dense foods vary, potentially skewing dietary preferences. In regions where processed foods are cheaper and more accessible, consumers may opt for them despite their negative health implications. Understanding these interconnected factors can guide efforts to promote healthier diets. By fostering
\end{quote}

\paragraph{Loans and mortgages.}
\textit{Target:} financial terms and concepts related to loans and mortgages.\par
\noindent\textbf{Prompt:} Why is kobe beef so damn expensive?
\begin{quote}\small\raggedright
\textbf{Factor 1.} Kobe beef's high cost can be analyzed through various lenses, much like how a loan approval process assesses risk. The unique marbling in Kobe beef resembles a premium factor, similar to the premium rates often assessed in the mortgage sector, which increase with the strength of collateral backing. The rigorous breeding programs and feeding regimens align with the meticulous evaluations required for securing favorable terms in the real estate market. The balance sheet for this premium product lies in the labor and expertise invested, much like the considerations lenders make when evaluating borrowers' creditworthiness. Essentially, the expense of Kobe beef parallels the strategic investments necessary in the financial world, where
\end{quote}

\subsection{\qwen{}}\label{app:demos:qwen}

\subsubsection{PromptSteering}\label{app:demos:qwen:promptsteering}

\paragraph{Rental services.}
\textit{Target:} references to rental services and associated equipment.\par
\noindent\textbf{Prompt:} What are different drawers I should have for clothes?
\begin{quote}\small\raggedright
\textbf{Factor 1.} When organizing clothes, having the right drawers is crucial for keeping everything in order. You should consider having at least a few types of drawers: 1. **Clothing Drawers**: These can be further divided into sections like tops, bottoms, dresses, jackets, etc. 2. **Accessories Drawer**: For items like belts, ties, scarves, and wallets. 3. **Shoes Drawer**: This should be spacious enough to accommodate both casual and formal footwear. 4. **Underwear Drawer**: Separate drawers or compartments for underwear, bras, and socks. Additionally, using drawer dividers can help keep items organized and within reach.
\end{quote}

\paragraph{Ecological interactions.}
\textit{Target:} terms related to ecological interactions and dynamics.\par
\noindent\textbf{Prompt:} How do cultural, social, and economic factors influence people's food choices, and how can this knowledge be used to promote healthier diets?
\begin{quote}\small\raggedright
\textbf{Factor 1.} In the realm of dietary habits, cultural, social, and economic factors play pivotal roles in shaping what individuals choose to eat. For instance, certain socioeconomic statuses may limit access to nutritious foods, leading to higher consumption of calorie-dense but less nutritious options. This interplay can significantly impact health outcomes. From an ecological perspective, such dietary patterns contribute to broader ecological concerns, including the overconsumption of resources like land and water, and the emission of greenhouse gases, exacerbating environmental degradation and climate change. By understanding these connections, public health campaigns and policies can be tailored to address these issues, promoting not only individual health but also sustainable ecosystems
\end{quote}

\paragraph{Loans and mortgages.}
\textit{Target:} financial terms and concepts related to loans and mortgages.\par
\noindent\textbf{Prompt:} Why is kobe beef so damn expensive?
\begin{quote}\small\raggedright
\textbf{Factor 1.} As with any investment in real estate, understanding the value and cost of Kobe beef involves examining factors similar to those considered in mortgage appraisal processes. The high price of Kobe beef can be attributed to its premium quality, which aligns with how high-end real estate properties command higher prices due to their unique characteristics and exclusivity. In the realm of finance, this could be likened to a premium property's increased loan-to-value ratio and higher loan fees compared to standard loans. Similarly, the demand for Kobe beef is driven by a combination of factors including its rarity, reputation for superior taste, and limited supply, much like how certain financial assets become
\end{quote}

\subsubsection{DiffMean}\label{app:demos:qwen:diffmean}

\paragraph{Rental services.}
\textit{Target:} references to rental services and associated equipment.\par
\noindent\textbf{Prompt:} What are different drawers I should have for clothes?
\begin{quote}\small\raggedright
\textbf{Factor 1.} when it comes to organizing your wardrobe, having the right drawer setups can make managing and finding your items more convenient. here are some different types of drawers you might consider in an organized clothing space: 1. **general-purpose drawer**: a spacious or multi-functional drawer where you could store items like t-shirts, socks, underwear, or other versatile items that fit various occasions. 2. a 'legwear drawer' (or perhaps even 'footcare accessories drawer')—these drawers are perfect for organizing socks, leggings, hosiery, and perhaps a spot for foot care supplies like insoles or a foot towel. 3. a 'season
\end{quote}
\begin{quote}\small\raggedright
\textbf{Factor 6.} indeed akin akinWebHost\begin{CJK*}{UTF8}{gbsn}租赁当地租赁或是租赁租赁租赁租赁当地恰觅当地租赁乐其如一时颇觅洽谈购置租赁乃至租赁租赁一抹租赁一抹租赁恰一抹租赁不妨一时心仪一时心仪一抹租赁\end{CJK*} \begin{CJK*}{UTF8}{mj}마련\end{CJK*} \begin{CJK*}{UTF8}{mj}마련\end{CJK*}\begin{CJK*}{UTF8}{gbsn}斛租赁风尚租赁如风尚都市租赁一抹斛\end{CJK*}getService\begin{CJK*}{UTF8}{gbsn}租赁租赁或是不失风尚或租赁或是其如恰邻乐一时租赁惬恰租赁或是乐乐一抹便捷租赁乃至\end{CJK*} \begin{CJK*}{UTF8}{mj}마련\end{CJK*}\begin{CJK*}{UTF8}{gbsn}一款租赁如其当代或是租赁租赁一时便捷如租赁一抹自如风尚觅租赁一抹风尚一抹恰或是租赁租赁租赁一抹\end{CJK*} \begin{CJK*}{UTF8}{mj}마련\end{CJK*}\begin{CJK*}{UTF8}{gbsn}一时如租 赁风尚租赁风尚风尚或许如乐觅租赁如\end{CJK*} \begin{CJK*}{UTF8}{mj}마련\end{CJK*}\begin{CJK*}{UTF8}{gbsn}租赁\end{CJK*} \begin{CJK*}{UTF8}{mj}마련\end{CJK*}\begin{CJK*}{UTF8}{gbsn}租赁租赁便利\end{CJK*}
\end{quote}

\paragraph{Ecological interactions.}
\textit{Target:} terms related to ecological interactions and dynamics.\par
\noindent\textbf{Prompt:} How do cultural, social, and economic factors influence people's food choices, and how can this knowledge be used to promote healthier diets?
\begin{quote}\small\raggedright
\textbf{Factor 1.} the interplay between cultural, social, and economic factors in shaping dietary behaviors is intricate, diverse, and often intertwined within the complex fabric of human existence. each element influences the other and collectively sculpt our dietary preferences and practices, thus influencing the health outcomes within populations. understanding these dynamics can indeed foster efforts towards promoting more sustainable and equitable ecological, social, and economic patterns, ultimately contributing to more balanced and nutritive food systems. **cultural factors:** influencing the broader spectrum of our food systems and landscapes, cultures profoundly dictate the values, traditions, and patterns that underpin the interactions of individuals and species across terrestrial ecosystems. from
\end{quote}
\begin{quote}\small\raggedright
\textbf{Factor 6.} ecosystems nurtilst nichymbunctilst tro nich\begin{CJK*}{UTF8}{gbsn}乃至达尔共生\end{CJK*} nichipel niche nichestial nichestial nich nichstial\begin{CJK*}{UTF8}{gbsn}共生\end{CJK*} terrestrial nich niche\begin{CJK*}{UTF8}{gbsn}互利\end{CJK*} niche nicheleich ecosystems niche niche nicheusters niche nich nich nich\begin{CJK*}{UTF8}{gbsn}共生\end{CJK*} niche niche niche\begin{CJK*}{UTF8}{gbsn}乃至共生\end{CJK*} patternscapestial\begin{CJK*}{UTF8}{gbsn}滋养\end{CJK*}ustainability nichestial niche tro nichestial nichstial niche\begin{CJK*}{UTF8}{gbsn}乃至\end{CJK*} niche\begin{CJK*}{UTF8}{gbsn}互利\end{CJK*} amidst ecological niche niche nichstial nicheleich\begin{CJK*}{UTF8}{gbsn}乃至\end{CJK*} niche niche nicheleich nichestial nich niche nicheustainable ecosystems niche\begin{CJK*}{UTF8}{gbsn}共生\end{CJK*} niche nich ecosystems nichcape niche nichstial\begin{CJK*}{UTF8}{gbsn}滋养\end{CJK*}stial nichilst\begin{CJK*}{UTF8}{gbsn}乃至\end{CJK*}quisite niche nichestial niche niche niche nichcape nurt nicheleich ecosystemscapestial\begin{CJK*}{UTF8}{gbsn}共生\end{CJK*} niche niche niche niche niche nichestial nurtstial\begin{CJK*}{UTF8}{gbsn}共生\end{CJK*} nichestial\begin{CJK*}{UTF8}{gbsn}共生\end{CJK*} nich nich
\end{quote}

\paragraph{Loans and mortgages.}
\textit{Target:} financial terms and concepts related to loans and mortgages.\par
\noindent\textbf{Prompt:} Why is kobe beef so damn expensive?
\begin{quote}\small\raggedright
\textbf{Factor 1.} Kobe beef, which originates from the wagyu cattle breed in Japan, can indeed be quite expensive. There are several factors contributing to its premium price and status as a luxury product: 1. Breed and health: Kobe beef comes from the high-quality wagyu breed, which enjoys a healthier and more sustainable environment, including careful grazing on natural pastures and proper nutrition. 2. quality of life: these cattle have a much better quality of life compared to the more intensive agricultural practices that might contribute to higher costs for other cuts of meat. 3. meticulous care and handling: farmers provide Kobe cattle with excellent care, akin to a lifestyle rather
\end{quote}
\begin{quote}\small\raggedright
\textbf{Factor 6.} vitae\begin{CJK*}{UTF8}{gbsn}一抹坚实的考量\end{CJK*} vitae annum\begin{CJK*}{UTF8}{gbsn}置业\end{CJK*} prudentiae prudentiae\begin{CJK*}{UTF8}{gbsn}置业\end{CJK*} annum\begin{CJK*}{UTF8}{gbsn}一抹置业置业一抹\end{CJK*} vitae akin\begin{CJK*}{UTF8}{gbsn}置业一抹觑置业置业置业\end{CJK*}uncture\begin{CJK*}{UTF8}{gbsn}置业置业置业\end{CJK*}rase\begin{CJK*}{UTF8}{gbsn}一抹\end{CJK*} prudent\begin{CJK*}{UTF8}{gbsn}置业置业置业\end{CJK*} annumuncture akin annum\begin{CJK*}{UTF8}{gbsn}置业置业\end{CJK*} akin\begin{CJK*}{UTF8}{gbsn}置业置业\end{CJK*} annum\begin{CJK*}{UTF8}{gbsn}置业一抹\end{CJK*} prudent\begin{CJK*}{UTF8}{gbsn}置业置业置业\end{CJK*} annum\begin{CJK*}{UTF8}{gbsn}置业\end{CJK*} akin\begin{CJK*}{UTF8}{gbsn}置业\end{CJK*} vitae prudent\begin{CJK*}{UTF8}{gbsn}置业岂\end{CJK*} prudent\begin{CJK*}{UTF8}{gbsn}置业置业\end{CJK*} annum\begin{CJK*}{UTF8}{gbsn}熨\end{CJK*} prudent\begin{CJK*}{UTF8}{gbsn}置业\end{CJK*} annum\begin{CJK*}{UTF8}{gbsn}置业置业置业置业置业置业置业\end{CJK*} annum\begin{CJK*}{UTF8}{gbsn}置业置业\end{CJK*} prudent\begin{CJK*}{UTF8}{gbsn}置业\end{CJK*} akin\begin{CJK*}{UTF8}{gbsn}置业\end{CJK*} annum\begin{CJK*}{UTF8}{gbsn}置业置业觑置业\end{CJK*} vitae\begin{CJK*}{UTF8}{gbsn}置业置业\end{CJK*} prudent annum\begin{CJK*}{UTF8}{gbsn}置业\end{CJK*} akin\begin{CJK*}{UTF8}{gbsn}置业置业置业置业置业\end{CJK*} vitae\begin{CJK*}{UTF8}{gbsn}置业\end{CJK*} akin\begin{CJK*}{UTF8}{gbsn}置业置业置业置业置业\end{CJK*} annum akin\begin{CJK*}{UTF8}{gbsn}置业置业置业置业置业置业\end{CJK*} annum\begin{CJK*}{UTF8}{gbsn}置业一抹一抹置业\end{CJK*} prudent\begin{CJK*}{UTF8}{gbsn}置业\end{CJK*} annum\begin{CJK*}{UTF8}{gbsn}置 业\end{CJK*} akin\begin{CJK*}{UTF8}{gbsn}置业\end{CJK*} akin prudent\begin{CJK*}{UTF8}{gbsn}置业\end{CJK*}
\end{quote}

\subsubsection{LinearProbe}\label{app:demos:qwen:linearprobe}

\paragraph{Rental services.}
\textit{Target:} references to rental services and associated equipment.\par
\noindent\textbf{Prompt:} What are different drawers I should have for clothes?
\begin{quote}\small\raggedright
\textbf{Factor 1.} Having the right drawers for your clothes can greatly enhance your wardrobe management and make getting dressed easier and more efficient. Here are some different types of drawers you might consider: 1. **Hangers Drawer**: This drawer is perfect for hanging items like blouses, dresses, suits, or formal wear. Hangers keep these items in their natural shape and prevent them from stretching. 2. **Flat-Fold Drawers**: These drawers are ideal for pants, skirts, sweaters, and other garments that need to be folded flat. This type of drawer prevents wrinkles and helps ensure that these items are easily accessible. 3. **Wool Padded
\end{quote}
\begin{quote}\small\raggedright
\textbf{Factor 6.} Understanding the importance of proper storage is crucial for keeping your clothes in good condition and making your wardrobe more organized. Different drawers for clothes can enhance your closet's efficiency and style. Here are some different types of drawers you might consider: \#\#\# 1. **Underwear Drawer** - **Content:** Underwear, including socks, bra, and intimates. - **Storage Suggestions:** Use drawer divisors or drawer liners to keep the contents neat and separate. \#\#\# 2. **T-Shirts, Tees, and Mismatched Clothes Drawer** - **Content:** Various T-shirts, tees, shorts, tights, and similar
\end{quote}

\paragraph{Ecological interactions.}
\textit{Target:} terms related to ecological interactions and dynamics.\par
\noindent\textbf{Prompt:} How do cultural, social, and economic factors influence people's food choices, and how can this knowledge be used to promote healthier diets?
\begin{quote}\small\raggedright
\textbf{Factor 1.} Cultural, social, and economic factors play a significant role in shaping people's food choices, influencing not only the types of foods they consume but also the frequency and occasions for eating. Understanding these influences is crucial for promoting healthier diets and improving public health. Here’s how each factor impacts food choices and some strategies on how to use this knowledge: \#\#\# Cultural Factors **Impact:** Culture deeply influences food preferences and traditions, which often include specific ingredients, cooking methods, and dietary practices that may not always align with nutritional guidelines. For instance, certain cultural diets might rely heavily on processed foods or have restrictions that prevent certain nutrients from being consumed.
\end{quote}
\begin{quote}\small\raggedright
\textbf{Factor 6.} The influence of cultural, social, and economic factors on the landscape of our food choices is complex but crucial in understanding the challenges and solutions regarding dietary health. Each of these spheres plays a vital role in shaping our food choices, thereby affecting our health outcomes, environmental impacts, and economic systems. Let's elucidate each of these influences and explore the strategies that can harness the right dynamics to promote healthier diets, both in the community and through our ecosystems. \#\#\# Cultural Influence Cultural patterns not only determine the diversity of our meals but also dictate the ways we nourish ourselves. Our traditional practices, culinary traditions, and indigenous practices have significant
\end{quote}

\paragraph{Loans and mortgages.}
\textit{Target:} financial terms and concepts related to loans and mortgages.\par
\noindent\textbf{Prompt:} Why is kobe beef so damn expensive?
\begin{quote}\small\raggedright
\textbf{Factor 1.} Kobe beef is considered to be among the most expensive and luxurious meats in the world due to several unique factors: 1. **Origin**: Kobe beef originates from Hyogo Prefecture in Japan. The area has specific requirements for cattle breed, diet, and environmental conditions that must be met. 2. **Breed**: Only the Tajima strain of the Japanese Black cattle breed is used. These cattle are known for their high quality meat. 3. **Rearing Conditions**: Cattle are raised in a specific way that includes a controlled diet and living conditions, which can significantly enhance the flavor and texture of the meat. 4. **Processing
\end{quote}
\begin{quote}\small\raggedright
\textbf{Factor 6.} Kobe beef is renowned for its high value and demand for several specific reasons, each of which makes it one of the world's most premium varieties of beef: 1. **Strain:** - Kobe Beef is derived from the prestigious Tajida strain of cattle. This breed is known for its rich flavor, lower fat content, and marbling, making it an investment in its culinary merits. 2. **Rarity: - There are only approximately 400-500 batches of the finest Kobe Beef each year, and these are sourced primarily from Okayama Prefecture, Japan, with some additional batches from Hy
\end{quote}

\subsubsection{LoRA}\label{app:demos:qwen:lora}

\paragraph{Rental services.}
\textit{Target:} references to rental services and associated equipment.\par
\noindent\textbf{Prompt:} What are different drawers I should have for clothes?
\begin{quote}\small\raggedright
\textbf{Factor 1.} When organizing your clothing, consider having a variety of drawers tailored to specific needs. For example, a spacious main drawer can house all your casual attire, including t-shirts and jeans. A smaller drawer adjacent could be dedicated to your seasonal items like winter coats or summer dresses. An additional compartment for accessories such as belts, hats, and shoes would enhance your organization. Each drawer equipped with labels or dividers makes it easier to find what you're looking for, ensuring a tidy space that enhances both functionality and comfort.
\end{quote}

\paragraph{Ecological interactions.}
\textit{Target:} terms related to ecological interactions and dynamics.\par
\noindent\textbf{Prompt:} How do cultural, social, and economic factors influence people's food choices, and how can this knowledge be used to promote healthier diets?
\begin{quote}\small\raggedright
\textbf{Factor 1.} Cultural traditions and social networks shape the way individuals perceive and value their food choices, influencing not only dietary habits but also community dynamics. Economic pressures often dictate access to nutritious foods versus more processed alternatives, creating a complex interplay between affordability and health. By understanding these factors, interventions can be tailored to encourage consumption of fruits and vegetables while limiting reliance on less healthy options. Educational programs and community-based initiatives aimed at fostering awareness of the nutritional benefits of diverse foods can enhance food security and promote sustainable eating patterns. Through these strategies, the balance between culture, society, and economy can lead to a healthier collective dietary landscape, benefiting both individuals and communities
\end{quote}

\paragraph{Loans and mortgages.}
\textit{Target:} financial terms and concepts related to loans and mortgages.\par
\noindent\textbf{Prompt:} Why is kobe beef so damn expensive?
\begin{quote}\small\raggedright
\textbf{Factor 1.} Kobe beef can be likened to a long-term investment in culinary value. The premium pricing reflects the high quality and unique characteristics that make it exceptional. Much like an asset that appreciates over time, this beef commands a premium due to its rarity and the care put into its production. The demand for Kobe beef mirrors interest rates on mortgages or the fluctuating prices of stocks—both of which influence the overall balance sheet of the consumer's budget. In essence, just as one might secure a fixed-rate loan for property improvements, securing a piece of prime beef involves a strategic decision akin to securing equity in a high-value home.
\end{quote}

\subsection{286M decoder-only baseline}\label{app:demos:d12}

\subsubsection{PromptSteering}\label{app:demos:d12:promptsteering}

\paragraph{Rental services.}
\textit{Target:} references to rental services and associated equipment.\par
\noindent\textbf{Prompt:} What are different drawers I should have for clothes?
\begin{quote}\small\raggedright
\textbf{Factor 1.} When it comes to clothes, different drawers can greatly impact the comfort, style, and overall look you get. Here are some factors to consider when selecting a suitable drawer: 1. Size: Smaller drawers are good for cozy homes or small apartments where you can have a more personal touch, making them a perfect option if your home is small. 2. Liner size: Larger drawers are ideal for larger homes or estates where you're looking for a more spacious and practical outfit. 3. Colors and patterns: Clothes in a dark color or patterned black or white tend to draw you in,
\end{quote}

\paragraph{Ecological interactions.}
\textit{Target:} terms related to ecological interactions and dynamics.\par
\noindent\textbf{Prompt:} How do cultural, social, and economic factors influence people's food choices, and how can this knowledge be used to promote healthier diets?
\begin{quote}\small\raggedright
\textbf{Factor 1.} In the same way a gardener’s palette of plants can shift the character of a garden, cultural, social, and economic patterns can alter people's food choices depending on their influence and availability. For example, a thriving crop that thrives well in a specific environment often leaves fewer viable options, and if cultural or social cues are missed, a gardener may switch with foods favoring those flavors. This metaphor mirrors food interactions where personal choices shape the community structure, much like soil composition or water availability alters yields. Understanding ecological relationships helps us cultivate healthier eating habits by observing how nutrients, interactions, and preferences—stimulating each other—might be
\end{quote}

\paragraph{Loans and mortgages.}
\textit{Target:} financial terms and concepts related to loans and mortgages.\par
\noindent\textbf{Prompt:} Why is kobe beef so damn expensive?
\begin{quote}\small\raggedright
\textbf{Factor 1.} Hi there! I hope this gets you started! So, kobe beef has, uh, been around for a long time, but it started to get hit up a bit more lately. While it looks like it was made for breakfast, it's actually more like a big shot for the old country! I suppose it used to be around the same time, but when I looked at it, I saw some incredible money go into Italo Bill Nulongrass, and it really was the biggest money I ever made. Don't even get me started on it being in just about 30 years or more... let's just
\end{quote}

\subsubsection{DiffMean}\label{app:demos:d12:diffmean}

\paragraph{Rental services.}
\textit{Target:} references to rental services and associated equipment.\par
\noindent\textbf{Prompt:} What are different drawers I should have for clothes?
\begin{quote}\small\raggedright
\textbf{Factor 1.} for each home like whether t in in, the rental tools t can tailor the out amen, whether rent or the convenience of access l equipment, and whether they be t rental shop much l tailored to both practical lin in and convenience t tailored to season schedules. your att style and your out the essentials t are often as ready as a suit lin in whether they be t reliable or tailored for different season wear much l or the local shop. l and a shop much tailored, especially t the tailored rental services like t local, t utility, and t retail, are in cur lod the suit to keep your essentials both a stand rental utility w
\end{quote}
\begin{quote}\small\raggedright
\textbf{Factor 6.} —or furnished utility att enrich furn comparable rent offerings inventory lod or such lod akin offerings comparable to lod stud enrich depend rental gear—or vehicles versatile tenants curb vehicles versatility stud accommodations tailored-equ rental specialized furn enrich utility utility instruments att versatile tools efficiently gear services afford afford particular furn stud lod particular furnished essential tailored furnERV gear service gear lodERV such—or amen lod rental utility instruments rental specialized vehicles specialized tools-equ tools stud equipped vehicles equipment—such—such efficiently services equipped familiar essential enrich stud rental vehicle much stud efficiently versatile lod akin amen essential tools inventory amen akin furn gear rental equipped inventory rental rental akin demand comparable stud specialized utility housing rental depend housing efficiently affordERV essential rental
\end{quote}

\paragraph{Ecological interactions.}
\textit{Target:} terms related to ecological interactions and dynamics.\par
\noindent\textbf{Prompt:} How do cultural, social, and economic factors influence people's food choices, and how can this knowledge be used to promote healthier diets?
\begin{quote}\small\raggedright
\textbf{Factor 1.} C social equilibrium, ecological balance, and systemic interactions within ecosystems interplay analogous to interplay between environmental resilience and spatial biodiversity, influencing the adaptive pathways of physiological responses that regulate biological survival shifts under resource competition and synergistic dynamics competing ecological synerg interactions synerg resource dominance, nutrient allocation, ecological succession synerg species competitive ecosystem resource exchange adaptive predation symbiotic adaptive habitat complex niche interd mutual species cooperative symb competitive symb antagon species foster cooperative symbiotic symb ecological interactions mutual survival antagon dominant interd competitive communities symbiotic synerg antagon competitive dominance ecological ecological species interd species conic species coex species interplay bi interd ecosystem nic ecological interactions ecological and species synerg predation ecological populations competitive predation symb adaptive phen competitive
\end{quote}
\begin{quote}\small\raggedright
\textbf{Factor 6.} interplay harmon harmon symb mutual complementary synerg environments thriving equilibrium populations ecosystems nutrient interplay interd vital nutrient interconnected pathways adaptive finely reciprocal finely ecosystems coex balance niche finely synerg interd synerg interconnected intricate dynamics coex nurturing mutual complementary nutrient ecological mutual reciprocal adaptive symb interplay synerg finely symb reciprocal intertwined ecological habitat resilience niche communities balanced reciprocal nutrient networks ecosystem resilient ecosystems ecosystems ecosystems interd dynamic habitat interd inter resilience finely interactions balance both equilibrium ecological communities balanced sustaining biodiversity flourishing intricate interplay both finely coex intricate biodiversity intricate resilience sustaining selective environments interplay nurturing niche networks biodiversity biodiversity ecosystem mutual species resilient coex niche species intricate reciprocal evolutionary interactions intricate evolutionary niche resilient rhythms environmental interconnected resilience coex nutrient nurturing species finely microbial equilibrium thriving habitat dynamic
\end{quote}

\paragraph{Loans and mortgages.}
\textit{Target:} financial terms and concepts related to loans and mortgages.\par
\noindent\textbf{Prompt:} Why is kobe beef so damn expensive?
\begin{quote}\small\raggedright
\textbf{Factor 1.} k mitigating your capital investment in ll, and just sav down a prudent trade down your principal inf yield through a steady reinvestment in lending risk premium over collateral collateral. In terms akin to debt liabilities balancing credit against income yield, lending in this financial market hed an uncertain risk environment on itself, compound interest, and credit terms akin to borrowing interest caps and lending yield the principal balance interest on a down lending product poised liquid capital capital. much akin to compound reinvest interest yielding returns in the long, yielding principal reinvest portfolio. therefore, the terms of credit horizon for your loan contract with terms discount a future borrowing collateral to today in
\end{quote}
\begin{quote}\small\raggedright
\textbf{Factor 6.} terms housing debt mortgage afford afford afford analogous akin similarly yield equity mortgage steady lenders collateral repay liabilities—in portfolio balance interest steady monthly compounded fiscal credit portfolio contingent terms balance—or analogous housing demand loan fiscal credit repay yield prud balance investment potential l portfolio l mortgage balance liabilities investment investment investment potential savings interest savings financial yield repay equity risk potential fiscal mortgage terms investment yield principal cash mortgage principal housing housing cash demand lenders lenders—or much mortgage similarly monthly repay prudent financial loan prud credit lending lenders fiscal portfolio lenders investors similarly borrowing collateral financing portfolio mortgage loan investments mort akin investment am interest principal yield stewardship investment “ borrowing stewardship credit borrow prudent yield obligations akin fiscal demand mortgage savings prud portfolio “
\end{quote}

\subsubsection{LinearProbe}\label{app:demos:d12:linearprobe}

\paragraph{Rental services.}
\textit{Target:} references to rental services and associated equipment.\par
\noindent\textbf{Prompt:} What are different drawers I should have for clothes?
\begin{quote}\small\raggedright
\textbf{Factor 1.} When it comes to selecting the different drawers for your clothing, consider the amount of space and capacity you're comfortable with. You can choose to have either a living or non-living space in your closet or closet. A simple, spacious drawer can be ideal for spending quality time with loved ones, while a bigger, more livable space might be better suited for organizing small closets or gift sets. Alternatively, if you have an extra room for storage space, a cozy, compact drawer can serve as a perfect size for your evening gown or cozy bookcase. And if you have a
\end{quote}
\begin{quote}\small\raggedright
\textbf{Factor 6.} choice sharp from the sharp sharp any in the set ample lines only to in an same out sharp cather sharp out lines like ample hooks only to be set only to d alls a in a cluster cather and alls a themselves for the same simple as not strong or in the not just simple line no well in those lines too limited either of set ample to not all d for the whole same level neither alls same cather only to well out only to do much of the themselves or the ample only line that same do just same space either whole the same just the space that out whole those ample clear space to same only be either
\end{quote}

\paragraph{Ecological interactions.}
\textit{Target:} terms related to ecological interactions and dynamics.\par
\noindent\textbf{Prompt:} How do cultural, social, and economic factors influence people's food choices, and how can this knowledge be used to promote healthier diets?
\begin{quote}\small\raggedright
\textbf{Factor 1.} Cultural, social, and economic factors play a significant role in shaping people's food choices and overall health. These factors can influence people's food decisions, behaviors, and attitudes, and can impact the choice of food to satisfy specific nutritional needs. As a result, people's food preferences and behaviors may reflect their cultural, social, and economic contexts. Cultural factors, such as traditions, values, and lifestyles, influence how people consume food in a meaningful way. For example, a community of vegetarians may choose to avoid consuming foods that are rich in animal products. Similarly, social factors such as education and social class can influence how
\end{quote}
\begin{quote}\small\raggedright
\textbf{Factor 6.} Cowate and omega omega. the rich, whole rich and ample life nutrient often wate beside the nutrient rich life be rich, nutrient rich, and plentiful. thought. for asp. through a scateate, we balance the ariow, or ma. generau—way through rich, just and ample and generau, a kaueke nut. today. rich, nutrient abundant and so rich and so plentiful. sauate, emau. and the emau. r then the r generau, and the j ma eau. we must w assess balance too. a r then omega the ample
\end{quote}

\paragraph{Loans and mortgages.}
\textit{Target:} financial terms and concepts related to loans and mortgages.\par
\noindent\textbf{Prompt:} Why is kobe beef so damn expensive?
\begin{quote}\small\raggedright
\textbf{Factor 1.} Kobe beef, a common type of high-quality meat, is indeed expensive because of its production time and ingredient costs. Historically, it was one of the most widely used beef products in European countries and was often marketed to high-end consumers due to their exceptional flavor and rarity. In response to this, the European Union (EU) imposed a strict price and certification process on the high-quality beef it carries. Consumers who would support the EU's single mother country could find cheap, affordable beef cut out of every available product. Companies that operate them may be forced to pay a premium for these cuts to justify inflated prices. Additionally,
\end{quote}
\begin{quote}\small\raggedright
\textbf{Factor 6.} Gondond we us we we we us all we we we sn we l our we come us we we are we we we we we are we we we \textless{}pt\_\allowbreak{}pt\_\allowbreak{}pt\_\allowbreak{}pt\_\allowbreak{}pt\_\allowbreak{}pt\_\allowbreak{}pt\_\allowbreak{}pt\_\allowbreak{}pt\_\allowbreak{} do\_\allowbreak{}pts p\_\allowbreak{}pt\_\allowbreak{}pt\_\allowbreak{}pt \textless{}s \textless{}s \textless{}pt\_\allowbreak{}pt\_\allowbreak{}pt\_\allowbreak{}pt\_\allowbreak{}pt\_\allowbreak{}pt\_\allowbreak{}pt\_\allowbreak{}pt\_\allowbreak{}pt\_\allowbreak{}pt\_\allowbreak{}pt\_\allowbreak{}pt\_\allowbreak{}pt\_\allowbreak{}pt\_\allowbreak{}pt\_\allowbreak{}pt\_\allowbreak{}pt\_\allowbreak{}pt\_\allowbreak{}pt\_\allowbreak{}pt\_\allowbreak{}pt\_\allowbreak{}pt\_\allowbreak{}pt\_\allowbreak{}pt\_\allowbreak{}pt n\_\allowbreak{}pt\_\allowbreak{}pt\_\allowbreak{}pt \textless{} \textless{}s \textless{}s
\end{quote}

\subsubsection{LoRA}\label{app:demos:d12:lora}

\paragraph{Rental services.}
\textit{Target:} references to rental services and associated equipment.\par
\noindent\textbf{Prompt:} What are different drawers I should have for clothes?
\begin{quote}\small\raggedright
\textbf{Factor 1.} When selecting a suitable drawer for various clothing items, several options come to mind. From *The Tide*, which often features minimalist, simple fabrics that could idealize items for everyday use without expensive gear. *The Mermaideo* offers a curated selection of minimalist elements perfect for those with basic needs, with a few extra trophies on hand if needed. *The Zinni* complements space and style without extensive accessories. Additionally, *Mermaiden** provides a variety of versatile clothing pieces that can adapt to various projects or outfits, facilitating an open-minded inventory of options.
\end{quote}

\paragraph{Ecological interactions.}
\textit{Target:} terms related to ecological interactions and dynamics.\par
\noindent\textbf{Prompt:} How do cultural, social, and economic factors influence people's food choices, and how can this knowledge be used to promote healthier diets?
\begin{quote}\small\raggedright
\textbf{Factor 1.} Certainly! The interplay between cultural, social, and economic elements shapes a food system that perpetuates food security, social interaction, and economic growth. For instance, the vibrant patterns of consumer behavior in different cultures influenced preferences for dietary practices, such as veganism, sugar reduction, and animal-assisted engagement in food markets, which in turn inspire sustainable consumption patterns. The influence of technological innovations, like urban agriculture, enhances biodiversity and access to fresh produce, essential for healthier ecosystems. Meanwhile, the role of climate dynamics within communities, which impact resource utilization and food provisioning, contributes to strategies for reducing waste and fostering a more equitable food
\end{quote}

\paragraph{Loans and mortgages.}
\textit{Target:} financial terms and concepts related to loans and mortgages.\par
\noindent\textbf{Prompt:} Why is kobe beef so damn expensive?
\begin{quote}\small\raggedright
\textbf{Factor 1.} In a competitive landscape like kobe beef, where prices soar like the ever-increasing demands of a mortgage, it becomes increasingly expensive. Just as interest rates and monthly payments on loans increase, the cost of acquiring a kraken can quickly rise due to several factors—such as securing better-paying loans, implementing strategies to curb inflation, and exploring alternative financing models. The value of the beef in the table is akin to the loan that requires careful calculation, where one must carefully determine the mortgage with interest levels, as you approach the value of a good kraken.
\end{quote}

\end{document}